\documentclass[journal]{IEEEtran}
\ifCLASSINFOpdf
  \usepackage[pdftex]{graphicx}
  \graphicspath{{./Graphs/}}
    \graphicspath{{./VisResult/}}
  \usepackage{subfig}
  \usepackage{url}
\else
\fi
\usepackage{amsmath,amssymb, amsfonts}
\usepackage{color}
\usepackage{multirow}

\begin{document}
\title{Efficient Deep Learning of Non-local Features \\for Hyperspectral Image Classification}
% Efficient Deep Learning of Non-local features
%Deep Fully Convolutional Networks \\with Efficient Non-local Module  for \\Hyperspectral Image Classification


\author{Yu Shen, Sijie Zhu, Chen Chen,~\IEEEmembership{Member,~IEEE,} Qian Du,~\IEEEmembership{Fellow,~IEEE,}\\
        Liang Xiao,~\IEEEmembership{Member,~IEEE,}
        Jianyu Chen,~\IEEEmembership{Member,~IEEE,} and % <-this % stops a space
        Delu Pan
\thanks{%This work was partially supported by the National Major Research and Development Program of China (Grant No.2016YFF0103604, No.2016YFC1400903), the National Natural Science Foundation of China (Grant No.61871226, 61571230), Jiangsu Provincial Social Developing Project (Grant No.BE2018727), the Fundamental Research Funds for the Central Universities (Grant No.30918011104), and Key Special Project for Introduced Talents Team of Southern Marine Science and Engineering Guangdong Laboratory (Guangzhou) (GML2019ZD0602). This work was done when Yu Shen was a visiting student at University of North Carolina at Charlotte. (Corresponding Authors: L. Xiao \& J. Chen)

Y. Shen is with the School of Computer Science and Engineering, Nanjing University of Science and Technology, Nanjing 210094, the State Key Laboratory of Satellite Ocean Environment Dynamics, Second Institute of Oceanography, Ministry of Natural Resources, Hangzhou 310012, China, and the Department of Electrical and Computer Engineering, University of North Carolina at Charlotte, NC 28223, USA.

S. Zhu and C. Chen are with the Department of Electrical and Computer Engineering, University of North Carolina at Charlotte, NC 28223, USA.

Q. Du is with the Department of Electrical and Computer Engineering, Mississippi State University, MS 39762, USA.

L. Xiao is with the School of Computer Science and Engineering, Nanjing University of Science and Technology, Nanjing 210094, China. (e-mail:xiaoliang@mail.njust.edu.cn)

J. Chen and D. Pan are with the State Key Laboratory of Satellite Ocean Environment Dynamics, Second Institute of Oceanography, Ministry of Natural Resources, Hangzhou 310012 and also with Southern Marine Science and Engineering Guangdong Laboratory, Guangzhou 511458, China.

}% <-this % stops a space
}


\markboth{Journal of \LaTeX\ Class Files}%
{Shell \MakeLowercase{\textit{et al.}}: Bare Demo of IEEEtran.cls for IEEE Journals}



\maketitle

\begin{abstract}
Deep learning based methods, such as Convolution Neural Network (CNN), have demonstrated their efficiency in hyperspectral image (HSI) classification. These methods can automatically learn spectral-spatial discriminative features within local patches. However, for each pixel in an HSI, it is not only related to its nearby pixels but also has connections to pixels far away from itself. Therefore, to incorporate the long-range contextual information, a deep fully convolutional network (FCN) with an efficient non-local module, named ENL-FCN, is proposed for HSI classification. In the proposed framework, a deep FCN considers an entire HSI as input and extracts spectral-spatial information in a local receptive field. The efficient non-local module is embedded in the network as a learning unit to capture the long-range contextual information. Different from the traditional non-local neural networks, the long-range contextual information is extracted in a specially designed criss-cross path for computation efficiency. Furthermore, by using a recurrent operation, each pixel's response is aggregated from all pixels of HSI. The benefits of our proposed ENL-FCN are threefold: 1) the long-range contextual information is incorporated effectively, 2) the efficient module can be freely embedded in a deep neural network in a plug-and-play fashion, and 3) it has much fewer learning parameters and requires less computational resources. The experiments conducted on three popular HSI datasets demonstrate that the proposed method achieves state-of-the-art classification performance with lower computational cost in comparison with several leading deep neural networks for HSI.
\end{abstract}


% \fi
\IEEEpeerreviewmaketitle

\section{Introduction}
\IEEEPARstart{W}{ith} the rapid development of remote sensing technology, hyperpspectral imaging has been an active research topic in earth observation and remote sensing \cite{camps-valls2014}. Due to the abundant spectral and spatial information, hyperspectral images (HSIs) enable the ability for pixel-wise classification, which aims at assigning a class label to each pixel and generating a thematic map of HSI.

Many machine learning based classifiers have been developed for HSI classification. For example, some pixel-wise methods, such as support vector machine (SVM) \cite{melgani2004}, extreme learning machine (ELM) \cite{su2017} and dictionary representation \cite{chen2011}, have been successful in HSI classification. These methods focus on spectral features for class membership assignment. Due to the existence of noise and mixed pixels in HSIs, pixel-wise approaches can result in considerable errors in thematic maps. To address this problem, on the one hand, kernel tricks, such as the composite kernel SVM (SVMCK) \cite{camps-valls2006},  are proposed to improve the linear separability. On the other hand, researchers have exploited spectral-spatial contextual information and developed a variety of classification algorithms. These methods assume that neighboring pixels share similar spectral signatures and thus belong to the same land-cover types. Based on this assumption, spectral-spatial feature extraction methods, such as Gabor filtering \cite{bau2010, jia2018a}, wavelet transformation \cite{qian2013,hongInvariantAttributeProfiles2020}, object voting \cite{li2015a,zhang2017}, and local binary pattern (LBP) \cite{jia2018,li2015}, are proposed to improve the discrimination of classes. Random field based methods have also been used as post-processing approaches in HSI classification \cite{zhao2016, Wang2017,sun2015}. For example, in \cite{Li2012}, Markov Random Field (MRF) is used as the post-processing strategy after multi-nominal logistic regression (MLR) to improve pixel-wise classification results.

In recent years, deep learning algorithms have been introduced to HSI classification and achieved remarkable success \cite{paolettiCapsuleNetworksHyperspectral2019, fang2017a, chen2013}. Compared with traditional hand-crafted feature extractors, deep learning based methods can automatically extract high-level semantic features from HSI data with several hierarchical representation layers. Various deep learning algorithms have emerged for HSI classification \cite{audebertDeepLearningClassification2019,chengExploringHierarchicalConvolutional2018a}. For example, in \cite{zhouLearningCompactDiscriminative2019}, a stacked auto-encoder (SAE) is proposed to learn the high-dimensional spectral information of HSI. %Similar to SAE, deep belief network (DBN) \cite{zhongLearningDiversifyDeep2017} is also used to exploit the spectral signature from HSI.
Recently, convolutional neural network (CNN) \cite{chenDeepFeatureExtraction2016a,xu2020csa}, recurrent neural network (RNN) \cite{mou2017, hangCascadedRecurrentNeural2019}, generative adversarial network (GAN) \cite{zhuGenerativeAdversarialNetworks2018a, fengClassificationHyperspectralImages2019}, long short-term memory (LSTM) \cite{huSpatialSpectralFeatureExtraction2020}, and graph convolutional network (GCN) \cite{wanMultiscaleDynamicGraph2019,mouNonlocalGraphConvolutional2020} have been developed for HSI classification. Among these approaches, CNN based classifiers can efficiently extract spectral-spatial contextual information of HSI \cite{leeGoingDeeperContextual2017, heHeterogeneousTransferLearning2019}. Therefore, many CNN based algorithms have been proposed for improving HSI classification performance \cite{gongCNNMultiscaleConvolution2019, paolettiDeepLearningClassifiers2019,meng2019fully}. For instance, in \cite{makantasisDeepSupervisedLearning2015}, HSI is first divided into small size patches (e.g., $3\times3$ or $5\times5$), and then these patches are fed into a CNN for training. Through back propagation, the convolutional kernels of CNN gradually obtain the deep-level representation of HSI. Other CNN variants have also been successfully applied to HSI classification. For example, 3-D CNN with 3-D convolutional kernels (e.g., $3\times3\times3$) is employed for spectral-spatial feature extraction of HSI \cite{liSpectralSpatialClassification2017b}.  More recently, the fully convolutional network (FCN) is proposed for classification, which allows a whole HSI as input to the network without patch-based processing (i.e., extracting neighboring regions for each pixel) of HSI \cite{jiaoDeepFullyConvolutional2017, zhengFPGAFastPatchFree2020}. Correspondingly, the output of FCN is a per-pixel classification map.

Although the CNN framework is able to extract spectral-spatial contextual information by convolutional kernels, the extracted information is still limited in a fixed small patch and the non-local information is barely utilized. For a pixel in a remote sensing image, it is not only related to its nearby pixels but also has connections to some pixels far away, e.g., the same land cover may appear in different locations of the scene. Therefore, to extract long-range contextual information, an intuitive approach is to expand the kernel size in convolution operations. In \cite{li2019multiscale}, a multi-scale approach is used to fuse the spectral-spatial information extracted from different scales of convolutional kernels. Although a larger convolutional kernel can acquire more contextual information, it may lead to over-smooth and blur image details. 

Recently, two strategies have been presented to incorporate long-range information in deep learning networks. One solution is to combine a deep neural network with conditional random field (CRF) for HSI classification \cite{cao2018}. In \cite{xuPatchwiseClassificationSpectralSpatial2019}, Xu \emph{et al.} introduce CRF after a CNN to balance local and non-local information. However, this approach divides the training step into two individual phases and thus the global optimization is difficult to achieve. Another solution is to apply non-local based neural modules \cite{wangHyperspectralImageClassification2019}. 
Non-local based methods consider each pixel's response from the whole image, which was first used in image denoising \cite{buadesNonlocalAlgorithmImage2005}. 
After that, non-local based methods are integrated with a variety of machine learning methods, such as dictionary learning \cite{zhangNonlocalWeightedJoint2014} and collaborative representation \cite{liColumngenerationKernelNonlocal2014}, for HSI classification. Inspired by the non-local mean filtering algorithm \cite{buadesNonlocalAlgorithmImage2005}, a non-local network \cite{Wang_2018_CVPR} is proposed by Wang \emph{et al.} for incorporating long-range dependencies, which is able to compute the responses of a pixel with all other pixels in the image. In \cite{sunSpectralSpatialAttentionNetwork2019}, Sun \emph{et al.} propose to add non-local modules to a CNN framework, and then spectral-spatial information from non-local areas can be captured for HSI classification. However, the HSI is still processed in individual small patches and the long-range contextual information is difficult to be modeled. Moreover, the computational cost of the non-local module is very high. Compared with common RGB images, HSI is a data cube that has both high spatial and spectral dimensions. The non-local operation will consume \textit{significant computational resources} because the correlation of each pixel is calculated based on the whole HSI.
% added by sijie: 

From the analysis of prior work, \textit{how to efficiently model long-range contextual information remains unsolved for HSI classification.} %Therefore, it remains a challenging problem to efficiently model long-range contextual information of HSI.
To this end, in this paper, an FCN with an efficient non-local module, named ENL-FCN, is proposed to extract both \textit{local and non-local} spectral-spatial information. Specifically, the FCN extracts the local spectral-spatial information from HSI, while an efficient non-local module \cite{Huang_2019_ICCV} is embedded in the network to aggregate each pixel's response from the whole HSI. 
The proposed efficient non-local module is developed from the criss-cross attention module \cite{Huang_2019_ICCV}, which was used for semantic segmentation. Compared with the original non-local module, the efficient non-local module computes the pixels' relation in a criss-cross path for computation efficiency. By using a recurrent operation, each  pixel’s  response  is  aggregated  from  all  pixels of HSI.
Furthermore, to help understand the long-range contextual information modeling of the network, we conduct visualization analysis to observe the correlation of each pixel in the entire image.
%By visualising the single pixel's attention map, we can observe the pixel's response from whole image. 
%Finally, an end-to-end learning framework is generated for HSI classification. 
The main contributions of this paper can be summarized as follows.

% sijie:
%1) We propose a FCN with efficient non-local module which outperforms state-of-the-art methods on four popular HSI classification datasets. The whole pipeline is end-to-end without complex preprocessing like \cite{wanMultiscaleDynamicGraph2019}.
\begin{itemize}
\item 
An end-to-end trainable deep learning framework is proposed for HSI classification. By integrating FCN with the efficient non-local module, the framework is able to extract both local and non-local information.

%2) The efficient non-local module requires fewer parameters and the computational memory is saved. Compared with the traditional non-local neural network, the attention module has no need to generate a huge attention map to record the relations of each pair of pixels.

%Theoretical and empirical analyses show that the proposed framework can capture lone-range contextual information due to the efficient non-local module. Moreover, 
\item
The efficient non-local module offers computation and memory efficiency, making it feasible to stack multiple efficient non-local modules for further boosting the performance.

% sijie: 
%Compared with the original non-local module \cite{Wang_2018_CVPR}, the efficient non-local module saves more than 3 times computational memory and 100 times learning parameters. Extensive experiments demonstrate that the proposed method achieves state-of-the-art classification performance on several HSI datasets.

\item
Compared with the original non-local module \cite{Wang_2018_CVPR}, the efficient non-local module saves more than three times computational memory and one hundred times learning parameters. Extensive experiments demonstrate that the proposed ENL-FCN achieves state-of-the-art classification performance on several HSI datasets. Comprehensive ablation studies and parameter analysis are carried out to validate the effectiveness of each component of the proposed method.

%Comprehensive ablation studies and parameter analysis are carried out to validate the effectiveness of each component of the proposed method. 

\end{itemize}

%3) Extensive experiments are conducted to figure out the effect of every specific design in our framework, including the kernel size, component position, recurrent operation, ...... (add other ablation study here). We also show visualization results of the correlation map to illustrate how the long-range spatial information is represented in our model.
%3) By using FCN as backbone of the proposed network, the efficient non-local module is able to extract long-range features from an entire HSI. The proposed method achieves state-of-the-art classification performance over several HSI datasets.

% (should emphasize this in the introduction somewhere, other papers don't have the visualization analysis. As they only show the classification performance, the attention map may not work as they expected.)
The remainder of this paper is organized as follows. In Section \ref{SecRelatedWork}, we review the works related to HSI classification.  Section \ref{SecMethod} introduces the proposed framework. Section \ref{SecExperiment} provides comprehensive experimental results and parameter analysis. Finally, conclusion is drawn in Section \ref{SecConclusion}.

\begin{figure*}[!t]
	\centering
	\includegraphics[width=0.94\textwidth]{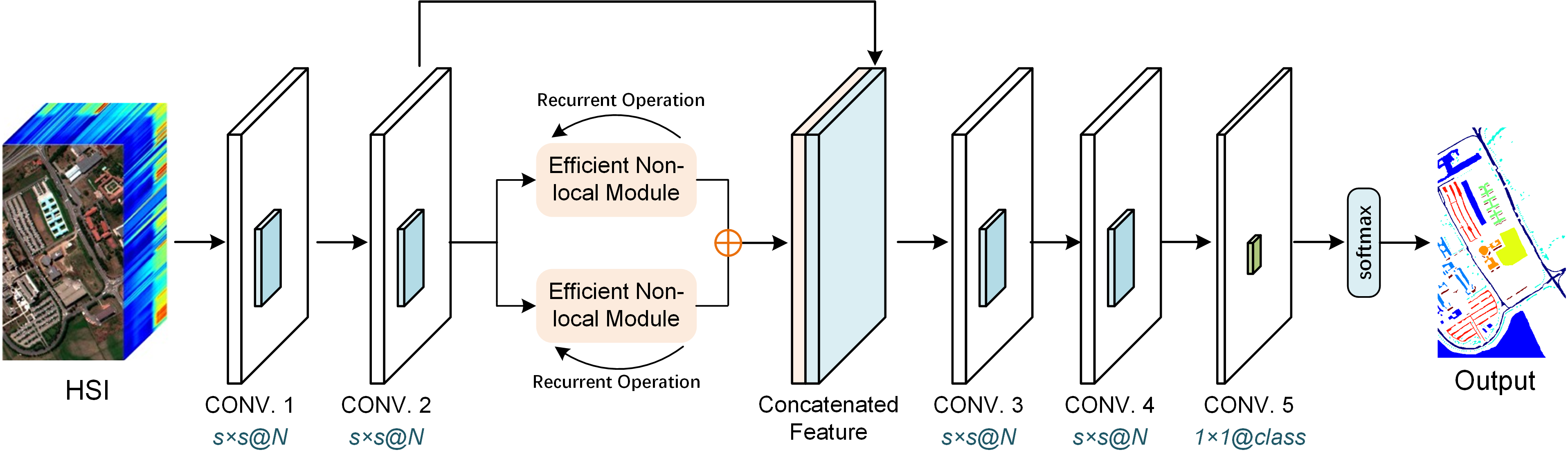}
	%\captionsetup{font={footnotesize}}
	\caption{Network architecture of the proposed ENL-FCN. The HSI is first fed into FCN for local spectral-spatial information extraction. After Conv.2 layer of the network, two efficient non-local modules are embedded in parallel for integrating long-range contextual information of HSI. $s$ denotes the convolutional kernel size, $N$ denotes the number of convolutional kernels. }
\label{figFlowchart}
\end{figure*}

\section{Related Work}
\label{SecRelatedWork}

%In this section, we present some related works about machine learning based HSI classification. Furthermore, we review the methods that integrate long-range contextual information under deep learning frameworks.

\subsection{HSI Classification with Machine Learning Methods}
Supervised HSI classification has been widely investigated and many machine learning based methods have been proposed, such as logistic regression \cite{Li2012}, SVM \cite{melgani2004}, ELM \cite{su2017} and sparse representation \cite{yuGlobalSpatialLocal2020}. Among these methods, SVM is effective in HSI classification with limited training samples. The classic SVM treats each pixel independently and the class label is determined by its spectral signature. Due to the existence of spectral noise, the classification results of these pixel-wise approaches are still unsatisfactory. To overcome this limitation, the spatial correlation of HSI is considered. 
%For instance, composite kernel SVM \cite{camps-valls2006} and ELM \cite{chen2014} are applied to improve the linear separability of HSI and have achieved better classification performance.
Therefore, more works include spectral-spatial information for HSI classification \cite{hongLearningRobustLocal2017, hongLearningPropagateLabels2019}. Filtering based methods, such as Gabor filtering \cite{jia2018a} and bilateral filtering \cite{kang2014}, are utilized for HSI classification. Random fields \cite{Wang2017}, relearning methods, and object voting \cite{zhong2014}, are also developed to exploit spatial correlation of pixels.

The aforementioned methods represent HSI pixels with handcrafting features, which highly depend on experts' experience. Different from these traditional classifiers, deep learning methods can automatically obtain high-level features by several representation layers. In \cite{chenDeepLearningBasedClassification2014}, Chen \emph{et al.} first introduce a deep learning method in HSI classification, which uses a stacked auto-encoder (SAE) for high-level feature extraction. 
%Similar to SAE, deep belief network (DBN) \cite{zhongLearningDiversifyDeep2017} is also employed to exploit spectral feature from HSI. 
Recently, CNN based approaches have emerged for HSI classification. 
For instance, in \cite{huDeepConvolutionalNeural2015}, a five-layer 1-D CNN is introduced to extract spectral features and has achieved promising performance. To further exploit the spectral-spatial information of HSI, more effective CNN architectures have developed. In \cite{huDeepConvolutionalNeural2015}, 2D-CNN is encoded to explore spatial correlation and the classification is conducted with a multi-layer perceptron.
%In \cite{zhangMultiScaleDenseNetworks2019}, dense residual connection is used to enhance the discriminative power of the network.  
In \cite{paolettiDeepPyramidalResidual2019}, more convolutional layers are grouped and used in pyramidal bottleneck to increase the depth of CNN. 
%In \cite{liDeepFeatureFusion2020}, a two-stream CNN is applied to separately extract spectral and spatial features. 
Moreover, a multi-scale representation is used to fuse the spectral-spatial information from different scales to obtain more discriminative features.

There are mainly two type frameworks or pipelines for CNN based HSI classification. One is the patch-based framework, where local patches are first extracted from HSI with fixed window size (e.g., $7\times7$). This framework involves modeling a patch classification task, where each extracted patch is assigned with a class label. Usually, the label of a small patch is determined by that of the center pixel in the patch. Although patch-based frameworks have achieved remarkable success in HSI classification, it requires much computational cost because a series of overlapping patches need to be generated, and it is difficult for CNN to sufficiently model non-local information. To alleviate these issues, FCN based frameworks have emerged as an alternative solution. In an FCN framework, the fully connected layer of CNN is replaced by the $1\times1$ kernel. In this way, FCN enables the whole HSI as input and produces a corresponding class map. In \cite{li2018a}, a pre-trained FCN is used to extract HSI feature. However, this framework is not end-to-end and the extracted feature is fed into SVM for classification. In \cite{zhengFPGAFastPatchFree2020}, an end-to-end FCN is proposed and achieves better classification performance than some patched based frameworks.

\subsection{Non-local Information Modeling}

With the rapid development of deep learning methods, researchers have exploited non-local information in deep neural networks. Similar to traditional methods, random field based methods are first combined with CNN for semantic segmentation \cite{vemulapalli2016gaussian}. After that, a trainable end-to-end CRF based network is introduced \cite{zheng2015conditional}. On the other hand, non-local based neural networks have also developed. Inspired by the classic non-local mean filtering \cite{buadesNonlocalAlgorithmImage2005}, Wang \emph{et al.} \cite{Wang_2018_CVPR} proposed a non-local neural network to capture long-range pixel dependencies, which achieves great improvement on the video classification task. Subsequently, a variety of non-local network extensions have been introduced \cite{wangHyperspectralImageClassification2019}. Fu \emph{et al.} \cite{fu2019dual} present a dual attention segmentation framework, which applies non-local modules to spatial and spectral domain respectively. In \cite{liu2018non}, Liu \emph{et al.} demonstrate that non-local similarity is an effective prior in image restoration. 

\begin{figure*}[!t]
	\centering
	\subfloat[]{
		\includegraphics[width=0.45\linewidth]{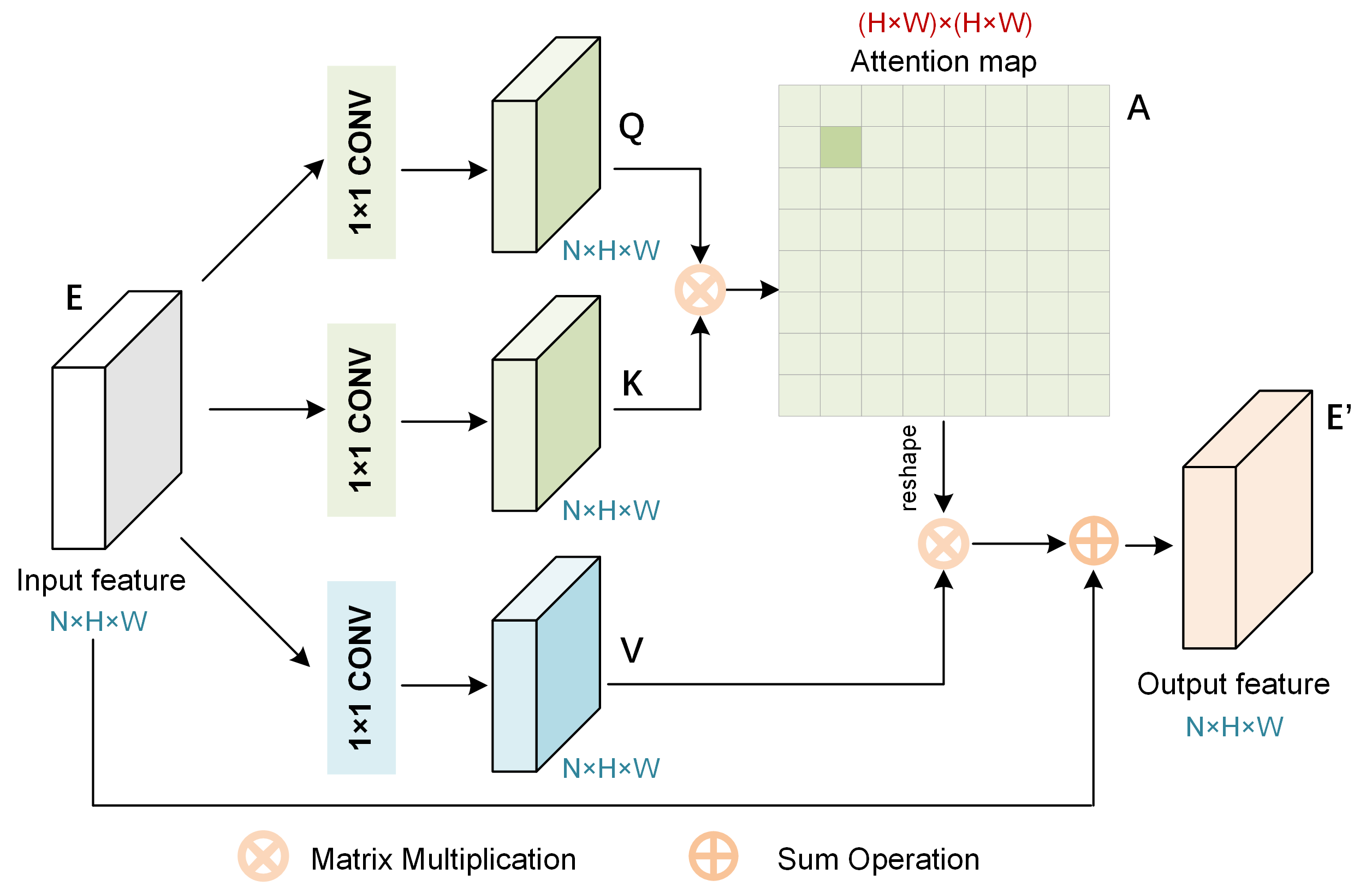}
	} 
	\subfloat[]{
		\includegraphics[width=0.45\linewidth]{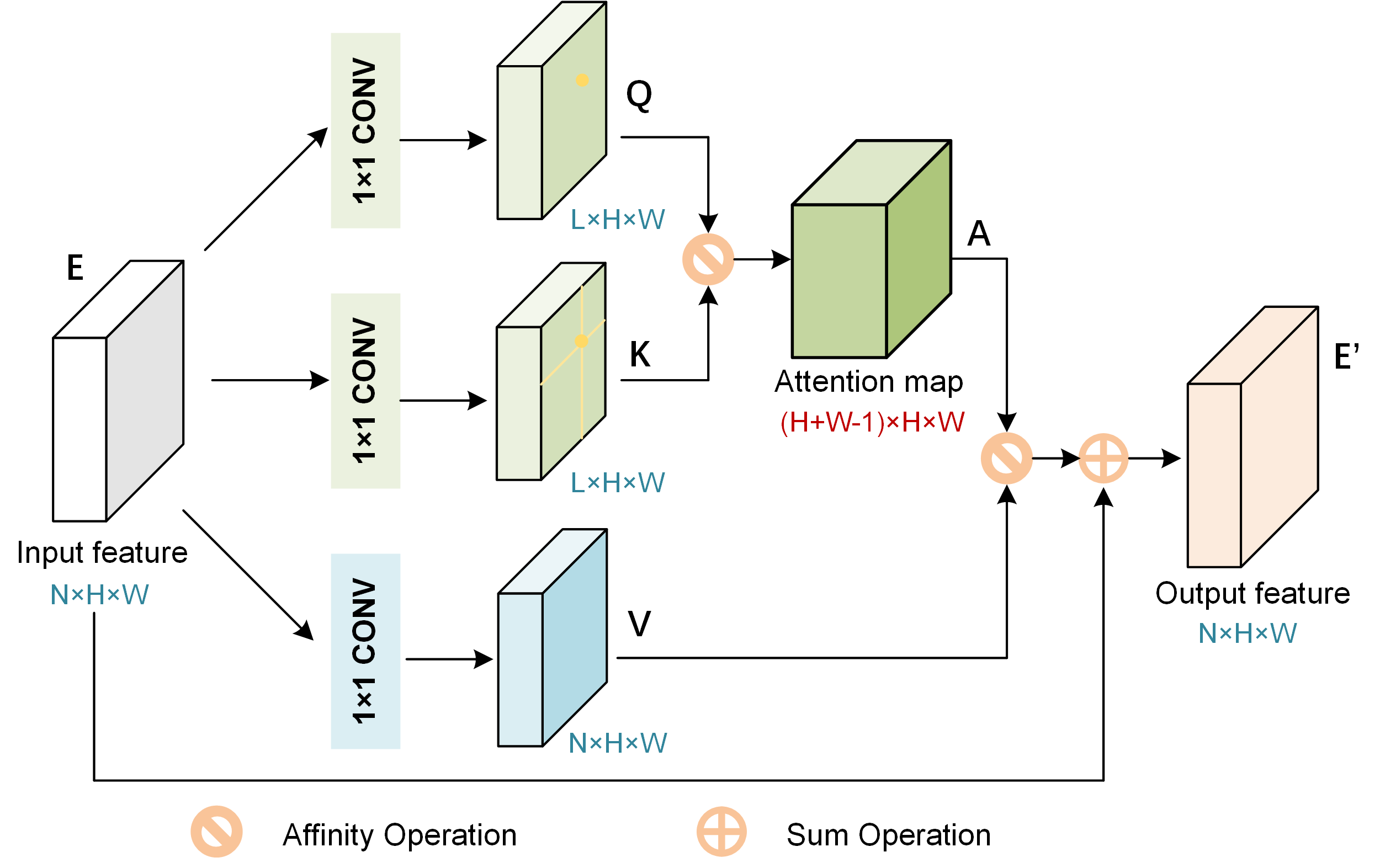}
    }  
	\caption{The architectures of (a) the original non-local module, and (b) the efficient non-local module. The generation of attention map $\mathbf{A}$ is different in two modules. In the original non-local module, $\mathbf{A}$ is obtained by simple matrix multiplication. In the efficient non-local module,  $\mathbf{A}$ is computed by the affinity operation. }
\label{figCC}
\end{figure*}

However, the computational memory usage of the non-local neural module \cite{Wang_2018_CVPR} is extremely high. Since each pixel's response is calculated on the whole image, the cost of computing memory is prohibitive when the input data has a large size. To overcome this drawback, the pooling strategy is widely used to reduce the input image or feature map size. However, this operation would lose fine image details. Therefore, in this study, our motivation is to design a deep FCN classification framework that \textit{efficiently integrates non-local information with a lower computation cost}. Specifically, an FCN is first built as the backbone for local spectral-spatial feature learning. Then an efficient non-local module \cite{Huang_2019_ICCV} is introduced to explore spatial correlation in the whole image. As a result, the entire framework is capable of capturing local and non-local information and can be trained end-to-end for HSI classification.

\section{The Proposed Method}
\label{SecMethod}

The proposed network architecture is illustrated in Fig. \ref{figFlowchart}. Let an HSI be $\mathbf{X}\in \mathbb{R}^{B\times H\times W}$, and the corresponding output be $\mathbf{Y}\in \mathbb{R}^{C\times H\times W}$ indicating the class probability of each pixel, where $H$, $W$, $B$, and $C$ represent the height, width, number of spectral bands, and number of land-cover classes, respectively. As shown in Fig. \ref{figFlowchart}, the original HSI is used as input to the network. Two main parts are in the proposed network, i.e., an efficient non-local module for long-range contextual information learning and an FCN as the backbone network. The details of each part are described in the following subsections.

\subsection{Original Non-local Module}
\label{subsecBasicNL}
The non-local network in \cite{Wang_2018_CVPR} integrates the non-local mean operation in the deep neural networks. Let $\mathbf{x}_i$ be the input signal at position $i$. Then a general non-local operation in a deep neural network can be defined as:
\begin{equation}
\mathbf{y}_i=\frac{1}{Z}\sum_{\forall j}f(\mathbf{x}_i, \mathbf{x}_j)g(\mathbf{x}_j),
\label{eqNLoperation}
\end{equation}
where $\mathbf{y}_i$ is the $i$-th output, and $\mathbf{x}_j$ is the $j$-th signal used to compute the response of $\mathbf{x}_i$, the pairwise function $f(\cdot)$ calculates the relationship of $\mathbf{x}_i$ and $\mathbf{x}_j$, the unary function $g(\cdot)$ calculates the representation of input signal at position $j$, and $Z$ is a normalization factor.

In a deep neural network, $g(\mathbf{x}_j)$ is considered as a weight matrix, which can be implemented by a $1\times 1$ convolutional kernel. For the pairwise function $f(\mathbf{x}_i, \mathbf{x}_j)$, a dot-product similarity is used, which is defined as
\begin{equation}
f(\mathbf{x}_i, \mathbf{x}_j)=\theta(\mathbf{x}_i)^T\phi(\mathbf{x}_j),
\label{eqDotProduct}
\end{equation}
where both $\theta(\mathbf{x}_i)$ and $\phi(\mathbf{x}_j)$ can be implemented with $1\times 1$ convolutional kernels. Then a non-local neural unit is created. Fig. \ref{figCC}(a) shows the basic architecture of a non-local module. 

\subsection{Efficient Non-local Module}
According to Eq. \ref{eqDotProduct}, each pixel's response is calculated from all other pixels of HSI. For an image with a size of $H\times W$, as denoted in Fig. \ref{figCC}(a), the attention feature acquired by Eq. \ref{eqDotProduct} has a size of $HW \times HW$. Obviously, this operation would consume much computational memory resources. Therefore, in order to reduce memory usage, an efficient non-local module is adopted as the learning unit to integrate long-range contextual information of HSI. As shown in Fig. \ref{figCC}(b),  let $\mathbf{E}\in \mathbb{R}^{N\times H\times W}$ be the output feature map after two convolutional operation, where $N$ is the number of convolutional kernels. At first, the efficient non-local module applies $1\times 1$ convolutional filter on $\mathbf{E}$ to generate an individual feature map, which can be formulated as
\begin{equation}
f_q^l(\mathbf{E})=\sigma (\mathbf{W}^l_q \ast \mathbf{E} + b^l_q), \quad l=1,...,L,
\label{eq1Conv}
\end{equation}
where $W^l_q$ denotes the $l$th convolutional kernel, $\ast$ is the convolutional operation, $b^l_q$ is the bias, and $\sigma$ represents the activation function (e.g., sigmoid). $L$ is the band number of output feature and is smaller than $N$. Here we denote the output feature as $\mathbf{Q}\in \mathbb{R}^{L\times H\times W}$. A similar $1\times 1$ convolution is applied and we can obtain another feature map $\mathbf{K}\in\mathbb{R}^{L\times H\times W}$ which has the same shape as $\mathbf{Q}$.

To reduce the memory cost, the dependencies of long-range pixels are aggregated in a criss-cross manner. Let $\mathbf{Q}_i \in \mathbb{R}^L$ be a feature vector at the spatial position of $i$ from $\mathbf{Q}$. Meanwhile, we can extract a feature vector set $\mathbf{\Omega}_i$ from  the same row or column $i$ in $\mathbf{K}$, where $\mathbf{\Omega}_i \in \mathbb{R}^{L\times(H+W-1)}$. $H+W-1$ is the number of feature vectors that are in the same row or column of $i$. Then we can calculate the first non-local correlation of pixel $i$ from its criss-cross pass by the affinity operation:
\begin{equation}
\mathbf{d}_i=\mathbf{Q}_i \mathbf{\Omega}_i, 
\label{eqdi}
\end{equation}
where $\mathbf{d}_i\in \mathbb{R}^{H+W-1}$. Obviously, $\mathbf{d}_i$ records the feature relations between $\mathbf{Q}_i$ and $\mathbf{\Omega}_i$. We apply the same operation on each pixel of $\mathbf{Q}$, and then a feature map $\mathbf{D}$ can be generated, where $\mathbf{D}\in \mathbb{R}^{(H+W-1)\times H\times W}$. A softmax layer is used along the spectral dimension of $\mathbf{D}$ to obtain an attention map $\mathbf{A}\in \mathbb{R}^{(H+W-1)\times H\times W}$.

Meanwhile, another $1\times 1$ covolutional filter is operated on feature $\mathbf{E}$ to generate $\mathbf{V}\in \mathbb{R}^{N\times H\times W}$. Similarly, a feature vector $\mathbf{A}_i\in \mathbb{R}^{H+W-1}$ can be extracted from $\mathbf{A}$ at spatial position $i$. Correspondingly, the feature vector set $\mathbf{\Phi}_i\in \mathbb{R}^{N\times(H+W-1)}$ is extracted from the same row or column at position $i$ in $\mathbf{V}$. Then non-local information related to pixel $i$ can be described as:
\begin{equation}
\mathbf{E}_i^{'}=\mathbf{A}_i \mathbf{\Phi}_i^T+\mathbf{E}_i,
\label{eqE}
\end{equation}
where $\mathbf{E}^{'}_i\in \mathbb{R}^N$. The same affinity operation is applied on each position of $\mathbf{A}$ and thus a new feature map $\mathbf{E}^{'}\in \mathbb{R}^{N\times H\times W}$ can be obtained. To enhance the local feature $\mathbf{E}$, the contextual information is added to $\mathbf{E}$. In the efficient non-local module, the output $\mathbf{E'}$ is  in the same shape ($N\times H\times W$) with the input feature $\mathbf{E}$. Therefore, this learning unit can be embedded in any position of a deep neural network for non-local information extraction. The architecture of the efficient non-local module is illustrated in Fig. \ref{figCC}(b).

\begin{figure}[t]
	\centering
	\subfloat[]{
		\includegraphics[width=0.48\linewidth]{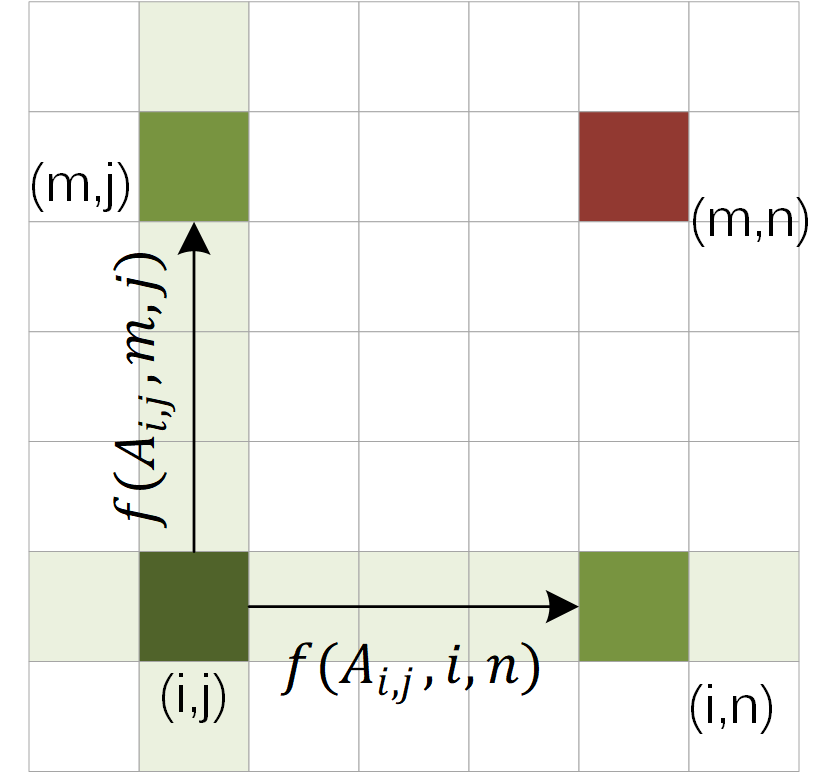}
		\label{fig:1} } 
	\subfloat[]{
		\includegraphics[width=0.48\linewidth]{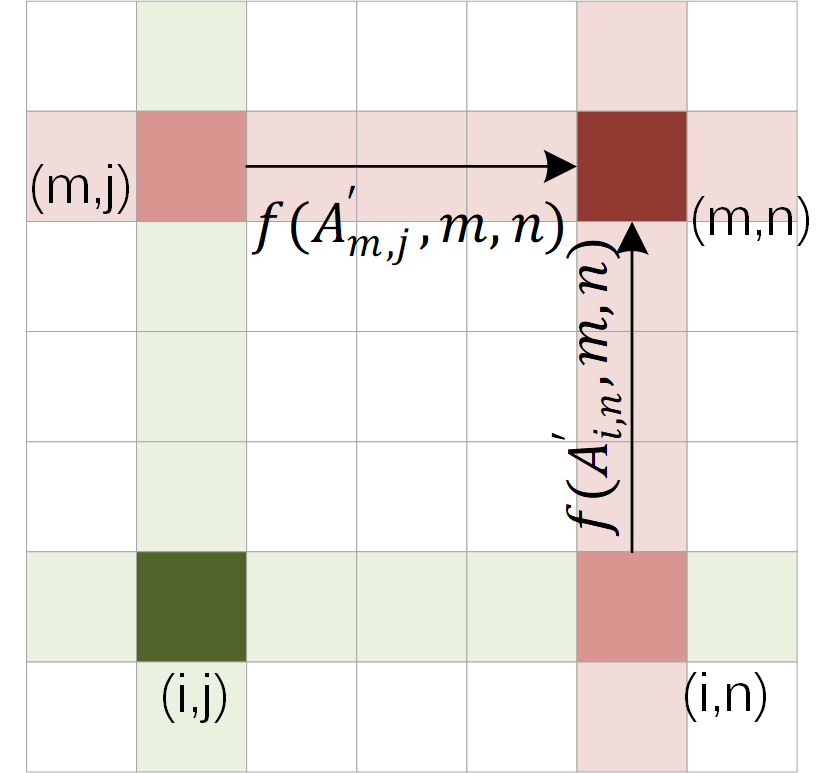}
		\label{fig:2}}        \\      	
	\caption{(a) Message passing before the recurrent operation. (b) Message passing after the recurrent operation.}
	\label{figMessagePass}
\end{figure}

\subsection{Message Passing via Recurrent Operation}
\label{subsecRCCA}
Although the non-local information can be aggregated via the criss-cross path, the relations between pixels are still sparse because the contextual information is mainly captured from horizontal and vertical directions. To obtain more contextual information, the recurrent operation is introduced based on the efficient non-local module. First, the feature $\mathbf{E}$ extracted from FCN is used as input of the efficient non-local module and the feature map $\mathbf{E'}$ can be obtained as output. Then we take $\mathbf{E'}$ as the new input of the efficient non-local module and a new output $\mathbf{E''}$ can be produced. 

Fig. \ref{figMessagePass} shows the message passing procedure between any two pixels that are not in the same row or column. Let  $\mathbf{A'}$ be the attention map obtained after the recurrent operation. Before the recurrent operation, the feature map is obtained from position $(i,j)$ to position $(m,j)$ and $(i,n)$. Thus, we define the mapping functions as $f(A_{i,j}, m, j)$ and $f(A_{i,j}, i, n)$. As pixel $(m,j)$ and $(i,n)$ are in the same row and column with the target position $(m,n)$, similar mapping functions can be formulated as $f(A'_{m,j}, m, n)$ and $f(A'_{i,n}, m, n)$. Therefore, we can represent the message passing procedure between pixels $(i,j)$ and $(m,n)$ as
\begin{equation}
\begin{split}
\mathbf{E''}_{m,n} \leftarrow [f(A_{i,j}, m, j)\cdot f(A'_{m,j},m,n)+ \\
f(A_{i,j},i,n)\cdot f(A'_{i,n},m,n)]\cdot \mathbf{E}_{i,j}.
\end{split}
\label{eqMessage}
\end{equation}

\begin{figure}[t]
	\centering
	\subfloat{
		\includegraphics[width=0.98\linewidth]{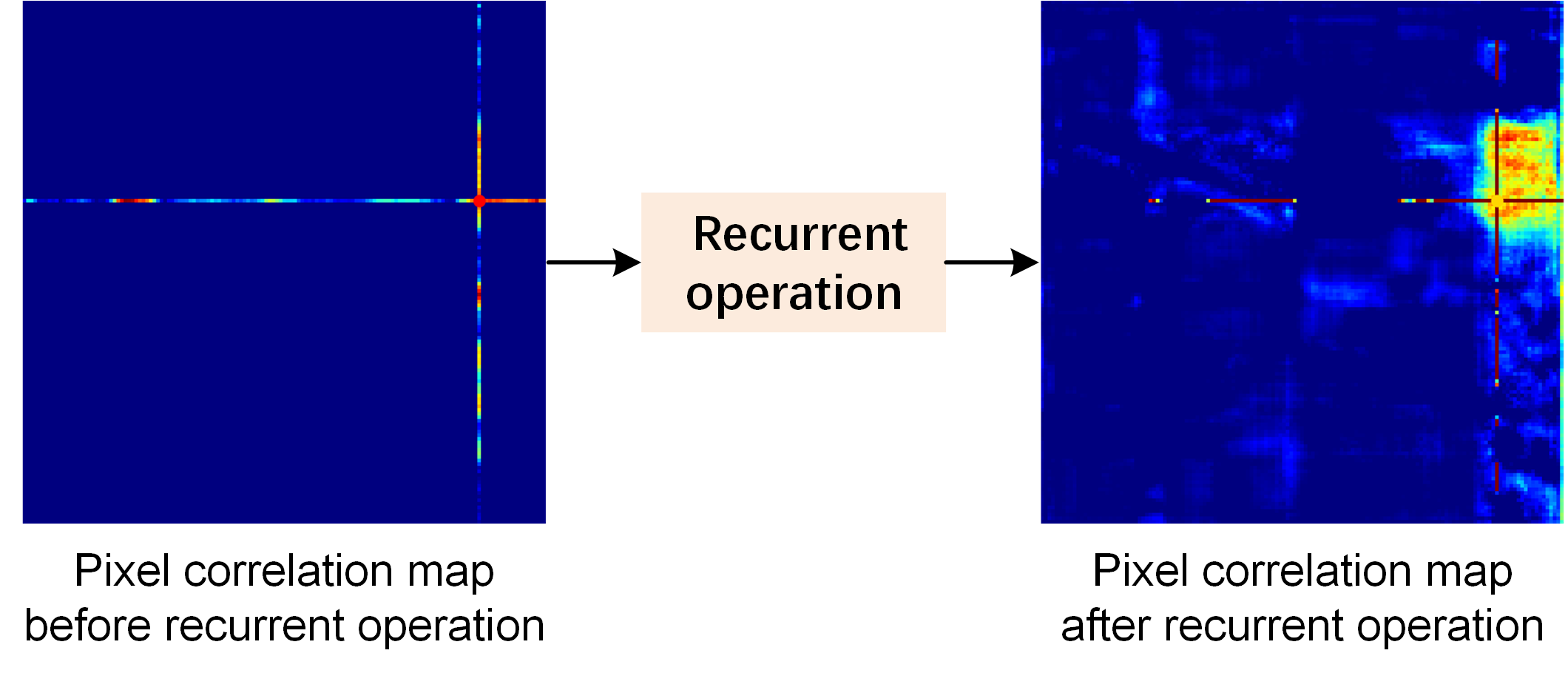}
		\label{fig:1} } 
	\caption{Pixel correlation map after using the recurrent operation in the efficient non-local module.}
	\label{figRCCApixelmap}
\end{figure}

Generally, in the recurrent efficient non-local module, the relation between two pixels can be captured by a criss-cross based message passing procedure. Fig. \ref{figRCCApixelmap} illustrates the pixel correlation map after applying the recurrent operation, where a pixel is selected to show the message passing procedure when applying the recurrent operation. From Fig. \ref{figRCCApixelmap}, we can observe more contextual information is captured around the selected pixel. The recurrent operation does not need more learning parameters, yet can provide more contextual information with one recurrent operation. Therefore, a better classification performance can be achieved.

\subsection{Computational Complexity and Memory Usage}
\label{subCost}
As discussed in Section \ref{subsecBasicNL}, the correlation between two pixels is calculated by a dot-production in the original non-local module. Since the response of a pixel is calculated from a whole image, the dot-product operation requires $HW\times HW\times L$ times of multiplication. In contrast, the affinity operation in the efficient non-local module needs  $(H+W-1)\times HW\times L$ times of multiplication. 
%edited by sijie:
For simplicity, we assume that $H=W=L=n$. Then the computational complexity of the original non-local module \cite{Wang_2018_CVPR} is $\mathcal{O}(n^5)$ and is reduced to $\mathcal{O}(n^4)$ by using the efficient module.

As displayed in Fig. \ref{figCC}, the attention map of the original non-local module has a size of $(H\times W)\times (H\times W)$, while the one from efficient non-local module is of size $(H+W-1)\times H\times W$. For an image with $145\times 145$ pixels, the attention map of the original non-local module would consume 1760MB of memory while the memory usage of the efficient non-local module is only about 24MB. Therefore, by using the efficient non-local module, the memory footprint can be significantly reduced. 

\subsection{Classification with Fully Convolutional Network}
Different from previous CNN based classifiers using local patches as training data, a whole HSI is fed into the FCN for feature extraction. In the convolutional layer,  a regular convolutional kernel is used to extract spectral-spatial features. The convolutional layer can be written as:
\begin{equation}
\mathbf{E}^i=\sigma(\sum_{j=1}^B(\mathbf{W}^i \ast \mathbf{X}^j)+b^i), \quad i=1,...,N,
\label{eqConv}
\end{equation}
where $\mathbf{E}^i$ is the $i$-th channel of the feature map of convolutional layer, $\mathbf{W}^i$ is the convolutional kernel, $\mathbf{X}^j$ is the $j$th channel of the input HSI or previous convolutional layer, $b^i$ is the bias term, and $\sigma(\cdot)$ is an activation function (e.g., sigmoid function in this study).

To further enhance the robustness of the network, we fuse the features from the low-level convolutional layer and efficient non-local module, which can be expressed as
\begin{equation}
\mathbf{E}_{cat}=[\mathbf{E}, \mathbf{E}^{''}],
\label{eqCat}
\end{equation}
where $\mathbf{E}_{cat}$ represents the concatenated feature, which is further passed to the subsequent convolutional layers. By using the concatenation operation, features from low-level convolutions can be reused, leading to enhanced feature representation.

Although an FCN can be trained using the whole HSI, there are unlabeled pixels in an HSI. To ensure only the labeled training samples are considered in the loss function, a training mask is used. Let $\mathbf{Y}^{train}\in \mathbb{R}^{H\times W}$ be the training map of an HSI, where $\mathbf{Y}^{train}(i,j)\in \{0, 1,...,C\}$ with $0$ representing an unlabeled pixel. A training mask can be defined as
\begin{equation}
\mathbf{H}(i,j)= \begin{cases}
1, & if  \quad \mathbf{Y}^{train}(i,j)>0 \\
0, & else.
\end{cases}
\label{eqMask}
\end{equation}
$\mathbf{H}(i,j)$ is used to indicate whether pixel $(i,j)$ should be included in the loss function. Then, the final cross-entropy loss function can be formulated as
\begin{equation}
\mathcal{L}=-\frac{1}{\sum\limits_{i=1}^H \sum\limits_{j=1}^W H(i,j)}\sum_{i=1}^H \sum_{j=1}^W \sum_{c=1}^C \mathbf{H}(i,j) t_{ij}^c \mathrm{log}\mathbf{P}^c_{ij},
\label{eqLossFunc}
\end{equation}
where $t_{ij}^c$ denotes the $c$th value of a label vector at position $(i,j)$, and $\mathbf{P}^c_{ij}$ represents the corresponding probability.

\section{Experiments}
\label{SecExperiment}
\subsection{Datasets}
To evaluate the classification performance of the proposed method, three widely used HSI datasets are tested in our experiments, including Indian Pines (IP), Pavia University (PU) and the Kennedy Space Center (KSC). The details of the three datasets are described as follows.

(1) \emph{Indiana Pines}: Indian Pines (IP) is captured by the AVIRIS sensor over Northwestern Indiana. It was generated with 220 spectral bands (400-2500nm) and 20m per pixel resolution. The whole dataset is a 145$\times$145$\times$ 200 data cube and the spectral bands are reduced to 200 after the water absorption and noisy bands are removed. The ground truth includes 10366 labeled pixels for 16 land cover classes. In our experiment, 10\%, 1\% and 89\% of samples from each class are randomly selected as training, validation and testing sets, respectively. The detailed sample numbers are listed in Table \ref{tabDataIP}.

\begin{table}[!t]
	\centering
	\caption{Land cover classes and numbers of samples for training, validation and testing on IP.}
	\begin{tabular}{ccccc}
		\hline
		\hline
		No. & Land cover class & Training & Validation & Testing\\
		\hline
		1 &	Alfalfa &	5  &1 & 48\\
		2&	Corn-notill & 143 &	14 &1277\\
		3&	Corn-mintill       &83     	 &8 &743 \\
		4&	Corn &23	&    2 &209\\
		5&	Pasture   &49  &     	4 &444\\
		6&	Trees/Grass	  &74       &  7    &444\\
		7&	Pasture-mowed	 &2         & 1        &23\\
		8&	Hay-windrowed &	48 &      4 &437\\
		9&	Oats &	2    &1   &17\\
		10&	Soybean-notill &96	   &9 &863\\
		11&	Soybean-mintill	   &246 &24      &2198 \\
		12&	Soybean-cleantill  &61	    &6     & 547  \\
		13&	Wheat	    &21        & 2 &189 \\
		14&	Woods	    & 129      & 12 &1153 \\
		15&	Building-Grass	    &38           &3 &339   \\
		16&	Stone-Steel-Towers &9	&1   &85\\
		\hline
		\multicolumn{2}{c}{Total}  &1029 &99	&9238  \\	
		\hline  \hline
	\end{tabular}
\label{tabDataIP}
\end{table}

\begin{table}[!t]
	\centering
	\caption{Land cover classes and numbers of samples for training, validation and testing on PU.}
	\begin{tabular}{ccccc}
		\hline
		\hline
		No. & Land cover class & Training & Validation & Testing\\
		\hline
		1 &	Asphalt &	67  &67 & 6497\\
		2&	Meadows & 187 &	187 &18275\\
		3&	Gravel      &21     	 &21 &2057 \\
		4&	Tress &31	&   31 &3002\\
		5&	Painted-metal-sheets   &14  &     14 &1317\\
		6&	Bare-soil	  &51      &  51    &4927\\
		7&	Bitumen	 &14       & 14       &1302\\
		8&	Self-Blocking Bricks &	37 &   37 &3608\\
		9&	Shadows &10    &10  &927\\
		\hline
		\multicolumn{2}{c}{Total}  &432 &432	&41912  \\	
		\hline  \hline
	\end{tabular}
\label{tabDataPU}
\end{table}

\begin{table}[!t]
	\centering
	\caption{Land cover classes and numbers of samples for training, validation and testing on KSC.}
	\begin{tabular}{ccccc}
		\hline
		\hline
		No. & Land cover class & Training & Validation & Testing\\
		\hline
		1 &	Scrub &	18  &4 & 325\\
		2& Willow swamp& 13 &	3 &227\\
		3&	CP hammock      &13     	 &3 &240 \\
		4&	Slash pine &13	&    3 &236\\
		5&	Oak/broadleaf  &9  &     2 &150\\
		6&	Hardwood	  &12       &  3    &214\\
		7&	Swamp	 &6        & 2       &97\\
		8&	Graminoid marsh &	20 &      4 &366\\
		9&	Spartina marsh &26    &6  &488\\
		10&	Cattail marsh &21	   &5 &378\\
		11&	Salt marsh   &21 &5      &393 \\
		12&	Mud flats  &26	    &6     & 471  \\
		13&	Water	    &47        & 10 &870 \\
		\hline
		\multicolumn{2}{c}{Total}  &245 &56	&4455  \\	
		\hline  \hline
	\end{tabular}
\label{tabDataKSC}
\end{table}

(2) \emph{Pavia University}: The second dataset is Pavia University (PU), which is an urban image captured over Northern Italy by Reflective Optics System Imaging Spectrometer (ROSIS) in 2002. The whole image has 9 classes with 610$\times$340 pixels. The image contains 115 spectral bands ranging from 430 to 860nm. The 12 noisy bands are removed. For this dataset, 1\%, 1\% and 98\% of samples from each class are randomly selected for training, validation and testing, respectively, as listed in Table \ref{tabDataPU}.

\begin{table*}[!t]
	\centering
	\caption{Classification accuracy (\%) and $\kappa$ coefficient for IP dataset. }
	\setlength{\tabcolsep}{1.5mm} {
		\begin{tabular}{c|cccccccccc}
			\hline \hline
			\textbf{class}  &\textbf{2D-CNN} \cite{makantasisDeepSupervisedLearning2015}	&\textbf{3D-CNN} \cite{liSpectralSpatialClassification2017b} &	\textbf{SSRN}  \cite{zhongSpectralSpatialResidual2018b}&	\textbf{PyraCNN}  \cite{paolettiDeepPyramidalResidual2019} & \textbf{HybridSN} \cite{royHybridSNExploring3D2020} &	\textbf{FCN-CRF} \cite{xuPatchwiseClassificationSpectralSpatial2019}& \textbf{Original non-local} &\textbf{ENL-FCN}  \\
			\hline
1	&	$100.00 \pm	0.00$ 	&	$99.41 	\pm	1.17$ 	&	$98.18 	\pm	3.63$ 	&	$97.26 	\pm	1.56 $	&$	97.16	\pm	2.73	$
&	$96.76 	\pm	3.25$ 	&	$97.56 	\pm	2.44 $	&	$97.15 	\pm	1.00$	\\
2	&	$94.11 	\pm	0.69$ 	&	$96.07 	\pm	2.27$ 	&	$96.25 	\pm	0.80$ 	&	$98.99 	\pm	0.56$ 	&$	97.15	\pm	0.56	$
&	$95.46 	\pm	1.28$ 	&	$97.81 	\pm	0.31$ 	&	$97.86 	\pm	0.41$ 	\\
3	&	$93.69 	\pm	1.52$ 	&	$94.65 	\pm	1.75$ 	&	$96.84 	\pm	1.02$ 	&	$98.96 	\pm	0.64$ 	&$	98.25	\pm	0.65	$
&	$94.78 	\pm	2.56$	&	$99.67 	\pm	0.30$ 	&	$99.75 	\pm	0.18$ 	\\
4	&	$95.40 	\pm	4.35$ 	&	$97.65 	\pm	1.97$ 	&	$97.16 	\pm	2.46$ 	&	$95.54 	\pm	2.14 $	&$	97.9	\pm	2.57	$
&	$90.40 	\pm	3.74$ 	&	$97.25 	\pm	0.52$ 	&	$96.60 	\pm	0.97$ 	\\
5	&	$96.87 	\pm	1.10 $	&	$98.76 	\pm	1.09$ 	&	$99.03 	\pm	0.17$	&	$98.79 	\pm	1.03$ 	&$	98.49	\pm	0.96	$
&  $94.24\pm	3.06$ 	&	$99.34 	\pm	0.26 $	&	$99.26 	\pm	0.50 $	\\
6	&	$98.35 	\pm	1.29$ 	&	$98.00 	\pm	1.34$ 	&	$98.61 	\pm	1.29$ 	&	$99.43 	\pm	0.39$ 	&$	98.92	\pm	0.38	$
&	$98.42 	\pm	1.12$ 	&	$99.11 	\pm	0.31$ 	&	$99.13 	\pm	0.25$ 	\\
7	&	$100.00 \pm	0.00 $	&	$98.82 	\pm	2.35$ 	&	$98.03 	\pm	2.41$ 	&	$89.00 	\pm	4.66 $	&$	100	\pm	0.00	$
&	$83.33 	\pm	13.44$ 	&	$100.00 	\pm	0.00 $	&	$100.00 	\pm	0.00$ 	\\
8	&	$96.58 	\pm	1.02$ 	&	$99.09 	\pm	0.97 $	&	$99.45 	\pm	0.68$ 	&	$100.00 	\pm	0.00 $&$	99.67	\pm	0.31	$
	&	$99.56 	\pm	0.59$ 	&	$99.86 	\pm	0.13 $	&	$99.84 	\pm	0.12$ 	\\
9	&	$100.00 	\pm	0.00$ 	&	$95.00 	\pm	6.12$ 	&	$97.64 	\pm	4.70$ 	&	$91.67 	\pm	6.64$ &$	92.38	\pm	5.25	$
	&	$88.54 	\pm	15.01$ 	&	$88.89 	\pm	3.93$ 	&	$85.19 	\pm	2.87$ 	\\
10	&	$94.27 	\pm	2.43$ 	&	$96.18 	\pm	0.91$ 	&	$95.21 	\pm	1.72$ 	&	$95.37 	\pm	1.16$ 	&$	98.74	\pm	0.81	$
&	$95.88 	\pm	1.27$ 	&	$97.84 	\pm	0.54$ 	&	$98.22 	\pm	0.33$ 	\\
11	&	$95.81 	\pm	1.23$ 	&	$96.08 	\pm	1.16$ 	&	$96.68 	\pm	0.92$	&	$98.98 	\pm	0.54$ 	&$	99.16	\pm	0.26	$
&	$98.22 	\pm	1.27$ 	&	$99.84 	\pm	0.03$ 	&	$99.82 	\pm	0.03$ 	\\
12	&	$93.74 	\pm	3.45$ 	&	$97.02 	\pm	1.36 $	&	$96.04 	\pm	1.43$ 	&	$95.23 	\pm	1.83$ 	&$	97.47	\pm	1.17	$
&	$92.46 	\pm	4.50$ 	&	$99.51 	\pm	0.17 $	&	$99.40 	\pm	0.33$ 	\\
13	&	$99.68 	\pm	0.63$ 	&	$99.78 	\pm	0.26$ 	&	$99.58 	\pm	0.51$ 	&	$100.00 	\pm	0.00$ &$	98.02	\pm	1.64	$
	&	$98.57 	\pm	1.72 $	&	$98.69 	\pm	0.30$ 	&	$98.91 	\pm	0.00$ 	\\
14	&	$98.39 	\pm	0.93$ 	&	$98.82 	\pm	0.59$ 	&	$99.30 	\pm	0.36$	&	$98.34 	\pm	0.77 $	&$	99.32	\pm	0.44	$
&	$98.36 	\pm	1.02$ 	&	$99.88 	\pm	0.10$ 	&	$99.91 	\pm	0.08$ 	\\
15	&	$95.67 	\pm	1.98$ 	&	$94.72 	\pm	2.13$ 	&	$95.71 	\pm	2.18$ 	&	$94.60 	\pm	1.46$ 	&$	97.64	\pm	1.58	$
&	$93.71 	\pm	3.57$ 	&	$92.12 	\pm	0.80$ 	&	$93.00 	\pm	0.87$ 	\\
16	&	$94.48 	\pm	4.08$ 	&	$95.48 	\pm	2.97$ 	&	$95.48 	\pm	2.20 $	&	$96.43 	\pm	2.77$ 	&$	91.02	\pm	4.07	$
&	$96.44 	\pm	2.48 $	&	$93.98 	\pm	1.70$ 	&	$93.57 	\pm	0.62$	\\
\hline
OA	&	$95.79 	\pm	0.44$ 	&	$96.84 	\pm	0.84 $	&	$97.19 	\pm	0.45$ 	&	$ 98.12 	\pm	0.24$ &$	98.44	\pm	0.16	$
	&	$96.48 	\pm	1.03$ 	&	$98.80 	\pm	0.06$ 	&	$\textbf{98.85} \pm	0.07$ 	\\
AA	&	$96.69 	\pm	0.40$ 	&	$97.22 	\pm	0.57$ 	&	$97.45 	\pm	0.53$ 	&	$96.79 	\pm	0.36$ 	&$	97.58	\pm	0.82	$
&	$94.80 	\pm	1.17$ 	&	$\textbf{97.58} \pm	0.19$ 	&	$97.35	\pm	0.15$ 	\\
Kappa	&	$95.10 	\pm	0.50$ 	&	$96.40 	\pm	0.95 $	&	$96.80 	\pm	0.52$ 	&	$97.85 	\pm	0.28$ &$	98.23	\pm	0.18	$
	&	$95.98 	\pm	1.18$ 	&	$98.64 	\pm	0.07$ 	&	$\textbf{98.69}\pm0.08$	\\
\hline	\hline	
	\end{tabular}}
	\label{tabIP}
\end{table*}

\begin{table*}[!t]
	\centering
	\caption{Classification accuracy (\%) and $\kappa$ coefficient for PU dataset. }
	\setlength{\tabcolsep}{1.5mm} {
		\begin{tabular}{c|cccccccccc}
		\hline	\hline	
			\textbf{class}  &\textbf{2D-CNN} \cite{makantasisDeepSupervisedLearning2015}	&\textbf{3D-CNN} \cite{liSpectralSpatialClassification2017b} &	\textbf{SSRN}  \cite{zhongSpectralSpatialResidual2018b}&	\textbf{PyraCNN}  \cite{paolettiDeepPyramidalResidual2019} & \textbf{HybridSN} \cite{royHybridSNExploring3D2020} &	\textbf{FCN-CRF} \cite{xuPatchwiseClassificationSpectralSpatial2019}& \textbf{Original non-local} &\textbf{ENL-FCN}  \\
			\hline
1	&	$92.72 	\pm	1.02$ 	&	$87.21 	\pm	3.39$ 	&	$99.66 	\pm	0.24$ 	&	$94.94 	\pm	1.37$ 	&$	95.13 	\pm	1.81 	$
&	$91.89 	\pm	2.39$ 	&	$98.41 	\pm	0.72$ 	&	$99.40 	\pm	0.45$ 	\\
2	&	$97.14 	\pm	0.59$ 	&	$94.10 	\pm	2.04$ 	&	$98.70 	\pm	1.02$ 	&	$99.41 	\pm	0.40$ 	&$	99.16 	\pm	0.49 	$
&	$95.83 	\pm	0.55 $	&	$99.95 	\pm	0.05$ 	&	$100.00 	\pm	0.01$ 	\\
3	&	$87.91 	\pm	3.40$ 	&	$64.08 	\pm	6.46$ 	&	$93.95 	\pm	5.15$ 	&	$81.90 	\pm	5.09$ 	&$	88.73 	\pm	4.90 	$
&	$95.82 	\pm	0.53$ 	&	$89.49 	\pm	3.66$ 	&	$91.45 	\pm	1.96$ 	\\
4	&	$99.35 	\pm	0.90$ 	&	$96.82 	\pm	1.76 $	&	$99.72 	\pm	0.27$ 	&	$93.75 	\pm	1.84$ 	&$	98.18 	\pm	0.77 	$
&	$98.23 	\pm	0.73$ 	&	$97.05 	\pm	0.45$ 	&	$97.55 	\pm	0.66$ 	\\
5	&	$98.92 	\pm	1.61$ 	&	$95.13 	\pm	4.87$ 	&	$99.93 	\pm	0.08$ 	&	$99.78 	\pm	0.29$ 	&$	98.98 	\pm	0.93 	$
&	$99.67 	\pm	0.35$ 	&	$99.94 	\pm	0.14$ 	&	$100.00 	\pm	0.00$ 	\\
6	&	$97.41 	\pm	0.77$ 	&	$94.07 	\pm	1.98$ 	&	$98.52 	\pm	2.11$ 	&	$93.91 	\pm	2.55$ 	&$	98.66 	\pm	0.96 	$
&	$94.76 	\pm	1.64$ 	&	$99.51 	\pm	0.61$ 	&	$99.28 	\pm	0.65$ 	\\
7	&	$91.99 	\pm	4.80$ 	&	$58.80 	\pm	5.37$ 	&	$96.84 	\pm	2.22 $	&	$83.03 	\pm	3.46$ 	&$	96.64 	\pm	2.37 	$
&	$95.42 	\pm	0.90$ 	&	$97.53 	\pm	1.76$ 	&	$98.66 	\pm	1.32 $	\\
8	&	$88.41 	\pm	1.07$ 	&	$77.11 	\pm	2.78$ 	&	$88.85 	\pm	5.74$ 	&	$89.20 	\pm	3.22$ 	&$	90.69 	\pm	2.72 	$
&	$94.95 	\pm	1.09$ 	&	$98.64 	\pm	0.78$ 	&	$99.26 	\pm	0.33$ 	\\
9	&	$99.41 	\pm	0.59$ 	&	$84.19 	\pm	4.87$ 	&	$99.53 	\pm	0.55$ 	&	$99.84 	\pm	0.18 $	&$	97.21 	\pm	1.86 	$
&	$99.77 	\pm	0.17$ 	&	$98.45 	\pm	0.78$ 	&	$98.24 	\pm	0.87$ 	\\
\hline
OA	&	$95.35 	\pm	0.17 $	&	$88.69 	\pm	1.99$	&	$97.54 	\pm	0.58$ 	&	$95.44 	\pm	0.24$ 	&$	97.01 	\pm	0.69 	$
&	$95.36 	\pm	0.33$ 	&	$98.72 	\pm	0.12$ 	&	$\textbf{99.08}\pm	0.12$ 	\\
AA	&	$94.81 	\pm	0.56$ 	&	$83.50 	\pm	2.14$ 	&	$97.30 	\pm	0.55$ 	&	$92.86 	\pm	0.69$ 	&$	95.93 	\pm	0.87 	$
&	$96.26 	\pm	0.35$ 	&	$97.66 	\pm	0.41$ 	&	$\textbf{98.20}	\pm	0.34 $	\\
Kappa	&	$93.81 	\pm	0.23$ 	&	$84.86 	\pm	2.72$ 	&	$96.74 	\pm	0.77$ 	&	$93.93 	\pm	0.33 $&$	96.02 	\pm	0.92 	$
	&	$93.83 	\pm	0.43$ 	&	$98.30 	\pm	0.16$ 	&	$\textbf{98.78}	\pm	0.15$ 	\\
\hline	\hline		
	\end{tabular}}
	\label{tabPU}
\end{table*}

\begin{table*}[!t]
	\centering
	\caption{Classification accuracy (\%) and $\kappa$ coefficient for KSC dataset. }
	\setlength{\tabcolsep}{1.5mm} {
		\begin{tabular}{c|cccccccccc}
		\hline	\hline	
			\textbf{class}  &\textbf{2D-CNN} \cite{makantasisDeepSupervisedLearning2015}	&\textbf{3D-CNN} \cite{liSpectralSpatialClassification2017b} &	\textbf{SSRN}  \cite{zhongSpectralSpatialResidual2018b}&	\textbf{PyraCNN}  \cite{paolettiDeepPyramidalResidual2019} & \textbf{HybridSN} \cite{royHybridSNExploring3D2020} &	\textbf{FCN-CRF} \cite{xuPatchwiseClassificationSpectralSpatial2019}& \textbf{Original non-local} &\textbf{ENL-FCN}  \\
			\hline
1	&	$97.65 	\pm	1.05$ 	&	$98.58 	\pm	0.17$ 	&	$98.84 	\pm	0.94$ 	&	$98.93 	\pm	0.86$ 	&$	98.47 	\pm	0.51 	$
&	$100.00 	\pm	0.02$ 	&	$98.68 	\pm	0.29$ 	&	$99.62 	\pm	0.29$ 	\\
2	&	$88.78 	\pm	4.42$ 	&	$82.31 	\pm	5.79$ 	&	$97.07 	\pm	5.63$ 	&	$92.54 	\pm	2.20$ 	&$	93.75 	\pm	5.50 	$
&	$99.34 	\pm	0.93$ 	&	$100.00 	\pm	0.00$ 	&	$100.00 	\pm	0.00$ 	\\
3	&	$64.38 	\pm	11.24$ 	&	$73.55 	\pm	8.35$ 	&	$97.13 	\pm	1.27$ 	&	$95.12 	\pm	3.00 $	&$	89.08 	\pm	2.36 	$
&	$99.17 	\pm	0.59$ 	&	$97.93 	\pm	1.76$ 	&	$99.59 	\pm	0.00$ 	\\
4	&	$67.29 	\pm	7.56$ 	&	$60.33 	\pm	8.67$ 	&	$89.77 	\pm	4.52$ 	&	$81.55 	\pm	4.10$ 	&$	89.93 	\pm	3.01 	$
&	$71.94 	\pm	15.22$ 	&	$99.37 	\pm	0.42$ 	&	$98.10 	\pm	1.22$ 	\\
5	&	$65.42 	\pm	14.22$ 	&	$64.64 	\pm	11.99$ 	&	$87.78 	\pm	12.10$ 	&	$81.30 	\pm	4.71$ 	&$	90.48 	\pm	6.70 	$
&	$68.33 	\pm	12.73$ 	&	$99.83 	\pm	0.33$ 	&	$100.00 	\pm	0.00$ 	\\
6	&	$79.15 	\pm	4.40 $	&	$79.29 	\pm	2.96$ 	&	$99.32 	\pm	0.91$ 	&	$89.86 	\pm	3.97$ 	&$	92.98 	\pm	6.09 	$
&	$99.30 	\pm	0.33$ 	&	$100.00 	\pm	0.00$ 	&	$100.00 	\pm	0.00$ 	\\
7	&	$80.14 	\pm	3.83$ 	&	$77.98 	\pm	11.43$ 	&	$93.55 	\pm	5.53$ 	&	$95.27 	\pm	5.30$ 	&$	91.07 	\pm	6.22 	$
&	$33.16 	\pm	36.80$ 	&	$99.74 	\pm	0.51$ 	&	$100.00 	\pm	0.00$ 	\\
8	&	$88.49 	\pm	8.07$ 	&	$93.37 	\pm	3.28$ 	&	$98.59 	\pm	1.23$ 	&	$99.28 	\pm	0.56$ 	&$	93.59 	\pm	1.90 	$
&	$99.01 	\pm	0.35$ 	&	$99.81 	\pm	0.37$ 	&	$100.00 	\pm	0.00 $	\\
9	&	$90.87 	\pm	6.95$ 	&	$88.06 	\pm	3.97$ 	&	$98.50 	\pm	1.90$ 	&	$99.76 	\pm	0.42$ 	&$	94.12 	\pm	2.96 	$
&	$100.00 	\pm	0.00$ 	&	$100.00 	\pm	0.00 $	&	$100.00 	\pm	0.00 $	\\
10	&	$99.52 	\pm	0.58$ 	&	$98.36 	\pm	1.18$ 	&	$99.32 	\pm	1.10$ 	&	$99.09 	\pm	0.60$ 	&$	98.36 	\pm	2.76 	$
&	$100.00 	\pm	0.00$ 	&	$99.20 	\pm	0.69$ 	&	$99.93 	\pm	0.13 $	\\
11	&	$99.64 	\pm	0.44$ 	&	$99.69 	\pm	0.49$ 	&	$99.74 	\pm	0.50$ 	&	$99.97 	\pm	0.09$ 	&$	97.69 	\pm	2.71 	$
&	$100.00 	\pm	0.00$ 	&	$100.00 	\pm	0.00$ 	&	$100.00 	\pm	0.00$ 	\\
12	&	$97.98 	\pm	1.04$ 	&	$89.44 	\pm	3.08 $	&	$98.16 	\pm	2.27$ 	&	$99.46 	\pm	0.66$ 	&$	99.19 	\pm	1.19 	$
&	$99.36 	\pm	0.90$ 	&	$95.44 	\pm	0.27$ 	&	$96.18 	\pm	0.00$ 	\\
13	&	$98.98 	\pm	1.18$ 	&	$98.56 	\pm	1.16$ 	&	$100.00 	\pm	0.00$ 	&	$100.00 	\pm	0.00$ &$	99.47 	\pm	0.58 	$
	&	$100.00 	\pm	0.00$ 	&	$100.00 	\pm	0.00$ 	&	$100.00 	\pm	0.00$ 	\\
\hline
OA	&	$91.06 	\pm	2.14$ 	&	$89.98 	\pm	1.52$ 	&	$97.88 	\pm	0.80 $	&	$97.04 	\pm	0.31$ 	&$	96.04 	\pm	1.16 	$
&	$96.08 	\pm	0.33$ 	&	$99.15 	\pm	0.20 $	&	$\textbf{99.46} 	\pm	0.10$	\\
AA	&	$86.02 	\pm	2.47$ 	&	$84.93 	\pm	2.46 $	&	$96.75 	\pm	1.30 $	&	$94.78 	\pm	0.75 $	&$	94.47 	\pm	1.94 	$
&	$89.97 	\pm	2.63$ 	&	$99.23 	\pm	0.31 $	&	$\textbf{99.49} 	\pm	0.12$ 	\\
Kappa	&	$90.04 	\pm	2.39$ 	&	$88.84 	\pm	1.70 $	&	$97.64 	\pm	0.89$ 	&	$96.70 	\pm	0.34$ &$	95.59 	\pm	1.29 	$
	&	$95.64 	\pm	0.37$ 	&	$99.05 	\pm	0.23$ 	&	$\textbf{99.40} \pm	0.11$	\\
\hline	\hline		
	\end{tabular}}
	\label{tabKSC}
\end{table*}

(3) \emph{Kennedy Space Center}: The third dataset is Kennedy Space Center (KSC), which is collected by the AVIRIS sensor over Kennedy Space Center, Florida, in 1996. After removing water absorption and other noisy bands, the KSC dataset contains 176 spectral bands. The spatial resolution of KSC is 18m with  $512\times 614$ pixels. It includes 13 land cover classes are included in KSC dataset. For each class, 5\%, 1\% and 94\% of samples are randomly selected for training, validation and testing, respectively. The detailed numbers are listed in Table \ref{tabDataKSC}.

\subsection{Experimental Setup}
In the proposed framework, a five-layer FCN is used as the backbone and the efficient non-local module is embedded as a learning unit. To further boost the classification performance, two parallel efficient non-local modules are used in FCN as shown in Fig. \ref{figFlowchart}. The convolutional kernel number is set to 150 and the kernel size is $5\times 5$. In the efficient non-local module, the number of convolutional kernels is also set to 150. The Adam optimizer is used to train the network. The learning rate is set to 0.0005 and the weight decay is 0.0002. The batch size is 1. The number of learning iterations is 800. The network is implemented under PyTorch environment with an Intel i9-9920X CPU and an NVIDIA TITAN-V GPU. Our code is available at \url{https://github.com/ShaneShen/ENL-FCN}.

\begin{figure*}[!t]
	\centering
		\subfloat[ ]{\centering
		\includegraphics[width=0.13\linewidth]{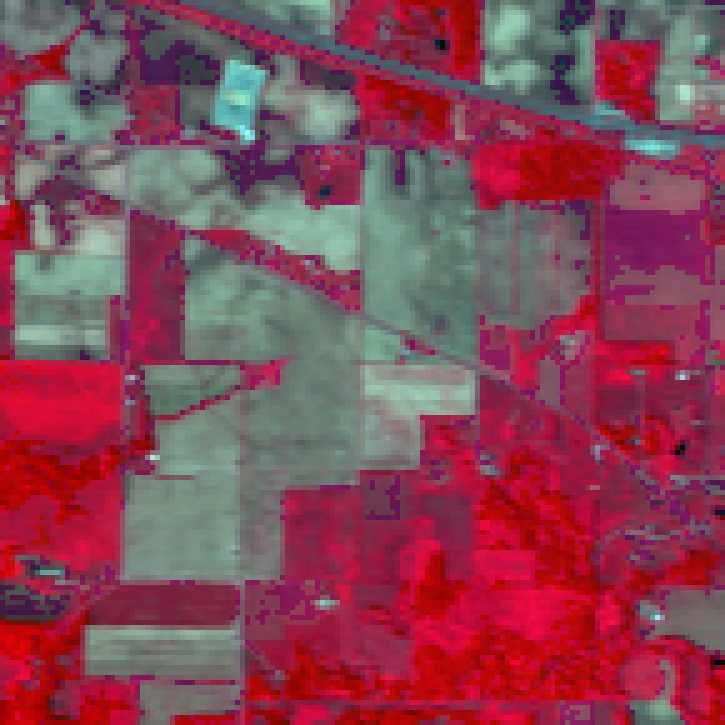}
		\label{fig:2}}    \hspace*{0.07em}
	\subfloat[ ]{\centering
		\includegraphics[width=0.13\linewidth]{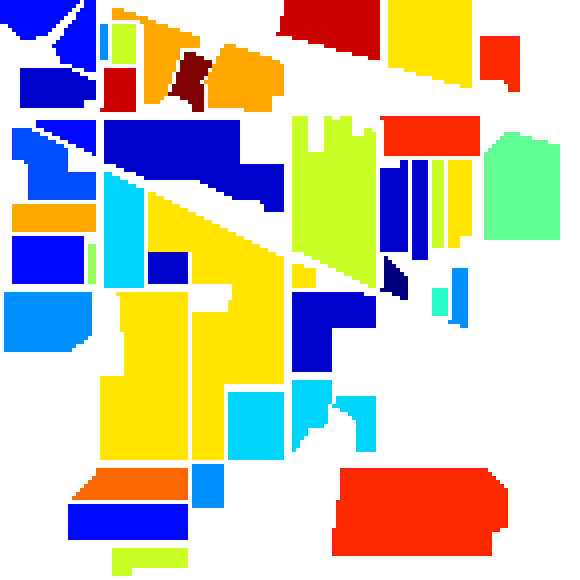}
		\label{fig:2}}    \hspace*{0.07em} 
	\subfloat[ ]{\centering
		\includegraphics[width=0.13\linewidth]{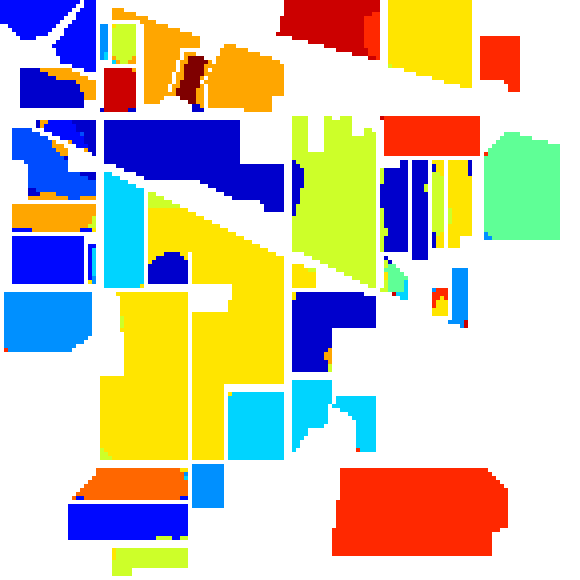}
		\label{fig:1} }  \hspace*{0.07em}    
	\subfloat[ ]{\centering
		\includegraphics[width=0.13\linewidth]{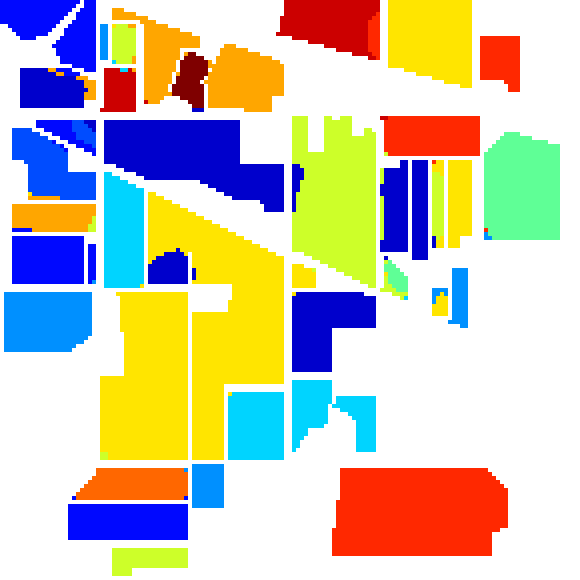}
		\label{fig:2}}    \hspace*{0.07em}      
	\subfloat[ ]{\centering
		\includegraphics[width=0.13\linewidth]{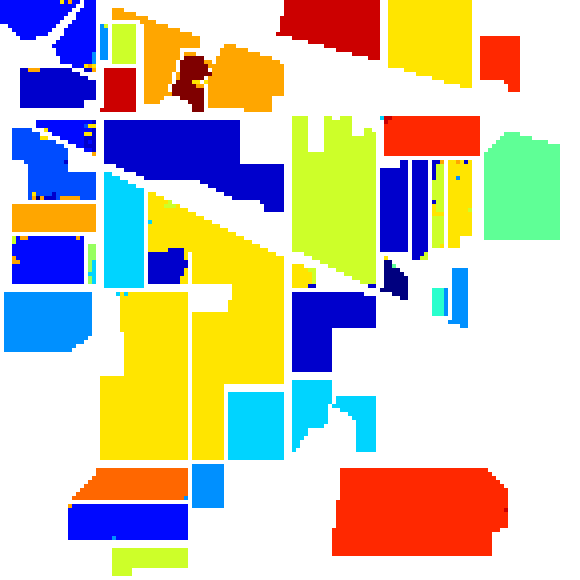}
		\label{fig:2}}    \hspace*{0.07em} \\
	\subfloat[ ]{\centering
	    \includegraphics[width=0.13\linewidth]{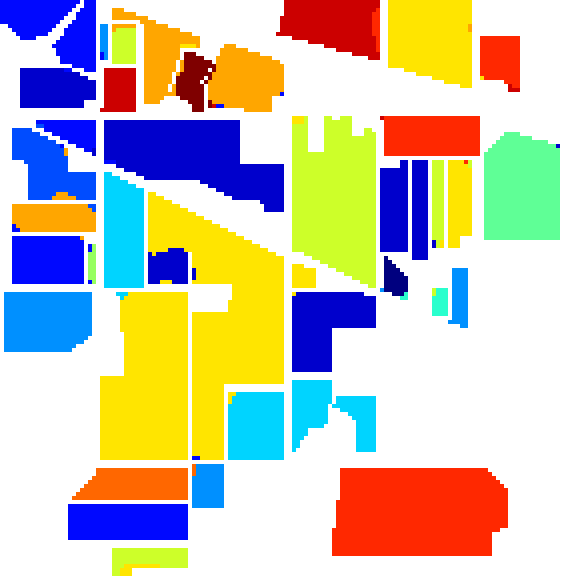}
	    \label{fig:1} }  \hspace*{0.07em}   
     \subfloat[ ]{\centering
	    \includegraphics[width=0.13\linewidth]{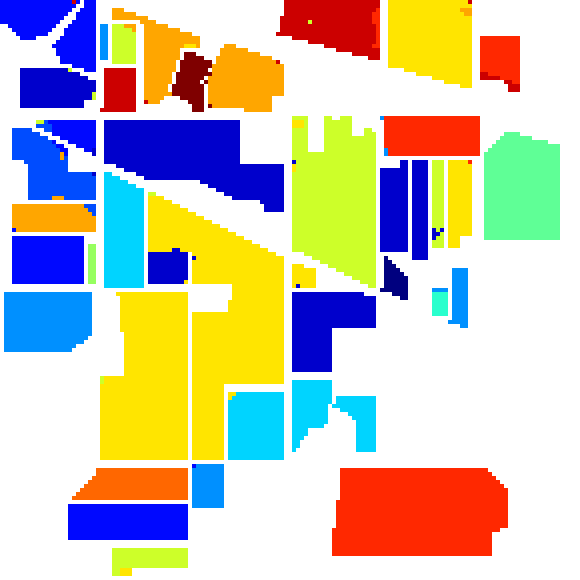}
	    \label{fig:1} }  \hspace*{0.07em}  
	\subfloat[ ]{\centering
	   \includegraphics[width=0.13\linewidth]{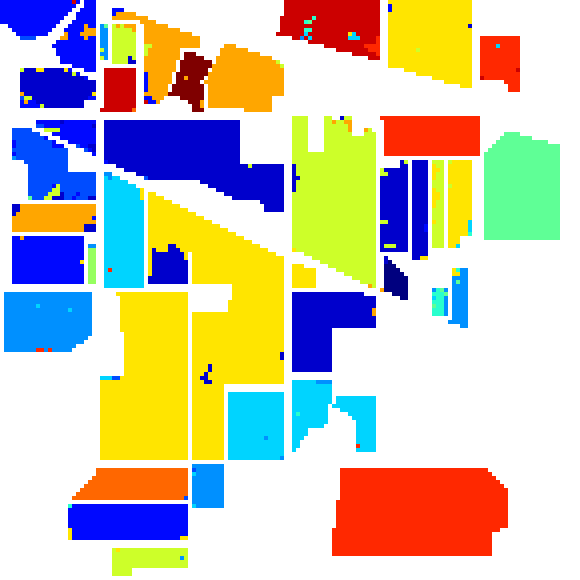}
	   \label{fig:1} }  \hspace*{0.07em}
	\subfloat[ ]{\centering
		\includegraphics[width=0.13\linewidth]{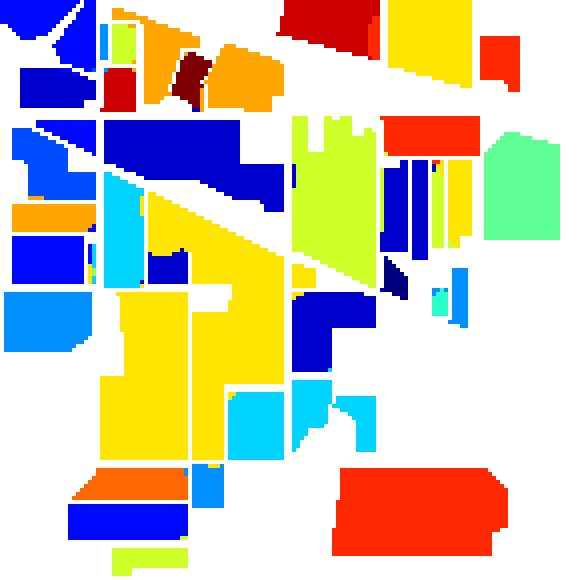}
		\label{fig:2}}    \hspace*{0.07em}    
	\subfloat[ ]{\centering
		\includegraphics[width=0.13\linewidth]{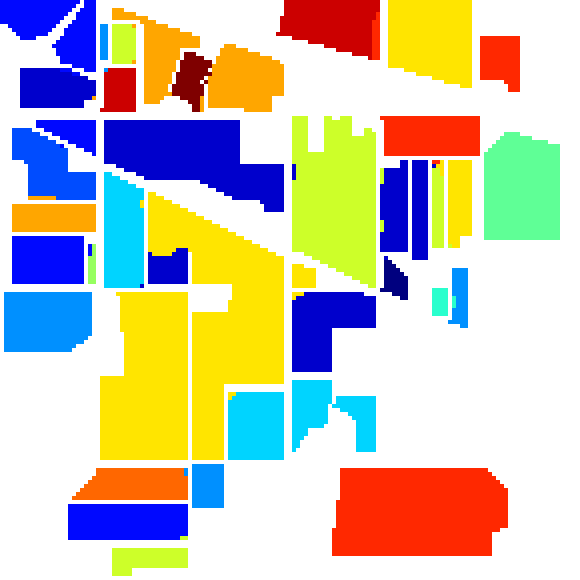}
		\label{fig:2}}    \hspace*{0.07em}    \\
		\subfloat{\centering
		\includegraphics[width=0.75\linewidth]{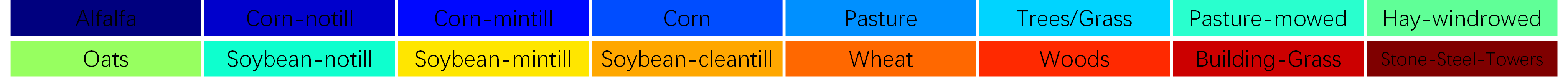}}  
	\caption{Classification results on IP dataset. (a) False color image, (b) Ground truth, (c) 2D-CNN (OA=95.79\%), (d) 3D-CNN (OA=96.84\%), (e) SSRN (OA=97.19\%), (f) PyraCNN (OA=98.12\%), (g) HybridSN (OA=98.44\%), (h) FCN-CRF (OA=96.48\%), (i) Original non-local (OA=98.80\%), and (j) the proposed ENL-FCN (OA=98.85\%).}
	\label{figIP}
\end{figure*}

\begin{figure*}[!t]
	\centering
		\subfloat[ ]{\centering
		\includegraphics[width=0.14\linewidth]{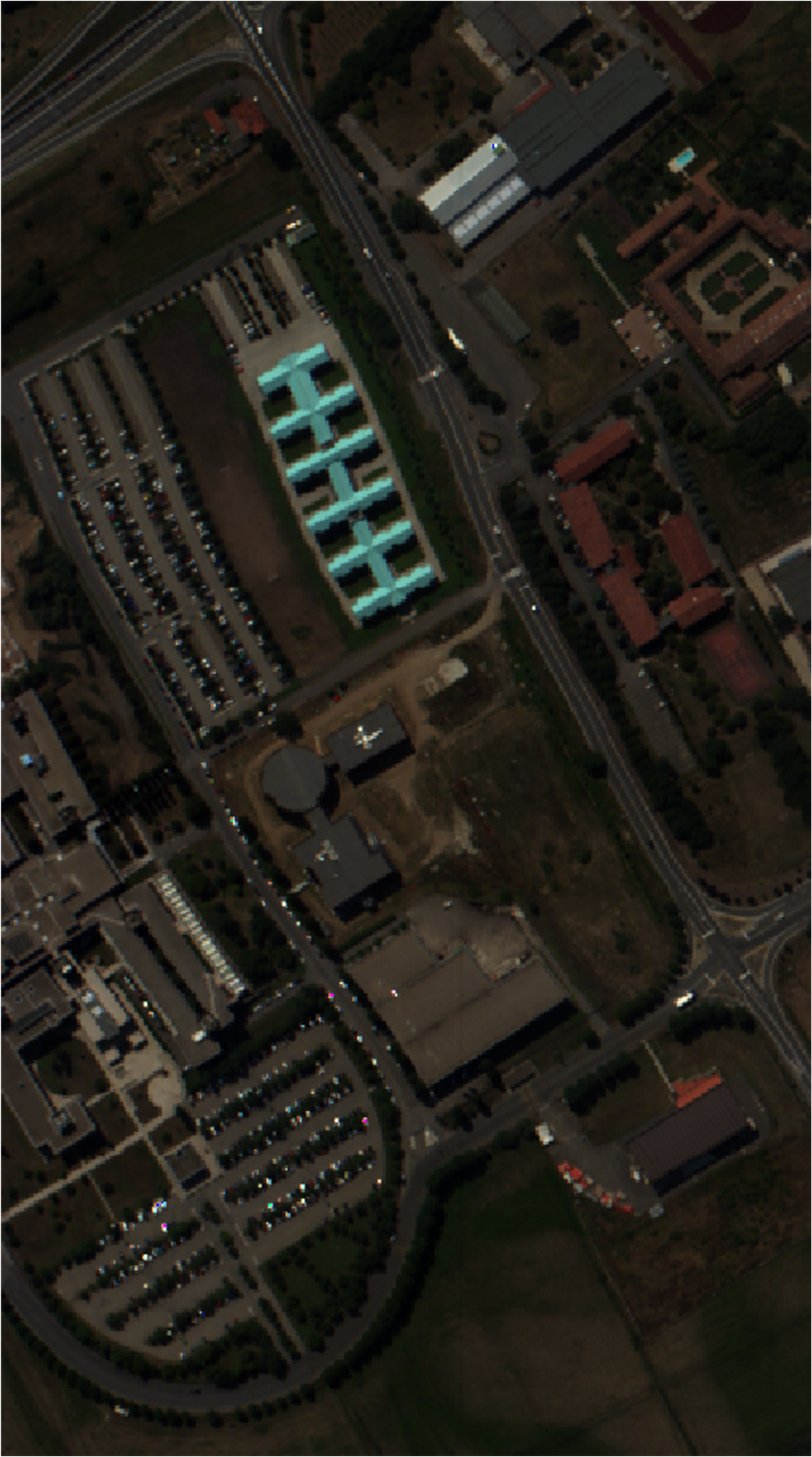}
		\label{fig:2}}    \hspace*{0.04em}
	\subfloat[ ]{\centering
		\includegraphics[width=0.14\linewidth]{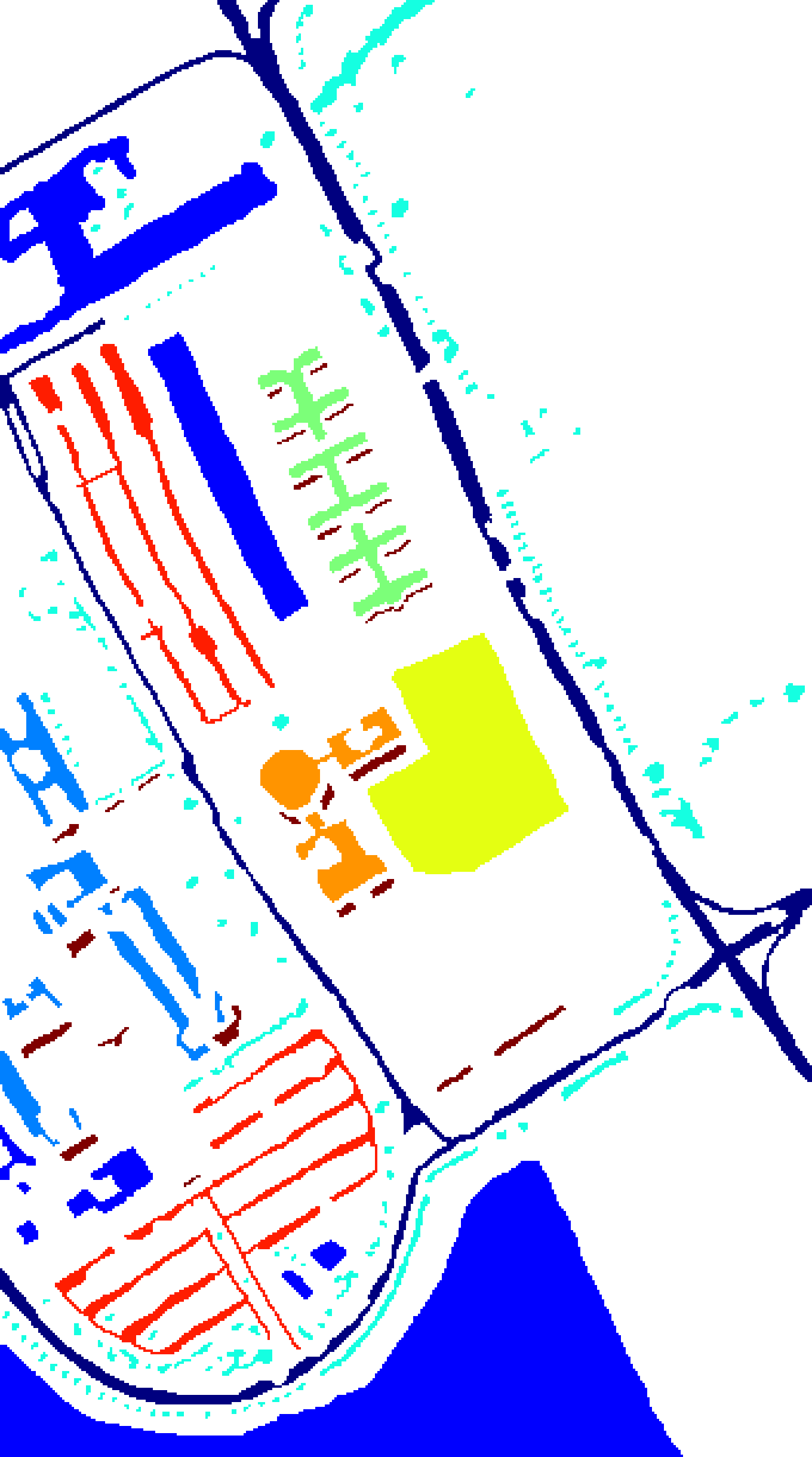}
		\label{fig:2}}    \hspace*{0.04em} 
	\subfloat[ ]{\centering
		\includegraphics[width=0.14\linewidth]{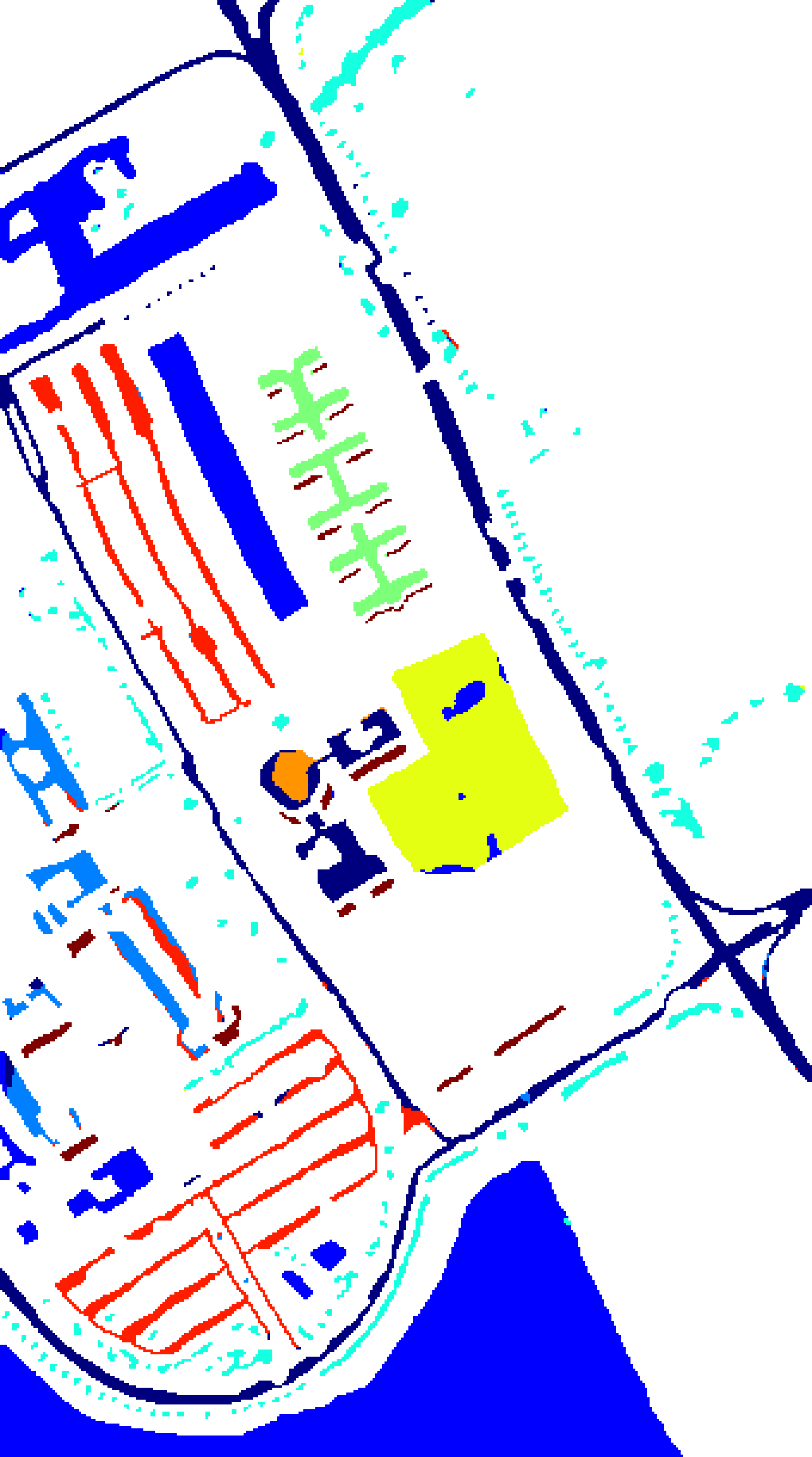}
		\label{fig:1} }  \hspace*{0.04em}    
	\subfloat[ ]{\centering
		\includegraphics[width=0.14\linewidth]{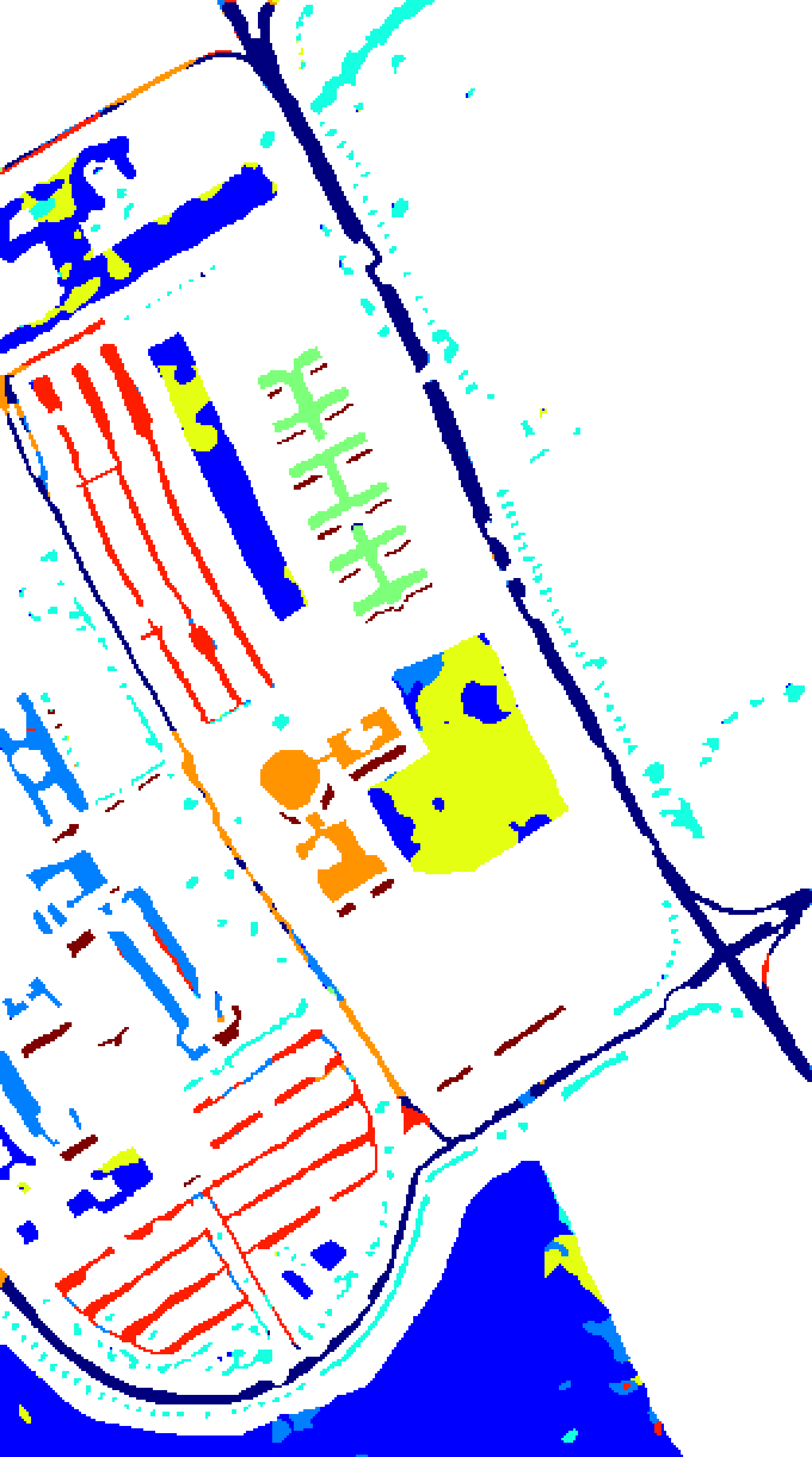}
		\label{fig:2}}    \hspace*{0.04em}
	\subfloat[ ]{\centering
		\includegraphics[width=0.14\linewidth]{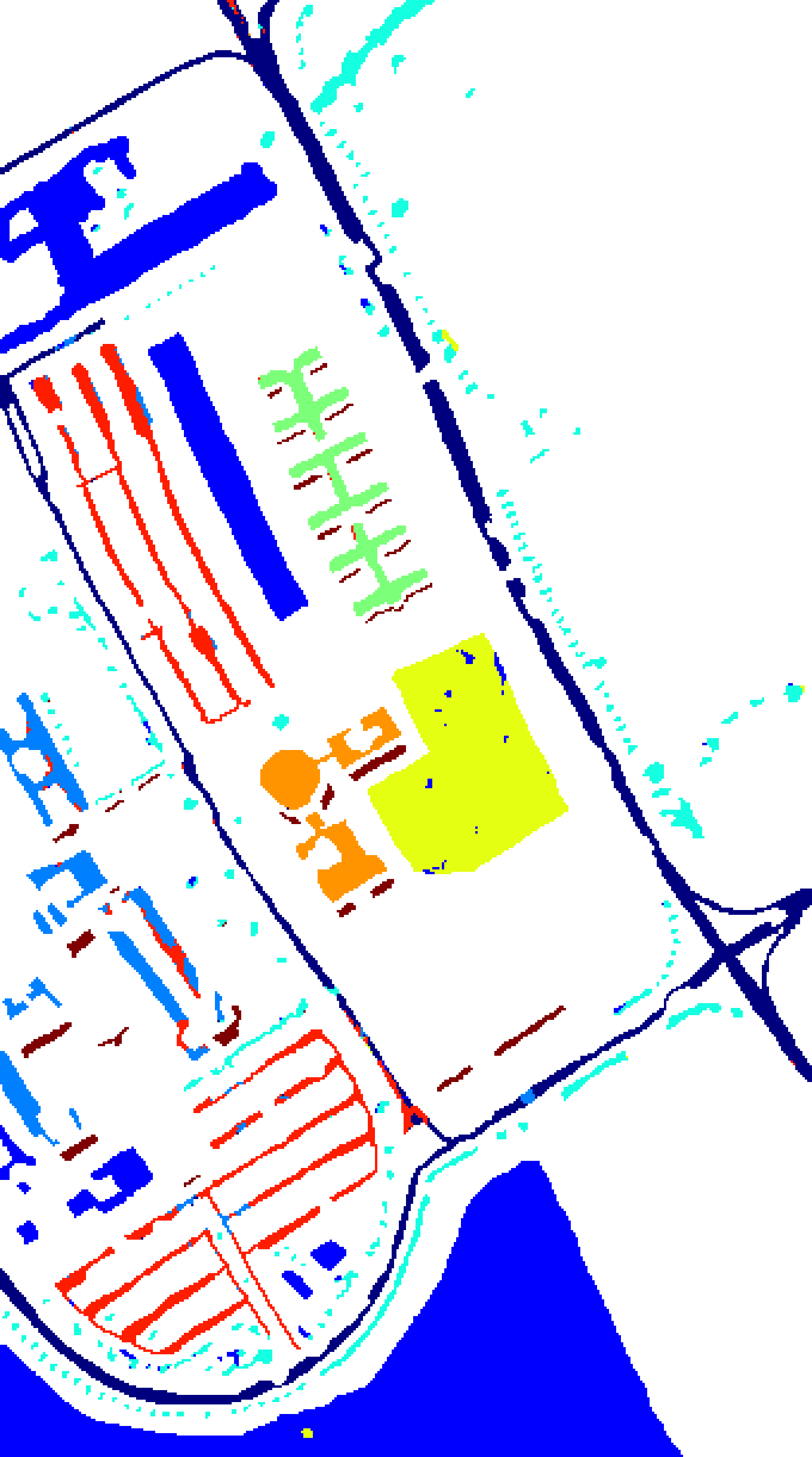}
		\label{fig:2}}    \hspace*{0.04em} \\
	\subfloat[ ]{\centering
	    \includegraphics[width=0.14\linewidth]{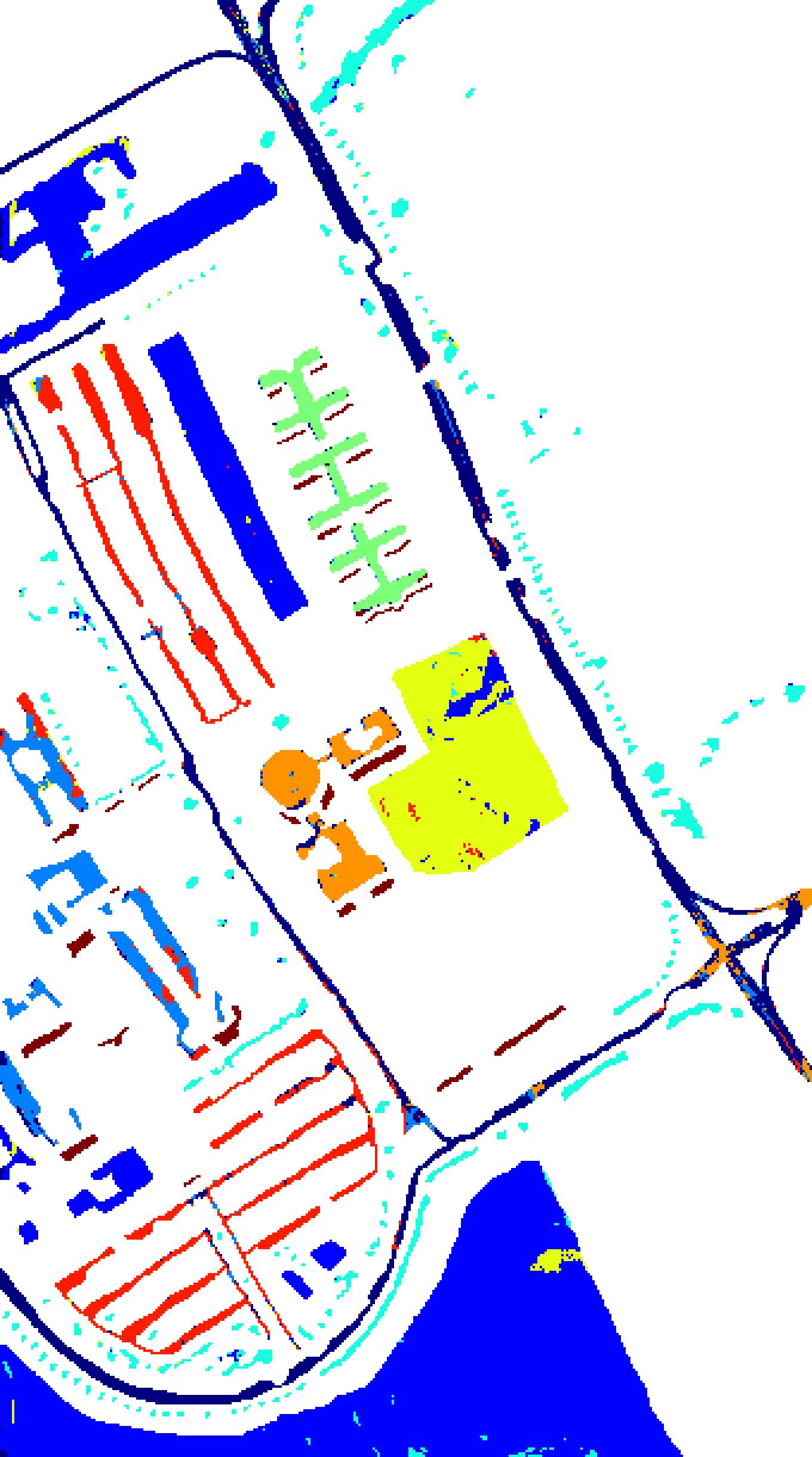}
	    \label{fig:1} }  \hspace*{0.04em}   
	\subfloat[ ]{\centering
	    \includegraphics[width=0.14\linewidth]{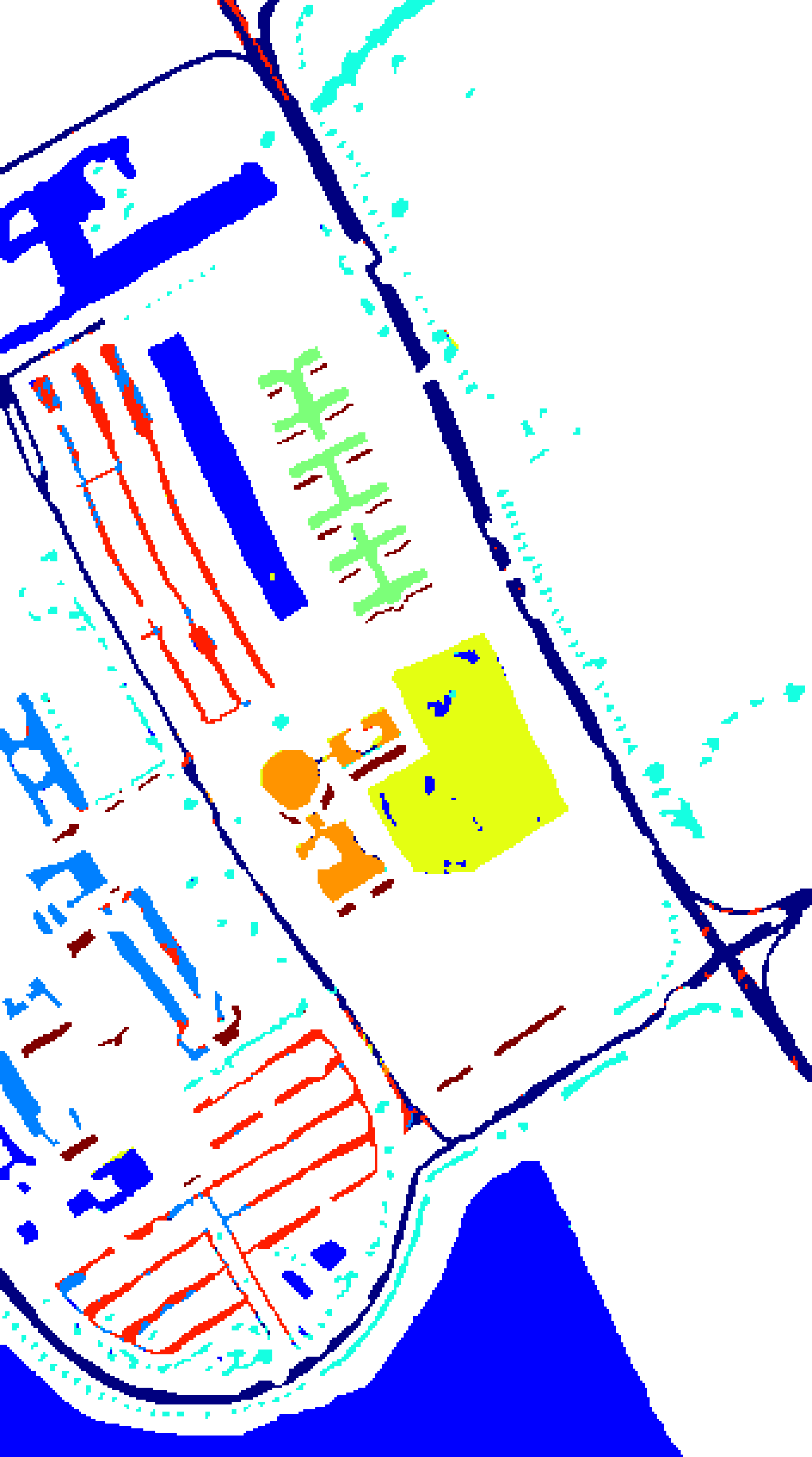}
	    \label{fig:1} }  \hspace*{0.04em}   
	\subfloat[ ]{\centering
	   \includegraphics[width=0.14\linewidth]{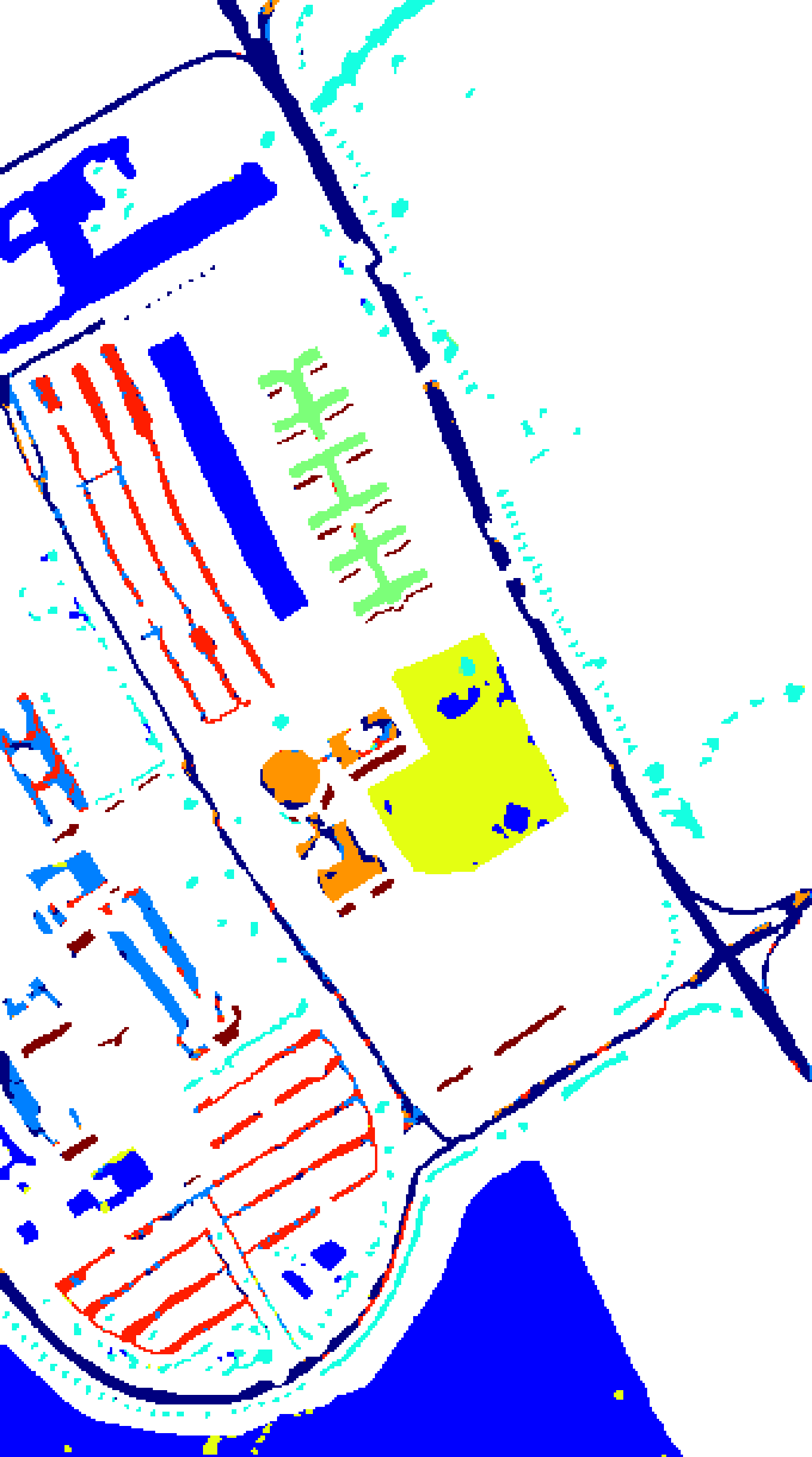}
	   \label{fig:1} }  \hspace*{0.04em}
	\subfloat[ ]{\centering
		\includegraphics[width=0.14\linewidth]{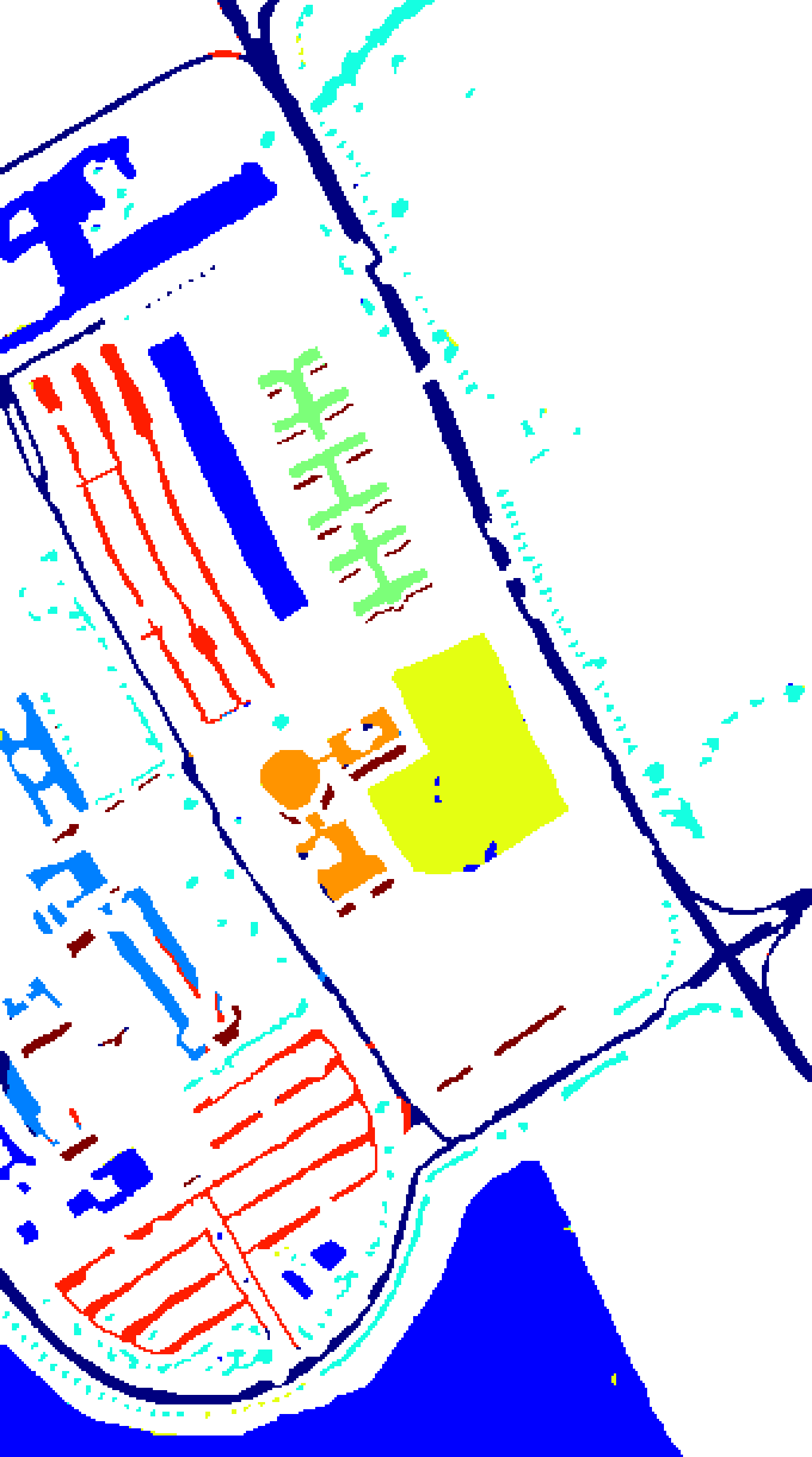}
		\label{fig:2}}    \hspace*{0.04em}    
	\subfloat[ ]{\centering
		\includegraphics[width=0.14\linewidth]{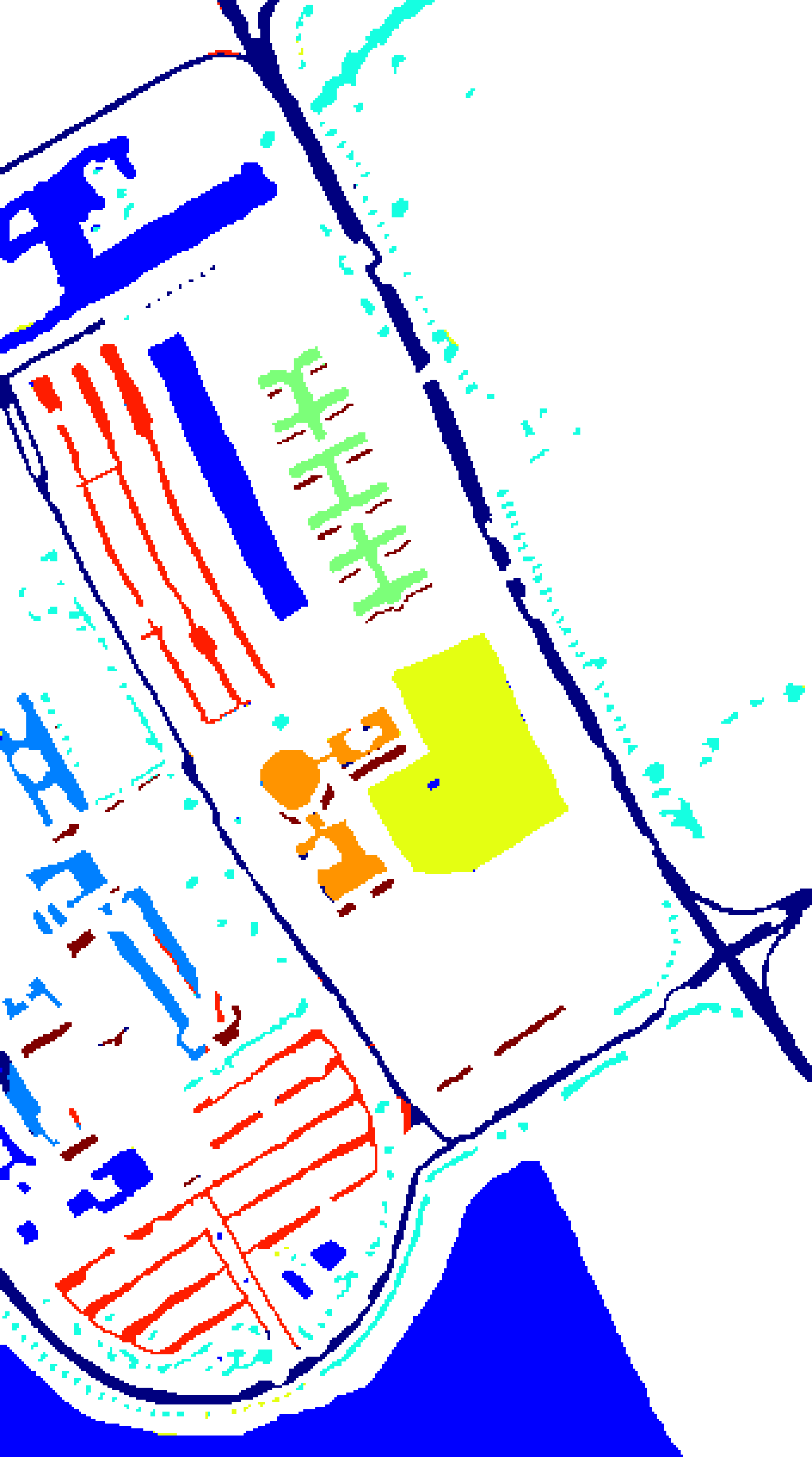}
		\label{fig:2}}    \hspace*{0.04em}     \\
	\subfloat{\centering
		\includegraphics[width=0.51\linewidth]{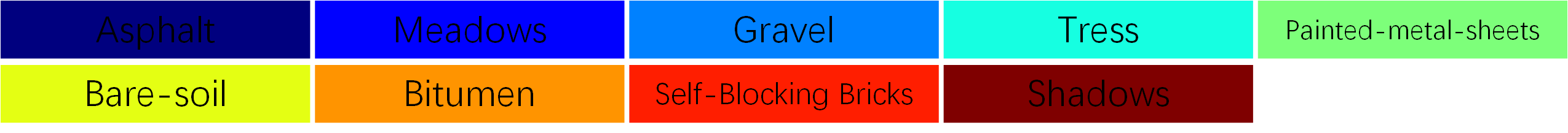}}  
	\caption{Classification results on PU dataset. (a) False color image, (b) Ground truth, (c) 2D-CNN (OA=95.35\%), (d) 3D-CNN (OA=88.69\%), (e) SSRN (OA=97.54\%), (f) PyraCNN (OA=95.44\%), (g) HybridSN (OA=97.01\%), (h) FCN-CRF (OA=95.36\%), (i) Original non-local (OA=98.72\%), and (j) the proposed ENL-FCN(OA=99.08\%).}
	\label{figPU}
\end{figure*}

\begin{figure*}[!t]
	\centering
		\subfloat[ ]{\centering
		\includegraphics[width=0.18\linewidth]{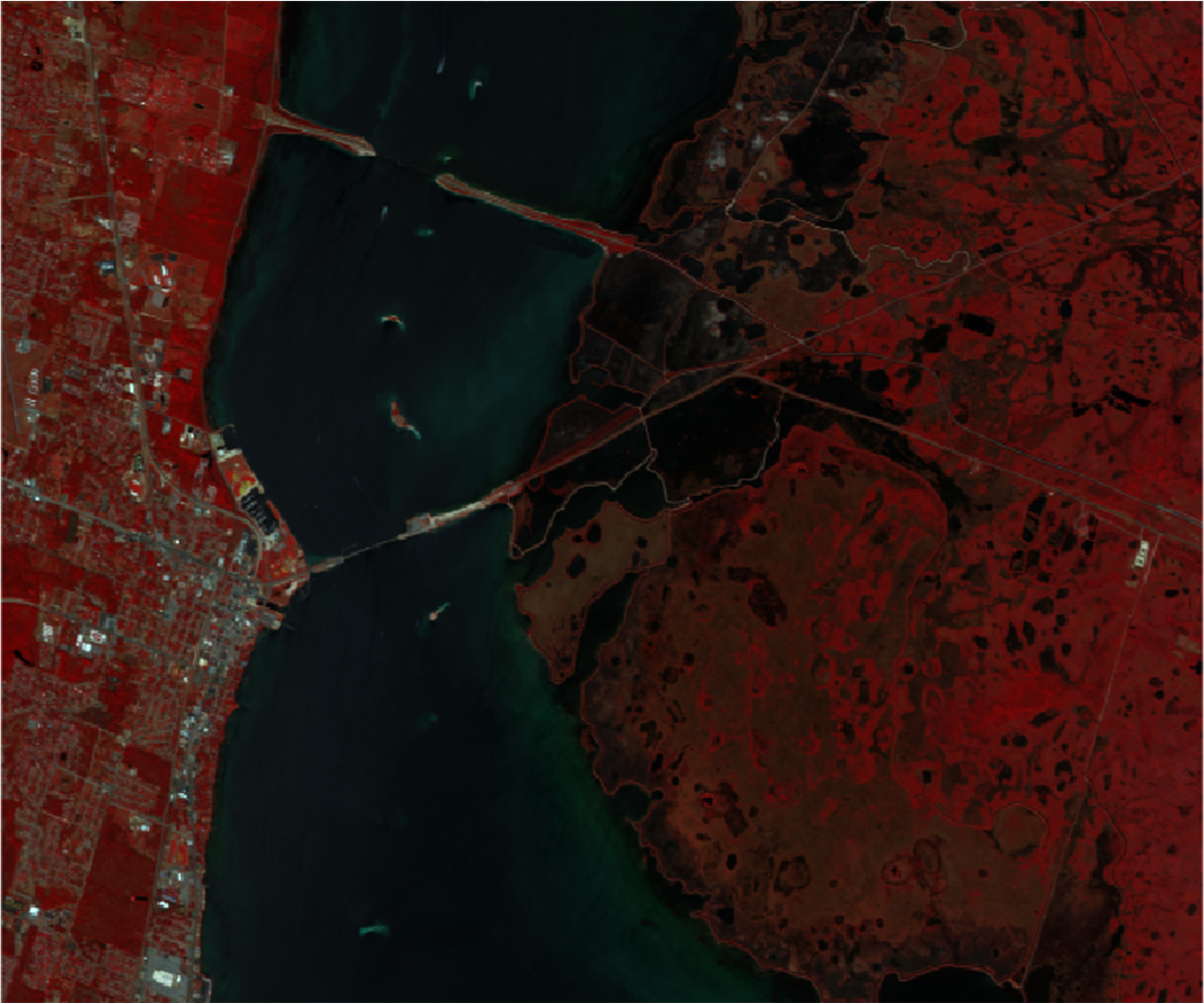}
		\label{fig:2}}    \hspace*{0.07em}
	\subfloat[ ]{\centering
		\includegraphics[width=0.18\linewidth]{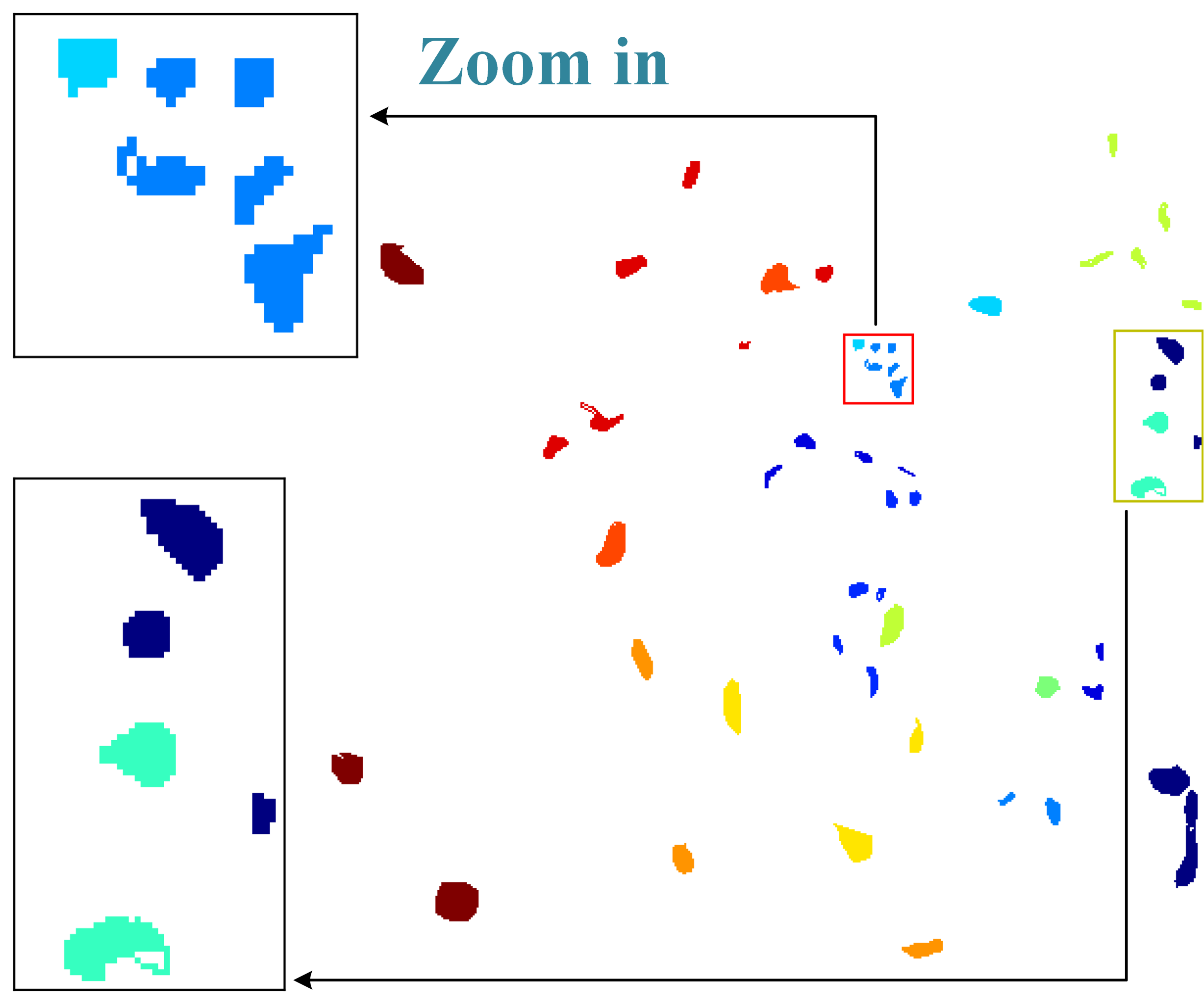}
		\label{fig:2}}    \hspace*{0.07em} 
	\subfloat[ ]{\centering
		\includegraphics[width=0.18\linewidth]{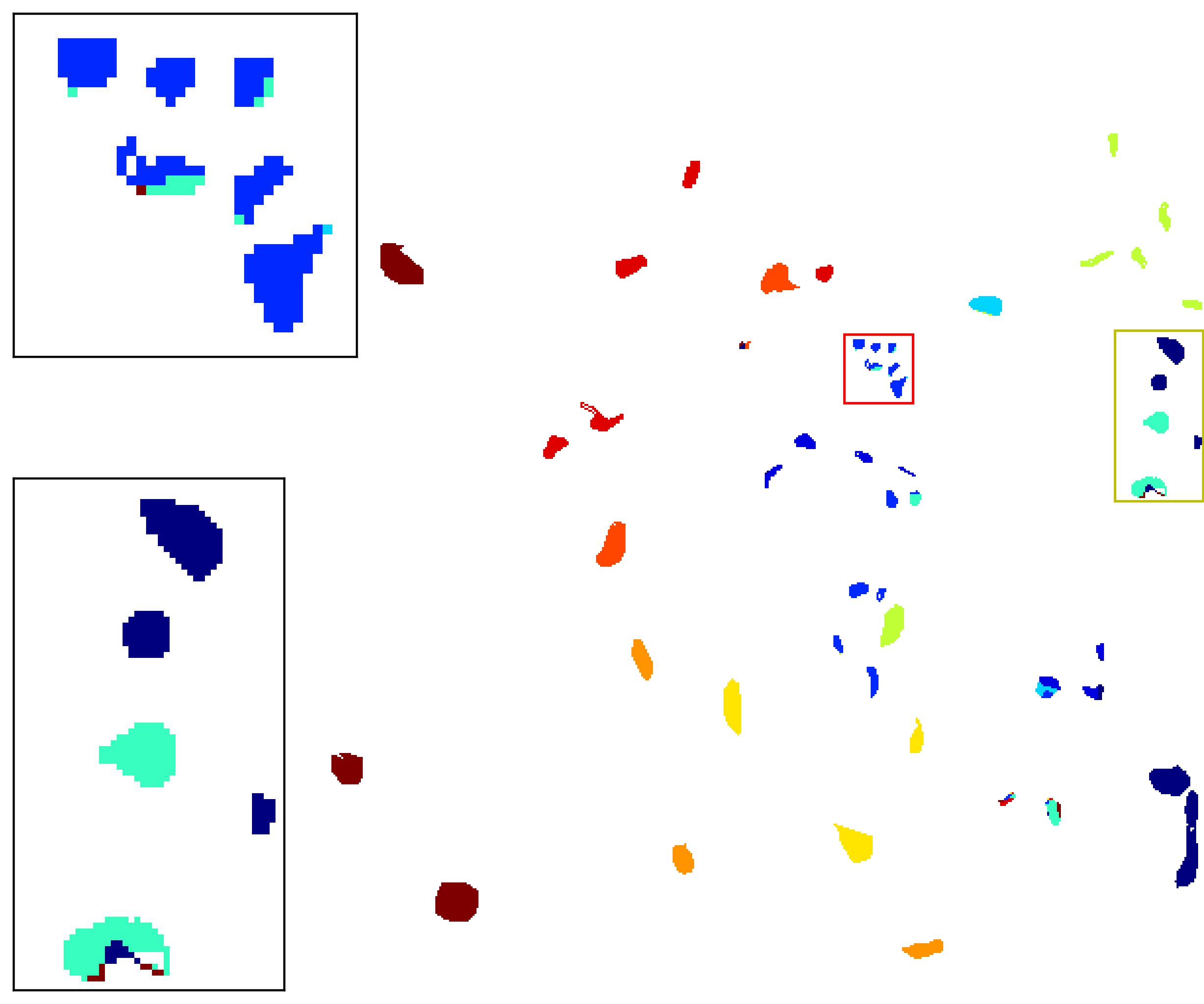}
		\label{fig:1} }  \hspace*{0.07em}    
	\subfloat[ ]{\centering
		\includegraphics[width=0.18\linewidth]{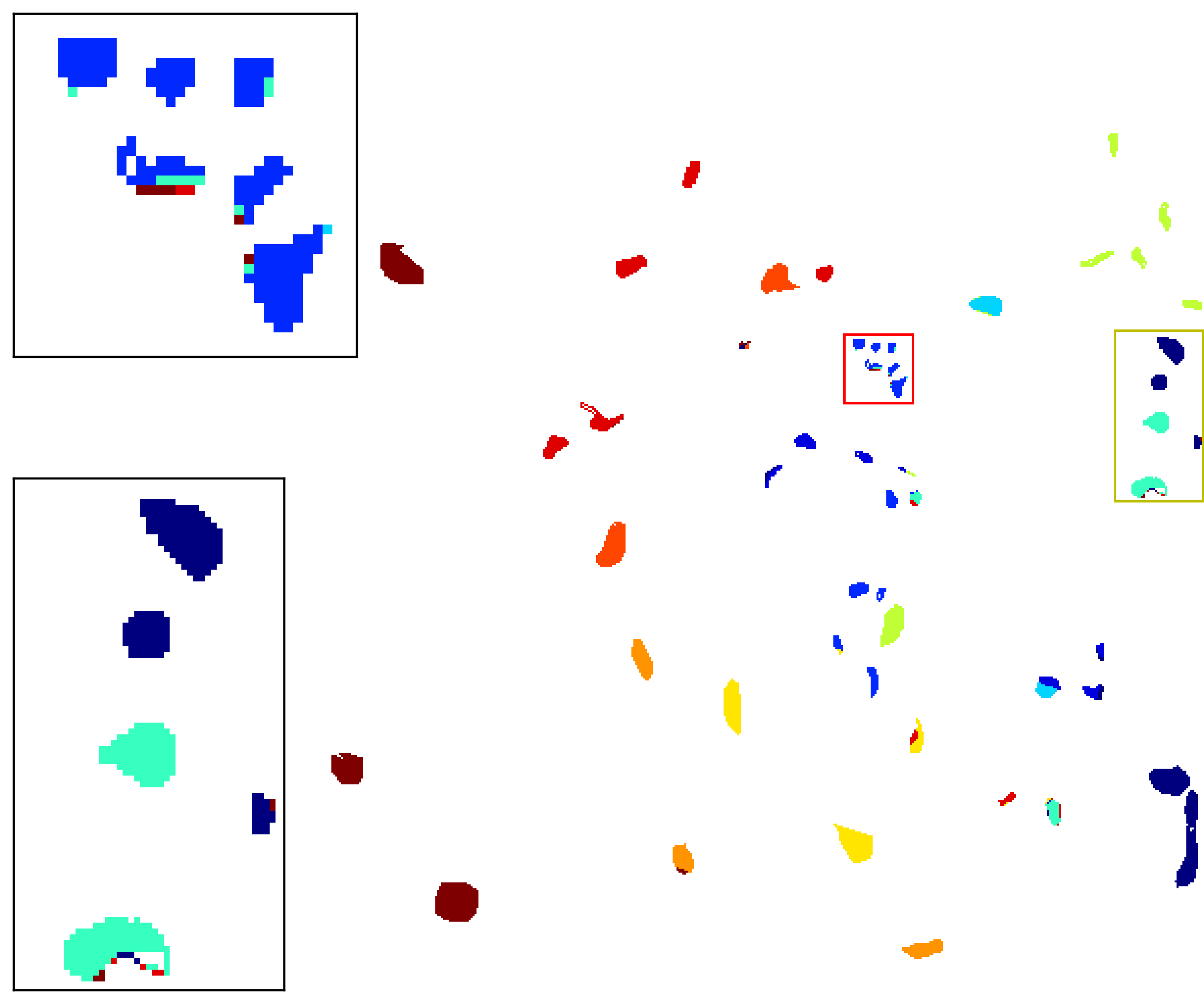}
		\label{fig:2}}    \hspace*{0.07em}      
	\subfloat[ ]{\centering
		\includegraphics[width=0.18\linewidth]{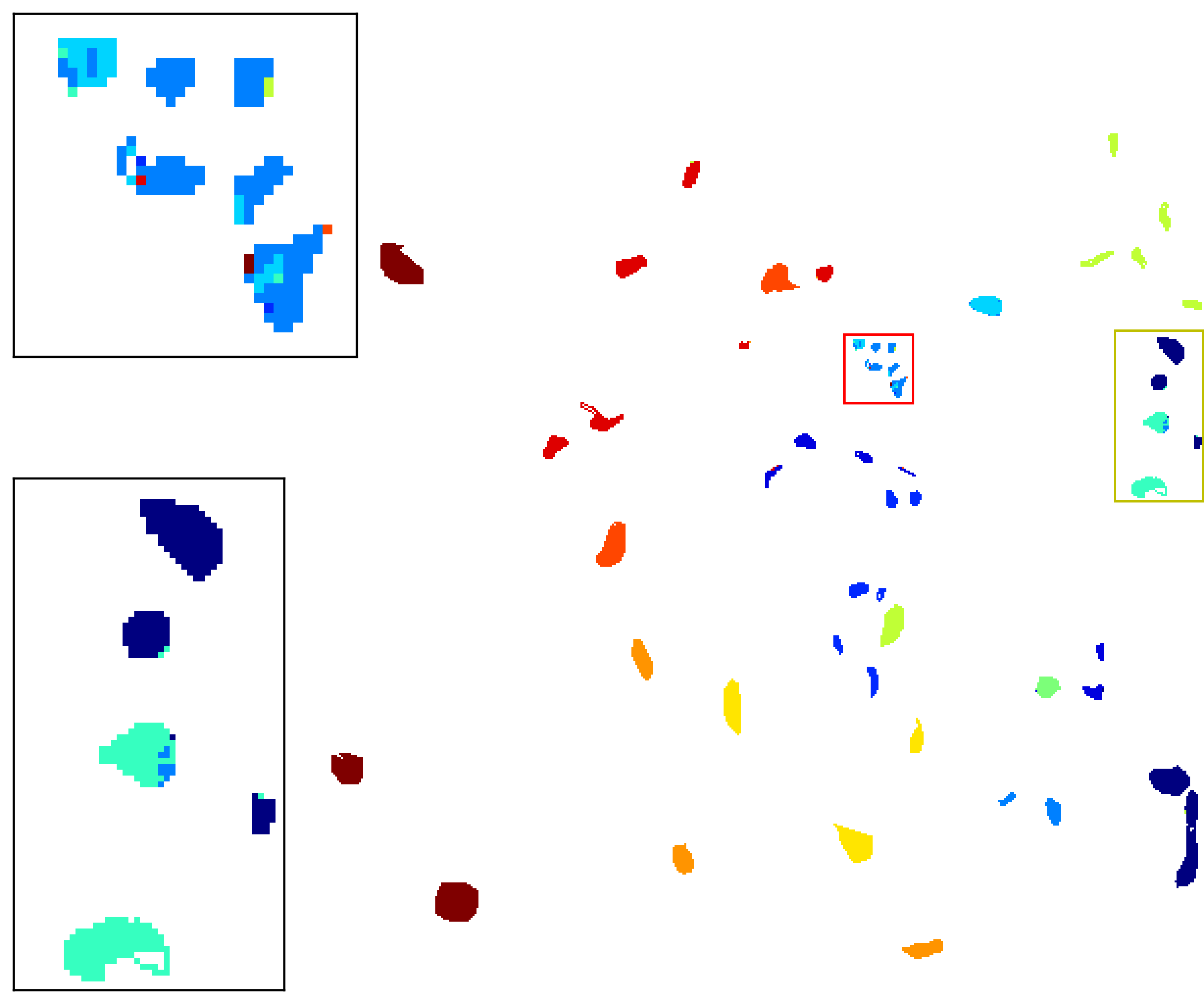}
		\label{fig:2}}    \hspace*{0.07em} \\
	\subfloat[ ]{\centering
	    \includegraphics[width=0.18\linewidth]{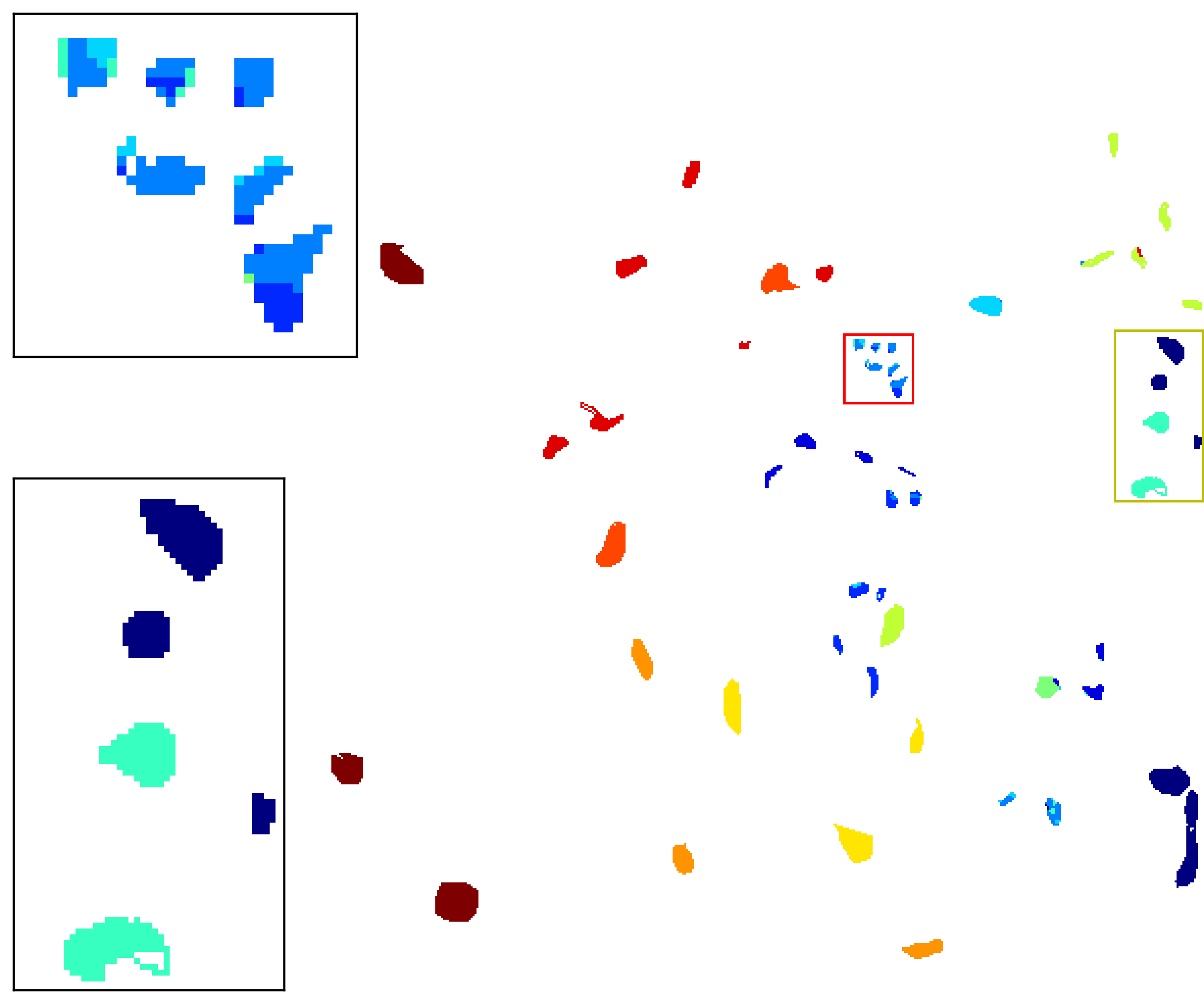}
	    \label{fig:1} }  \hspace*{0.07em}   
    \subfloat[ ]{\centering
	    \includegraphics[width=0.18\linewidth]{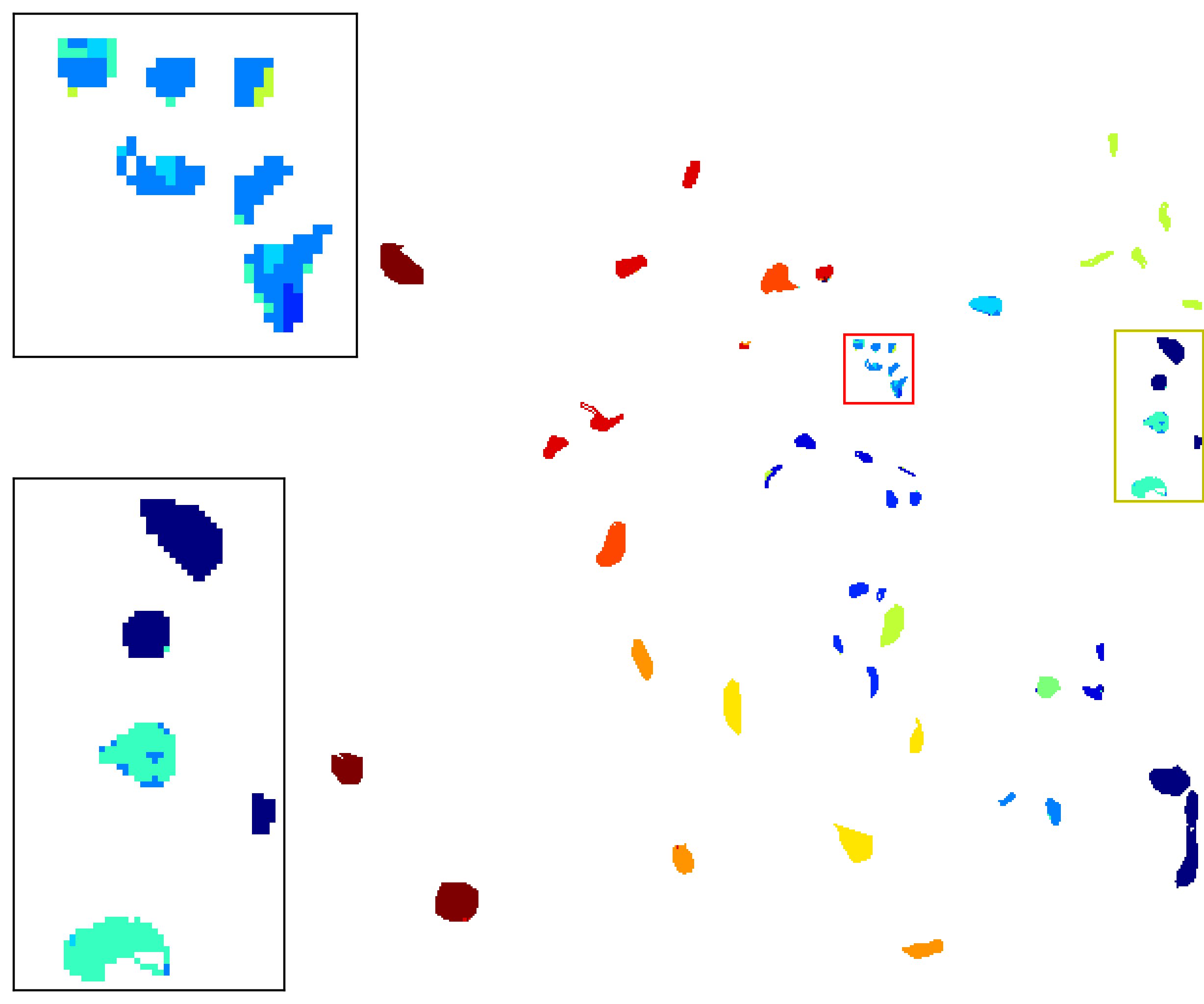}
	    \label{fig:1} }  \hspace*{0.07em}   
	\subfloat[ ]{\centering
	   \includegraphics[width=0.18\linewidth]{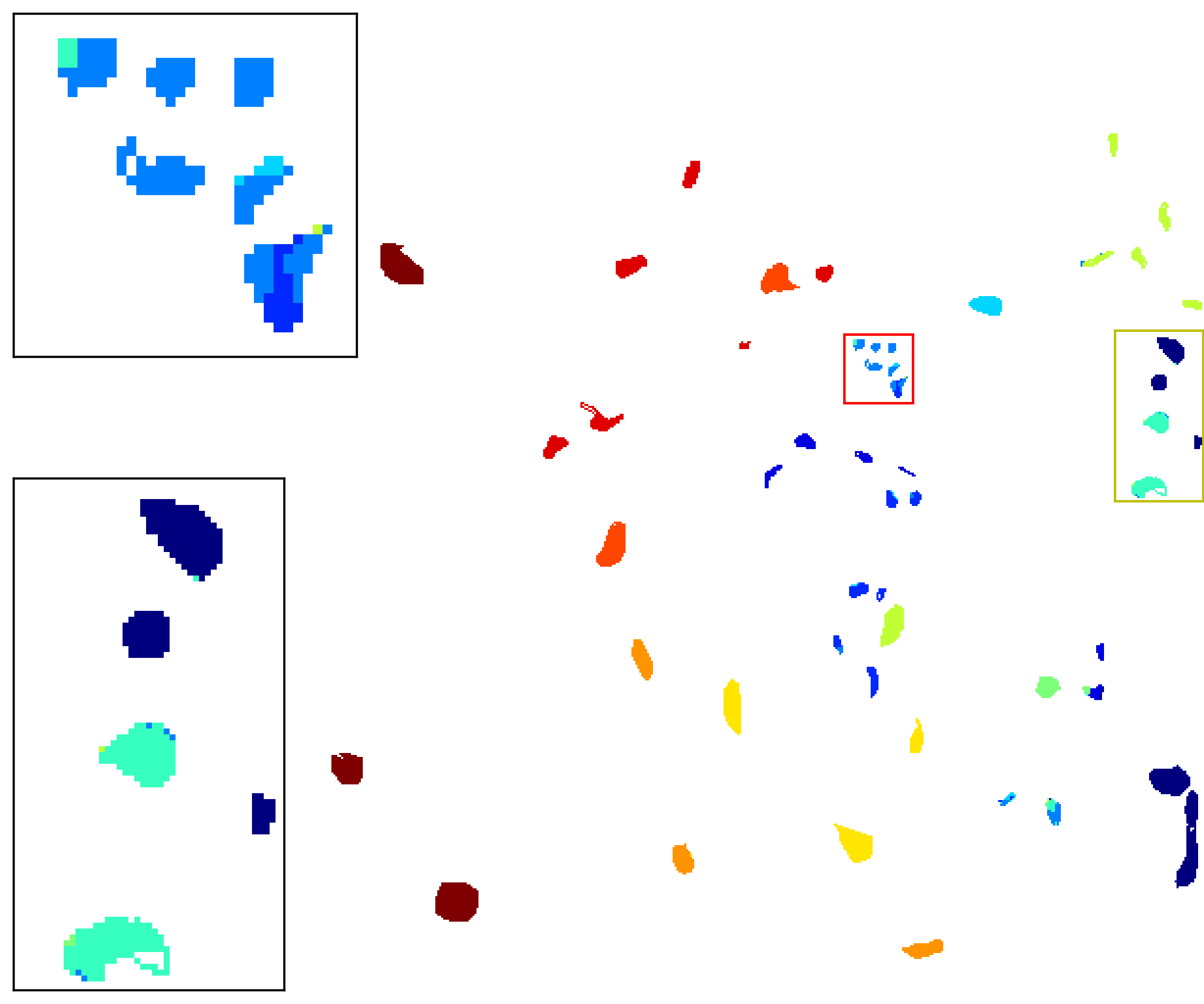}
	   \label{fig:1} }  \hspace*{0.07em}
	\subfloat[ ]{\centering
		\includegraphics[width=0.18\linewidth]{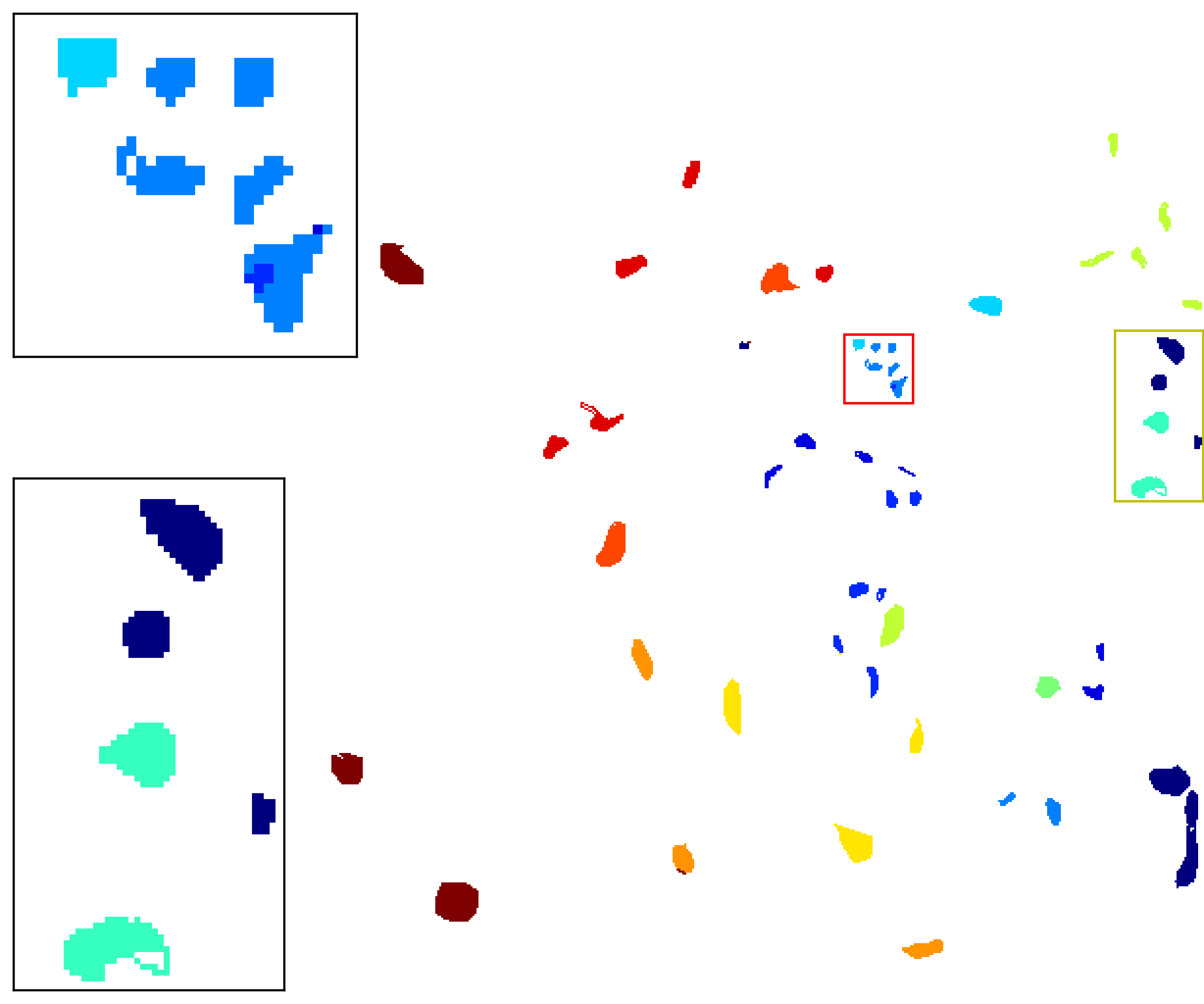}
		\label{fig:2}}    \hspace*{0.07em}    
	\subfloat[ ]{\centering
		\includegraphics[width=0.18\linewidth]{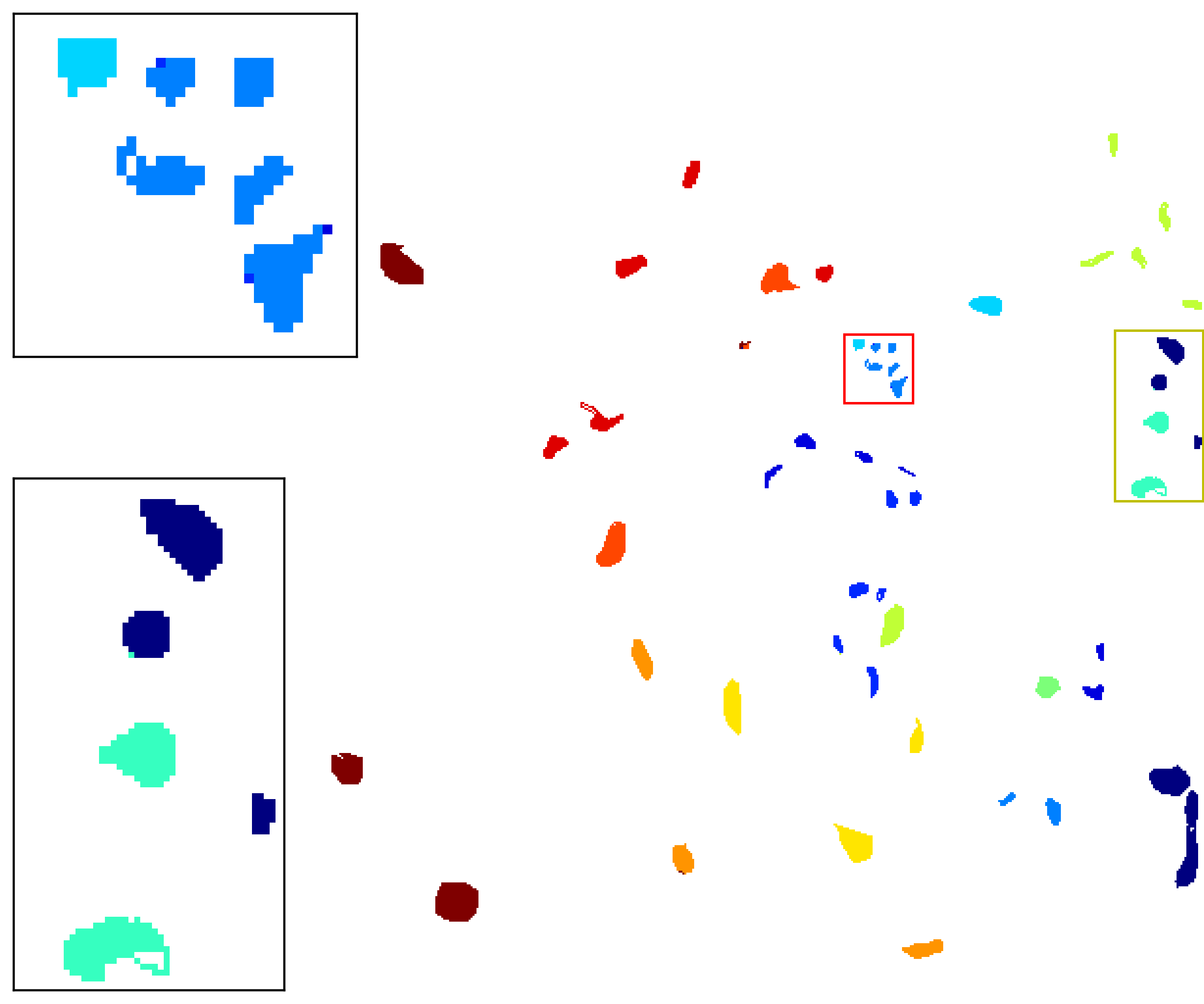}
		\label{fig:2}}    \hspace*{0.07em}         \\
	\subfloat{\centering
		\includegraphics[width=0.72\linewidth]{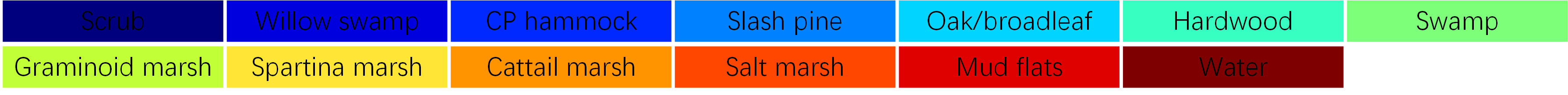}}  
	\caption{Classification results on KSC dataset. (a) False color image, (b) Ground truth, (c) 2D-CNN (OA=91.06\%), (d) 3D-CNN (OA=89.98\%), (e) SSRN (OA=97.88\%), (f) PyraCNN (OA=97.04\%), (g) HybridSN (OA=96.04\%), (h) FCN-CRF (OA=96.08\%), (i) Original non-local (OA=99.15\%), and (j) the proposed ENL-FCN (OA=99.46\%).}
	\label{figKSC}
\end{figure*}

\begin{figure*}[!t]	
	\centering
	\subfloat[]{\centering
		\includegraphics[width=0.32\linewidth]{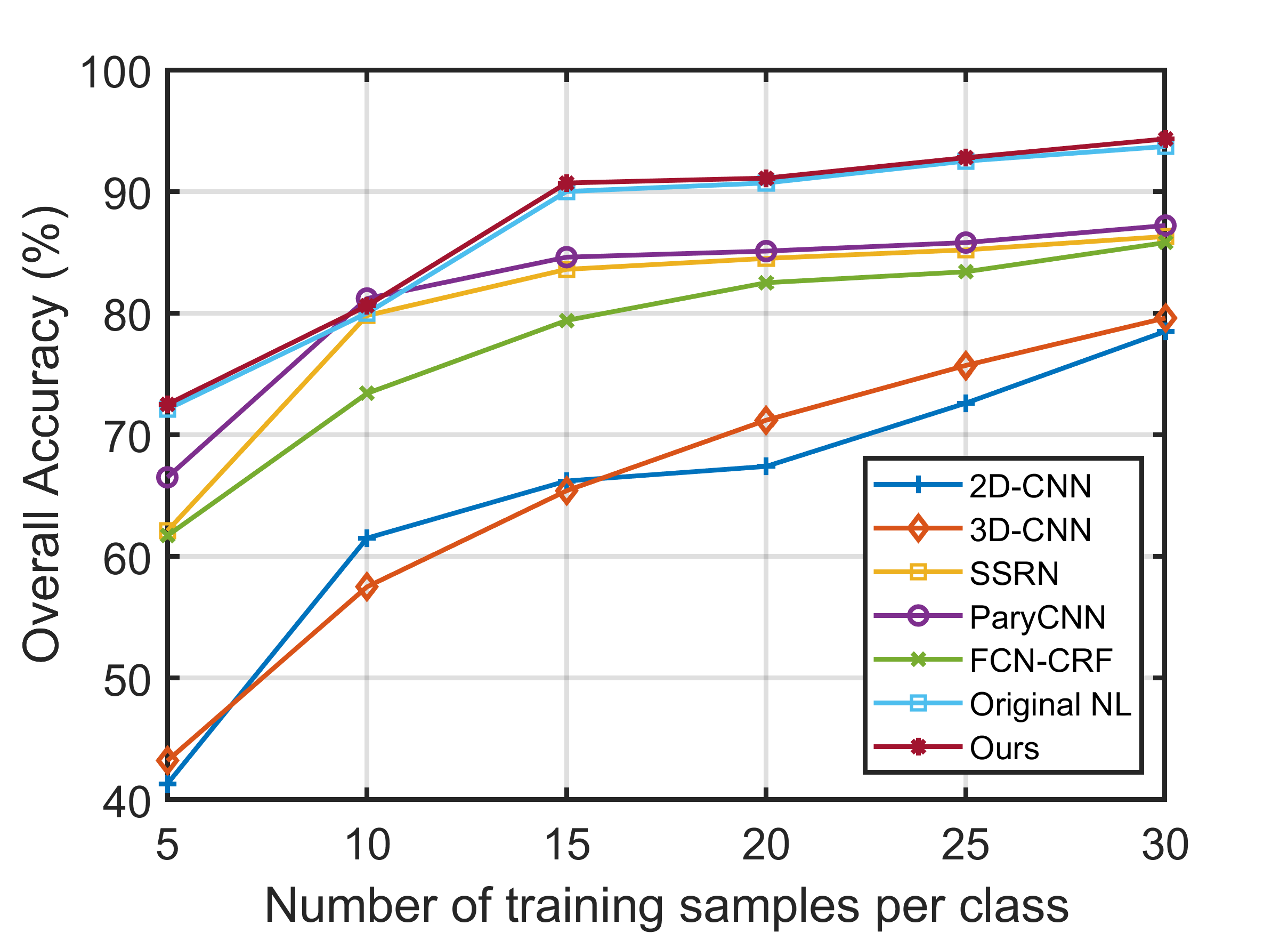}
		\label{fig:1} }  \hspace*{0.01em}
	\subfloat[]{\centering
		\includegraphics[width=0.32\linewidth]{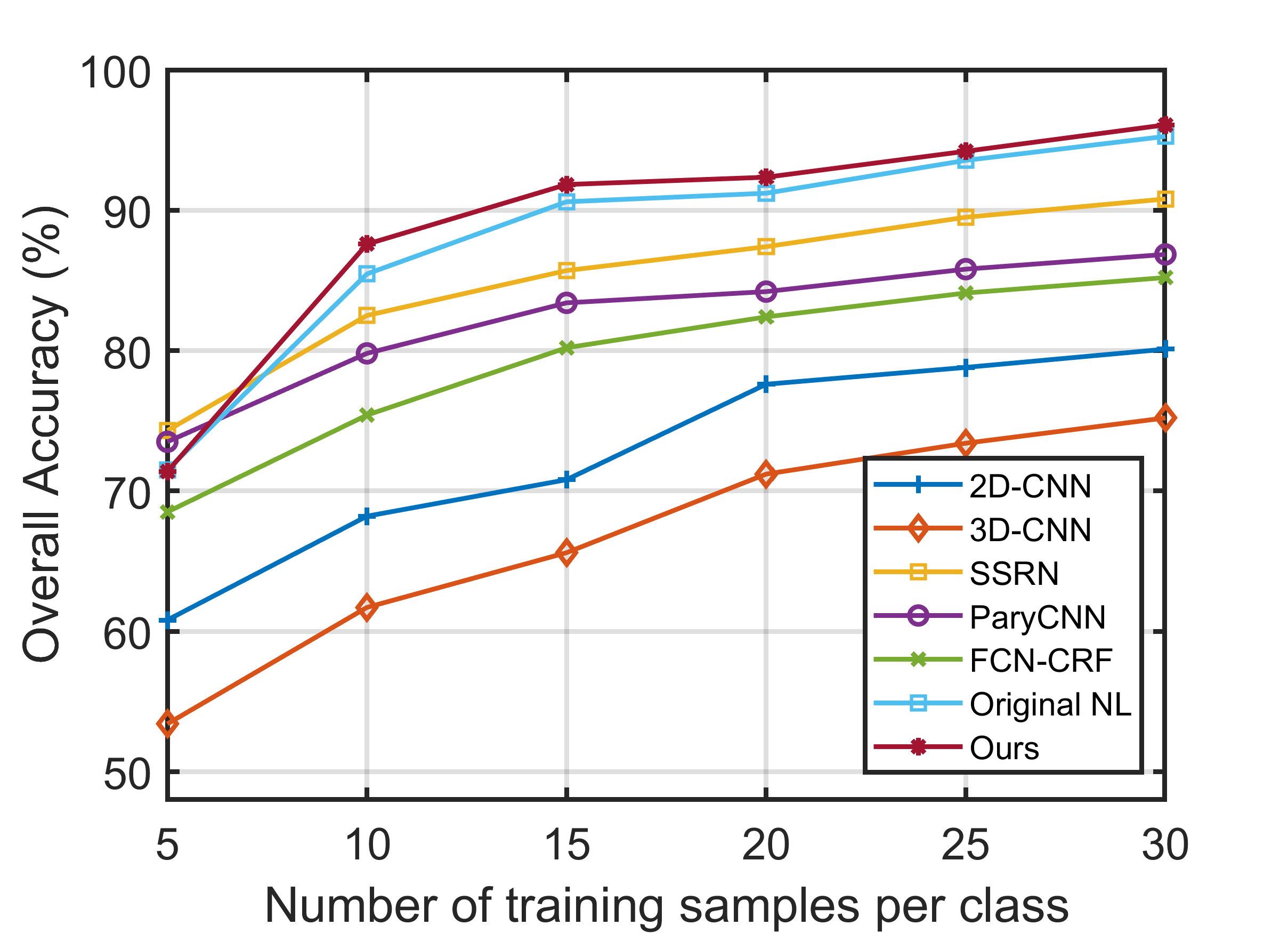}
		\label{fig:2}}    \hspace*{0.01em}
	\subfloat[]{\centering
		\includegraphics[width=0.32\linewidth]{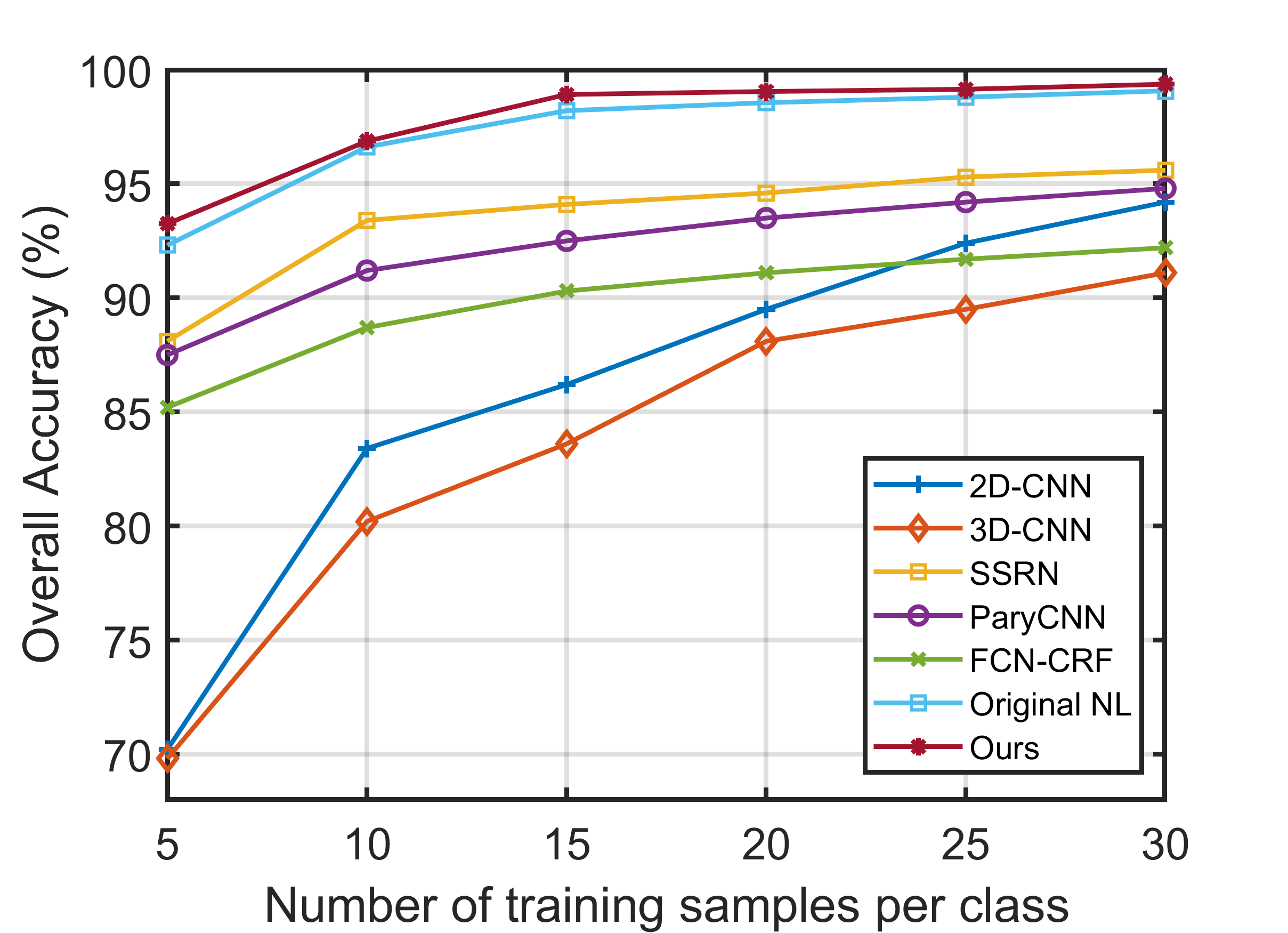}
		\label{fig:2}}    \hspace*{0.01em}
	\caption{OA of different methods with different numbers of training samples. (a) IP, (b) PU, (c) KSC.}
	\label{figSmallSample}
\end{figure*}

\subsection{Classification Performance}
In the experiments, several state-of-the-art deep learning based classifiers are used in comparison with the proposed ENL-FCN, including 2D-CNN \cite{makantasisDeepSupervisedLearning2015}, 3D-CNN \cite{liSpectralSpatialClassification2017b}, spectral-spatial residual network (SSRN) \cite{zhongSpectralSpatialResidual2018b}, deep pyramidal residual network (PyraCNN) \cite{paolettiDeepPyramidalResidual2019}, HybridSN \cite{royHybridSNExploring3D2020}, and FCN-CRF \cite{xuPatchwiseClassificationSpectralSpatial2019}. These algorithms exploit spectral-spatial information for HSI classification. Particularly, FCN-CRF uses the CRF model to incorporate long-range contextual information. The parameters of these deep learning methods are set according to their reference papers. We also employ the original non-local module \cite{Wang_2018_CVPR} in our FCN backbone as a baseline, denoted as Original non-local. In other words, we replace the efficient non-local module with the original non-local unit in the FCN. Due to the limitation of GPU memory, only one original non-local unit can be embedded in the FCN architecture. 

In the experiments, besides per-class accuracy, the overall accuracy (OA), average accuracy (AA) and kappa coefficient \cite{thompson1988reappraisal} are calculated to evaluate the classification performance comprehensively. OA is calculated by the ratio between the correctly classified test pixels and the total number of test pixels. AA is the average of the classification accuracy of each class. The kappa coefficient is computed by weighting the measurement accuracy \cite{thompson1988reappraisal}. It incorporates both of the diagonal and off-diagonal entries of the confusion matrix. The experiment is repeated ten times for all approaches to yield the average performance for a fair comparison. The classification maps and corresponding accuracies of different methods over three datasets are reported in Figs. \ref{figIP}-\ref{figKSC} and Tables \ref{tabIP}-\ref{tabKSC}.

For the IP dataset, our method provides the best classification performance in terms of OA and kappa index. For the classifiers that do not consider long-range contextual information (i.e., 2D-CNN and 3D-CNN), more mis-classification can be found at the edges of different land-cover regions, as shown in Fig. \ref{figIP}. Table \ref{tabIP} reveals that 2D-CNN shows a significant decline (up to 3\%) on OA compared with our method, indicating that a simple CNN framework fails to characterize the features of different regions. Due to the data imbalance, the OA of FCN based frameworks, including FCN-CRF, original non-local and our method, show an accuracy decline on Class 9. Overall, the proposed ENL-FCN generates the most accurate classification map.

From Table \ref{tabPU}, we can observe that our framework also achieves a better classification performance than other CNN based classifiers on PU dataset. The PU dataset contains more isolated points and small regions, which would increase the difficulty in distinguishing land cover classes. For example, 3D-CNN fails to provide a better performance on PU, whose OA declines to 88.69\%. Compared with other classifiers, our method offers obvious better classification accuracy in Table \ref{tabPU}, and the OA of our method is up to 99\%. Furthermore, according to the classification results of Fig. \ref{figPU}, less mis-classification can be found in the classification map of our method compared with other deep learning approaches.

Table \ref{tabKSC} reports the experimental results of different methods on KSC dataset.
%A similar observation of classification performance can be found on this dataset KSC. For KSC dataset, a more significant improvement on OA is achieved, as shown in Table \ref{tabKSC}. 
From this table, the OA of our method is much higher than other CNN based classifiers (up to 8\%). In Fig. \ref{figKSC}, we zoom in two critical regions of KSC for better visualization, which are marked with red and yellow rectangles. As shown in the zoomed regions, the class \emph{Oak/broadleaf} is difficult to distinguish. In Figs. \ref{figKSC}(c) and (d), the \emph{Oak/broadleaf} is almost mis-classified. Compared with other classifiers, our method achieves much better accuracy on class \emph{Oak/broadleaf}. 
%Overall, our method achieves a better visual effect.

\begin{table*}[!t]
	\centering
	\caption{Computational cost (FLOPs) and memory usage (MB) of two non-local modules over three datasets.}
	\begin{tabular}{cc|ccc}
		\hline 
		 & &  IP & PU & KSC \\
		\hline
		 \multirow{2}{*}{FLOPs} & Original non-local module & $1.32\times10^{11}$ & $1.329\times10^{13}$ & $2.97\times10^{13}$\\
	      & Efficient non-local module& $\mathbf{3.64\times10^9}$ & $\mathbf{1.18\times 10^{11}}$ & $\mathbf{2.12\times 10^{11}}$\\
		\hline
		\multirow{2}{*}{GPU memory (MB)} & Original non-local module &  6166 & 7122 & 8152\\
	     & Efficient non-local module  & \textbf{1928} & \textbf{1801} & \textbf{2342}\\
		\hline  
	\end{tabular}
\label{tabCost}
\end{table*}

Fig. \ref{figSmallSample} reports the classification performance under the case of small training sample size. In this experiment, the number of training samples varies from 5 to 30 per class over three datasets. It is evident that there is a clear improvement on the classification performance as the number of training samples increases. The 2D-CNN and 3D-CNN usually provide the lowest OA over three datasets. Some more complicated networks, such as SSRN, PyraCNN and FCN-CRF, achieve better classification performance than 2D-CNN and 3D-CNN. Since FCN-CRF separates the network into two phases, the OA is lower than SSRN and PyraCNN. In Fig. \ref{figSmallSample}(b), the OA of our method is slightly lower than SSRN and PyraCNN when the number of training samples is 5. Generally, the proposed method provides higher OA compared with other deep learning networks. 

From the above tables, we can also observe that the proposed network achieves the classification accuracy similar to the original non-local network. This is reasonable because both approaches are able to extract the long-range contextual information from HSI. Owing to the efficient non-local module being applied twice in the proposed framework, we can find that the proposed ENL-FCN provides a slightly better overall accuracy than the original non-local approach. 

\begin{figure*}[!t]
	\centering
		\subfloat[ ]{\centering
		\includegraphics[width=0.22\linewidth]{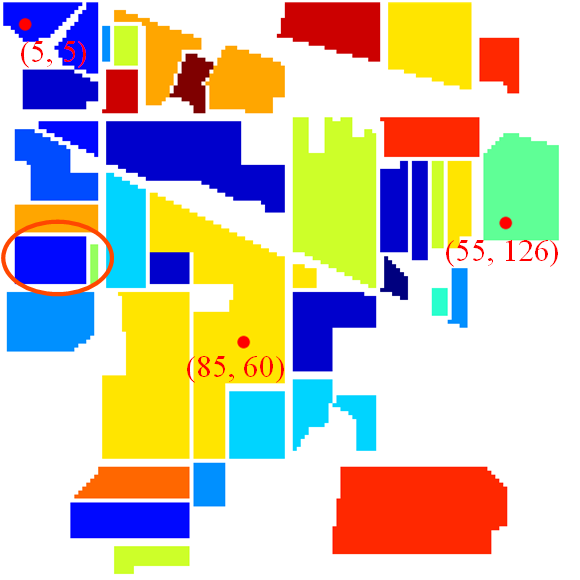}
		}    \hspace*{0.1em}
	\subfloat[ ]{\centering
		\includegraphics[width=0.22\linewidth]{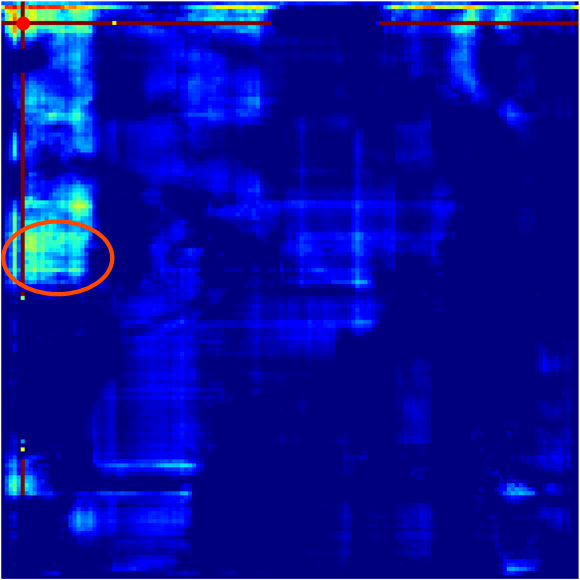}
		\label{figPoint55}}    \hspace*{0.1em} 
	\subfloat[ ]{\centering
		\includegraphics[width=0.22\linewidth]{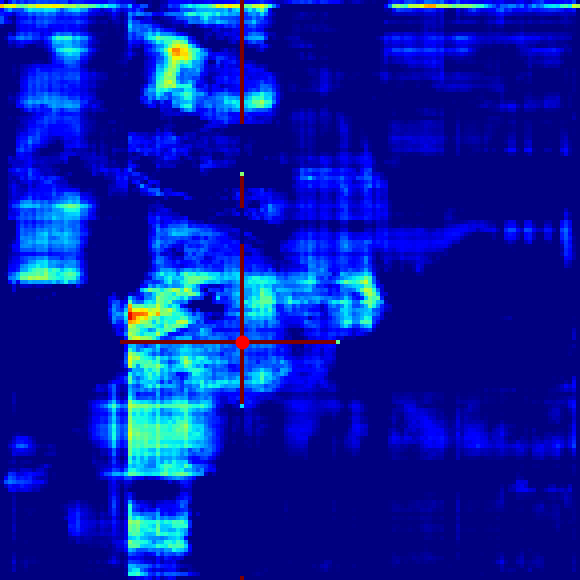}
		 }  \hspace*{0.1em}    
	\subfloat[ ]{\centering
		\includegraphics[width=0.22\linewidth]{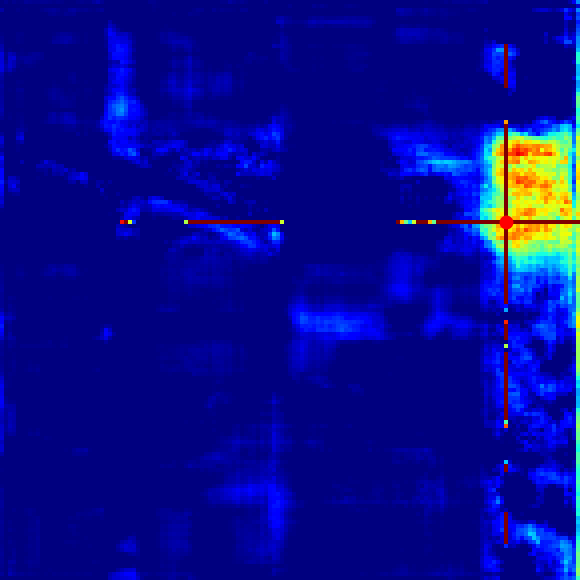}
		}    \hspace*{0.1em}      
\caption{Visualization of pixel correlation maps on IP dataset. (a) Ground-truth label map. The numbers in the parentheses denote the row and column numbers, e.g., (5, 5) denotes the pixel (red dot) position at the $5$th row and $5$th column. (b) Correlation map for pixel at position (5, 5). (c) Correlation map for pixel at position (85, 60). (d) Correlation map for pixel at position (55, 126). The red circles in (a) and (b) mark the region that is in the same class with position (5, 5). The red circle region and the nearby region of position (5, 5) have similar responses (i.e., correlation values).}
	\label{figPoints}
\end{figure*}

\begin{table}[!t]
	\centering
	\caption{Effectiveness analysis (OA\%) of the efficient non-local module.}
	\begin{tabular}{c|ccc}
		\hline 
		Dataset &  IP & PU & KSC \\
		\hline
		Baseline (only FCN) & 97.66 & 98.05 & 98.19 \\ 
		\hline
		Efficient non-local (one module) & 98.75 & 98.71 & 99.30\\
		Replaced with $1\times1$ kernel (one conv.) & 97.85 & 98.15 & 98.53\\
		\hline
		Efficient non-local (two modules in parallel) & \textbf{98.85} & \textbf{99.08} & \textbf{99.46}\\
		Efficient non-local (two modules in series) & 98.82 & 99.07 & 99.43\\
		Replaced with $1\times1$ kernel (two conv.) & 98.14 & 98.35 & 98.65\\
		\hline  
		Efficient non-local (three modules in parallel) & \textbf{98.92} & 99.15 & \textbf{99.60}\\
		Efficient non-local (three modules in series) & 98.91 & \textbf{99.18} & 99.55\\
		Replaced with $1\times1$ kernel (three conv.) & 98.82 & 99.07 & 99.12\\
		\hline
	\end{tabular}
\label{tabAblation}
\end{table}

\begin{table}[!t]
	\centering
	\caption{Effectiveness analysis (OA\%) of feature concatenation in our framework.}
	\begin{tabular}{c|ccc}
		\hline 
		Dataset &  IP & PU & KSC \\
		\hline
		With feature concatenation & \textbf{98.85} & \textbf{99.08} & \textbf{99.46}\\
		Without feature concatenation & 98.71 & 98.84 & 99.24\\
		\hline  
	\end{tabular}
\label{tabCat}
\end{table}

\subsection{Analysis of Computational Cost and Memory Usage}
As discussed in Section \ref{subCost}, the efficient non-local module consumes much less computational resources than the original one \cite{Wang_2018_CVPR}. It is worth mentioning that we have to divide the PU and KSC datasets into smaller data cubes  because the memory usage of the original non-local module exceeds our experimental limitation.  The PU dataset is equally divided into six regions. Although our proposed method does not have to divide a dataset for network training, we still keep the same setup with the original non-local module. Table. \ref{tabCost} compares the computational cost and GPU memory usage (MB) of the proposed framework and the original non-local network over three HSI datasets. The computational cost is calculated by floating point operations (FLOPs). From the results, we can observe that the original non-local network uses over three times GPU memory and one hundred times FLOPs than our proposed network on KSC dataset, indicating that the efficient non-local module can significantly save the computational resources.

\begin{figure}[t]
	\centering
	\subfloat{
		\includegraphics[width=0.8\linewidth]{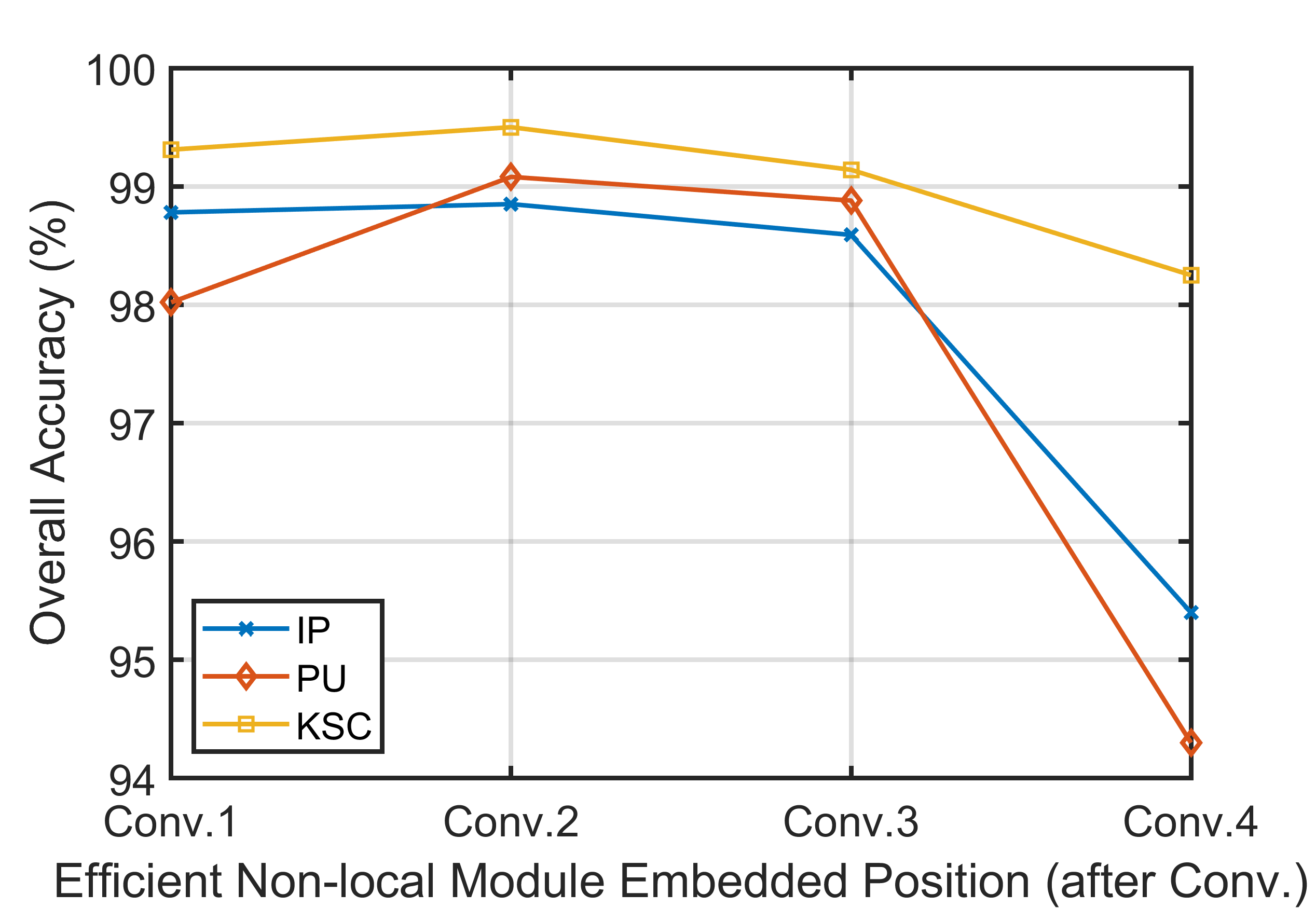}
		\label{fig:1} } 
	\caption{Influence of the efficient non-local module embedded position in convolutional layers.}
	\label{figCCPosition}
\end{figure}

\begin{figure}[t]
	\centering
	\subfloat{
		\includegraphics[width=0.8\linewidth]{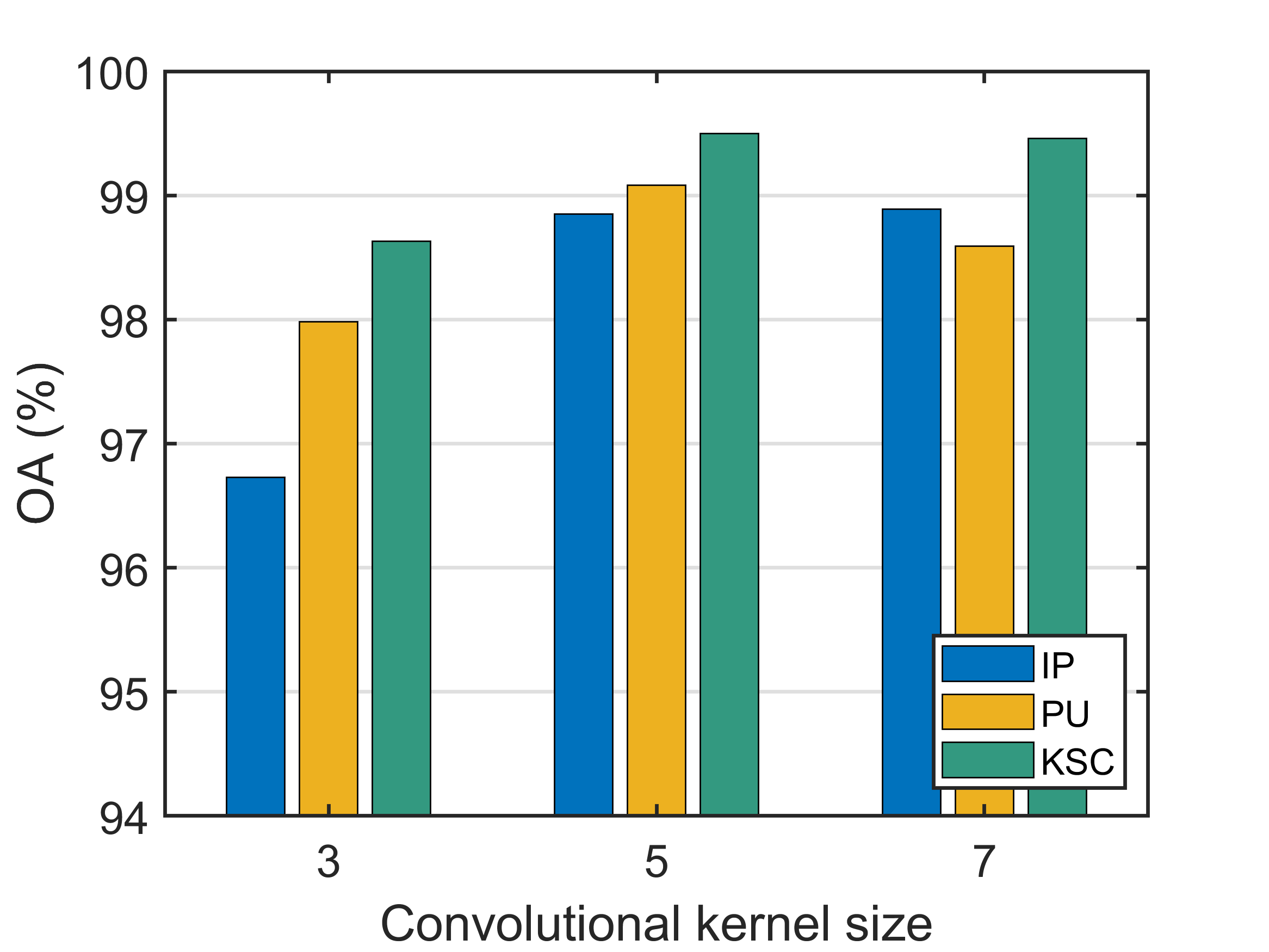}
		\label{fig:1} } 
	\caption{Sensitivity analysis over the convolutional kernel size.}
	\label{figKernelSize}
\end{figure}

\subsection{Ablation Study for Efficient Non-local Module}
In this section, to further evaluate the effectiveness of the efficient non-local module, several ablation experiments are conducted. (1) In the proposed network, the efficient non-local modules are replaced with the regular $1\times1$ convolutional kernels, while keeping the number of learning parameters the same. (2) The influence of applying multiple efficient non-local modules is analyzed. (3) The structure of employing multiple efficient non-local modules (in parallel or in series) is also discussed. 
%sijie:
% \paragraph{baseline vs 1X1 conv vs nonlocal}
% \paragraph{1 vs 2 vs 3 nonlocal modules}
% \paragraph{parallel vs series}
% \paragraph{concat vs no concat}

Compared with the performance of baseline (only FCN), the network with efficient non-local modules provides a significant improvement in OA, as shown in Table \ref{tabAblation}.  We can also observe that the OA of efficient non-local modules outperforms that of regular convolutions ($1\times1$ $conv$) when the number of learning parameters is the same, which demonstrates the effectiveness of the efficient non-local module. 

The impact of the number of efficient non-local modules on the classification performance is evaluated. For comparison, we embed different numbers of efficient non-local modules (i.e., 1$\sim$3 modules) in parallel in the network. Based on the results in Table \ref{tabAblation}, an improvement can be observed when more efficient non-local modules are added in the network. To balance between the computational efficiency and classification accuracy, we use two efficient non-local modules in the proposed framework.

We also investigate the structure of embedding the efficient non-local modules in the network, including two modules in parallel and two modules in series. From Table \ref{tabAblation}, we can observe a similar performance when the efficient non-local modules are placed in parallel or in series. The parallel architecture slightly outperforms the series one.

%Table \ref{tabAblation} reports the experimental results of our ablation study.  From Table \ref{tabAblation}, we can observe that the OA of efficient non-local modules outperforms the OA of regular convolution when the number of learning parameters is the same, demonstrating the effectiveness of efficient non-local module. In addition, when more efficient non-local modules are applied, the classification performance is improving. When the efficient non-local modules are placed in parallel, the OA is slightly better than the series format in most cases. However, it is time consuming when more efficient non-local modules are added in the network. To ensure the computational efficiency, we add two efficient non-local modules in parallel in the proposed framework.

To evaluate the effectiveness of feature concatenation in our framework, Table \ref{tabCat} reports the classification performance with and without feature concatenation in the proposed framework. As illustrated in Table \ref{tabCat}, the feature concatenation is beneficial for further improving the classification performance of our framework. 
%Compared the OA of without feature concatenation, the classification accuracy has an obvious increase when using feature concatenation.

The embedded position for the efficient non-local module is also discussed. As shown in Fig. \ref{figFlowchart}, it is placed after the second convolutional layer (Conv.2). In this experiment, it is placed after the first (Conv.1), third (Conv.3) or fourth (Conv.4) convolutional layer, and then the classification accuracy of each position is recorded.  As shown in Fig. \ref{figCCPosition}, the network performs better when the efficient non-local module is embedded after the second and third convolutional layers. The OA is the highest when the module is placed after the second convolutional layer (Conv.2). When the efficient non-local module is embedded after the fourth convolutional layer, a sharp accuracy decline can be observed in Fig. \ref{figCCPosition}. For the PU dataset, almost 5\% accuracy is dropped when the efficient non-local module is placed after Conv.4. This could indicate that a high-level feature representation may not be helpful for capturing the long-range contextual information. Therefore, based on this ablation study, we embed the efficient non-local module after the second convolutional layer to achieve the optimal performance.

\subsection{Visualization of Efficient Non-local Module}
As discussed in Section \ref{SecMethod}, non-local information is captured and saved in attention maps $\mathbf{A}$ and $\mathbf{A}'$. The response of a pixel with others is established by Eq. \ref{eqMessage} in a message passing procedure. Therefore, for each pixel in an HSI, there is a corresponding $H\times W$ pixel correlation map that records the responses from the whole image. In Fig. \ref{figPoints}, we select three pixels from IP dataset and visualize the corresponding pixel correlation maps according to Eq. \ref{eqMessage}. 
%Based on the correlation maps, we shed light on several key insights. 
As shown in Fig. \ref{figPoints}, the response of each pixel is significant in its criss-cross way. It can be seen that the efficient non-local module can capture semantic information from long-range regions in an HSI. For example, considering the pixel at position (5, 5) (the red dot at the top-left corner in Fig. \ref{figPoints}(a)), its correlation map (Fig. \ref{figPoints}(b)) reveals that the long-range region, which is highlighted in the red circle, yields the responses similar to the neighboring area of pixel (5, 5).
%since they belong to the same class as the pixel of interest. 
This indicates that the contextual information is captured not only from a pixel's surrounding area but also from other regions (highlighted in the red circle)  belonging to the same class.

\begin{figure*}[!t]
	\centering
		\subfloat[ ]{\centering
		\includegraphics[width=0.31\linewidth]{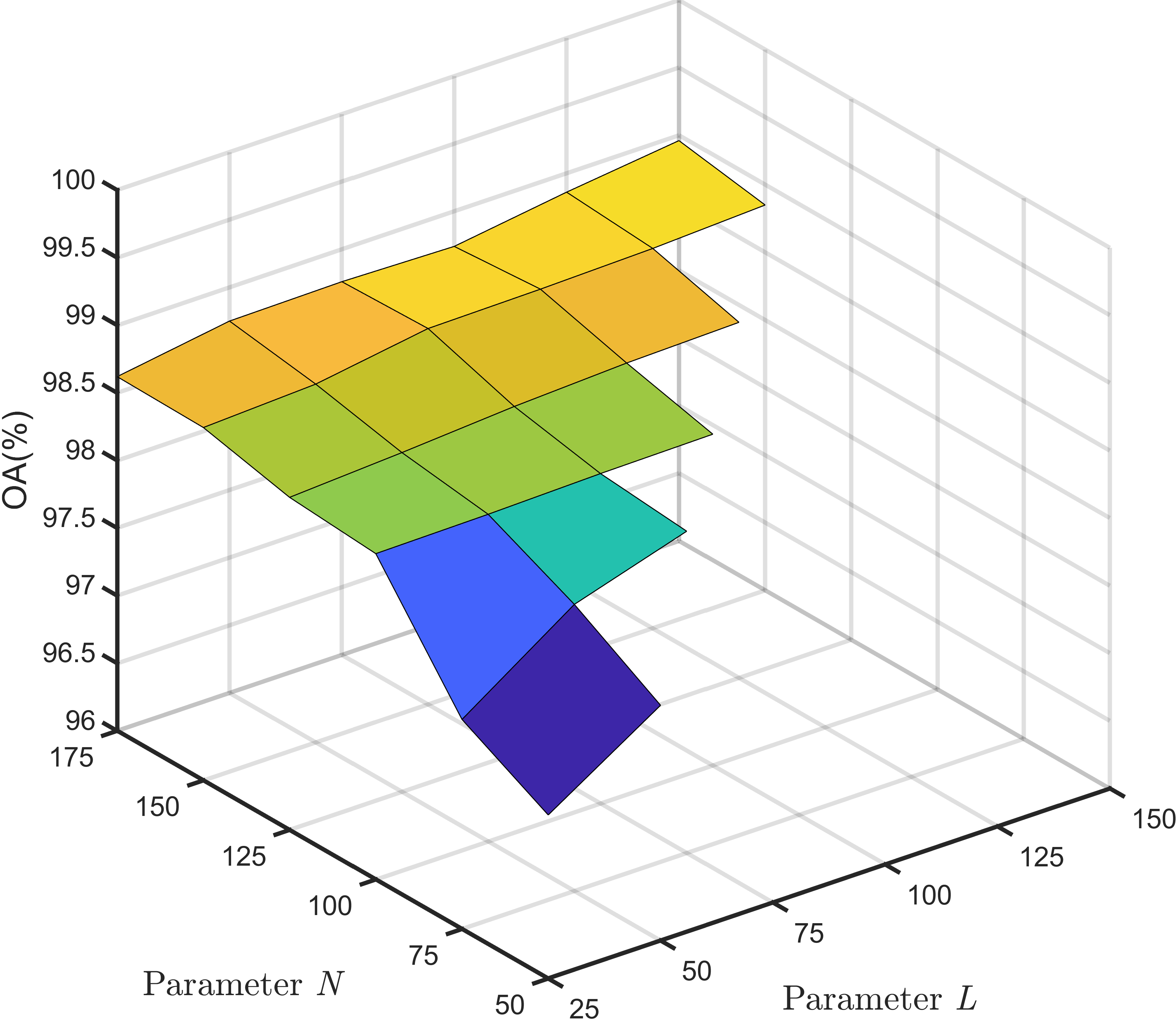}
		}    \hspace*{0.1em}
	\subfloat[ ]{\centering
		\includegraphics[width=0.31\linewidth]{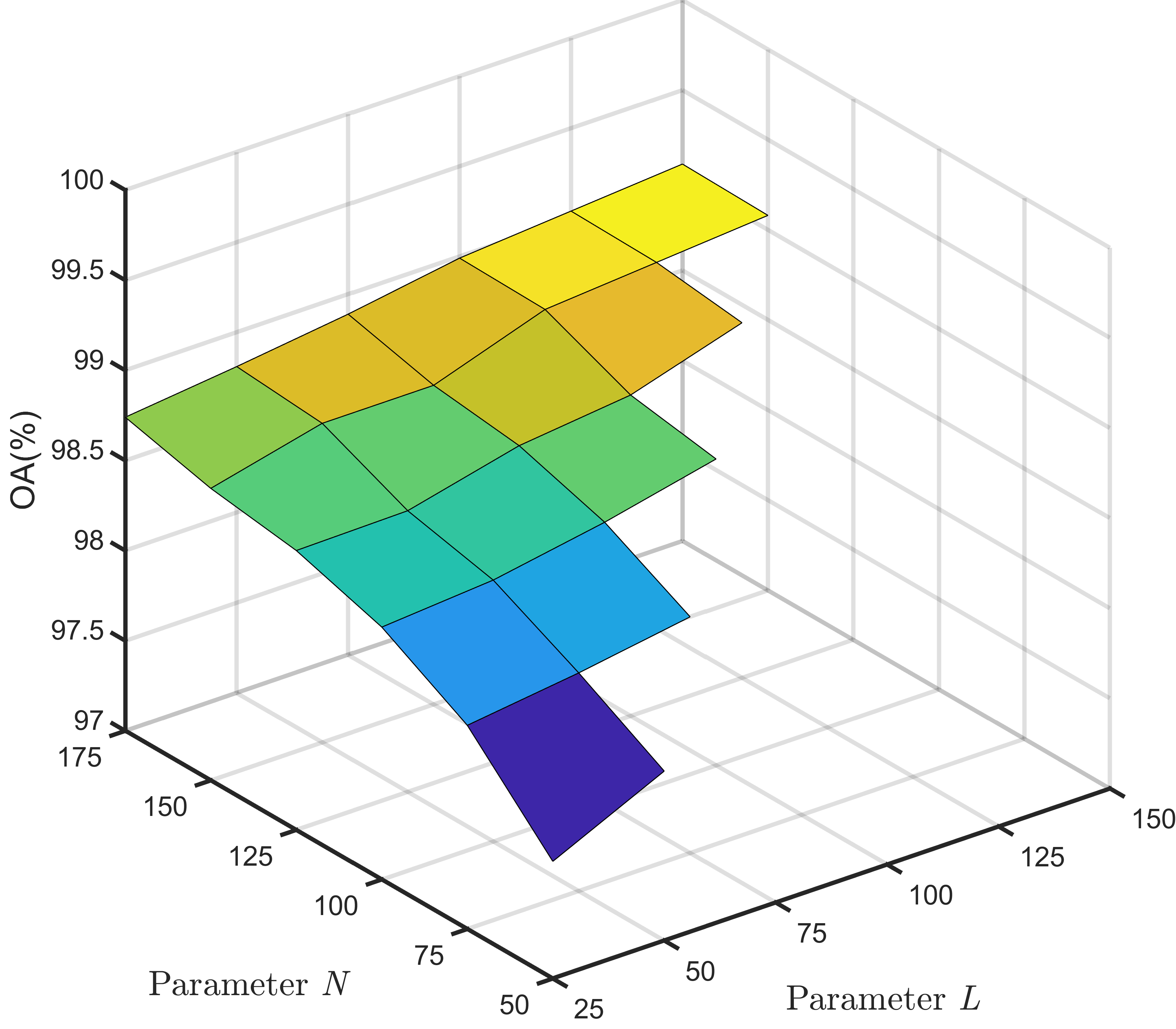}
		\label{figPoint55}}    \hspace*{0.1em} 
	\subfloat[ ]{\centering
		\includegraphics[width=0.31\linewidth]{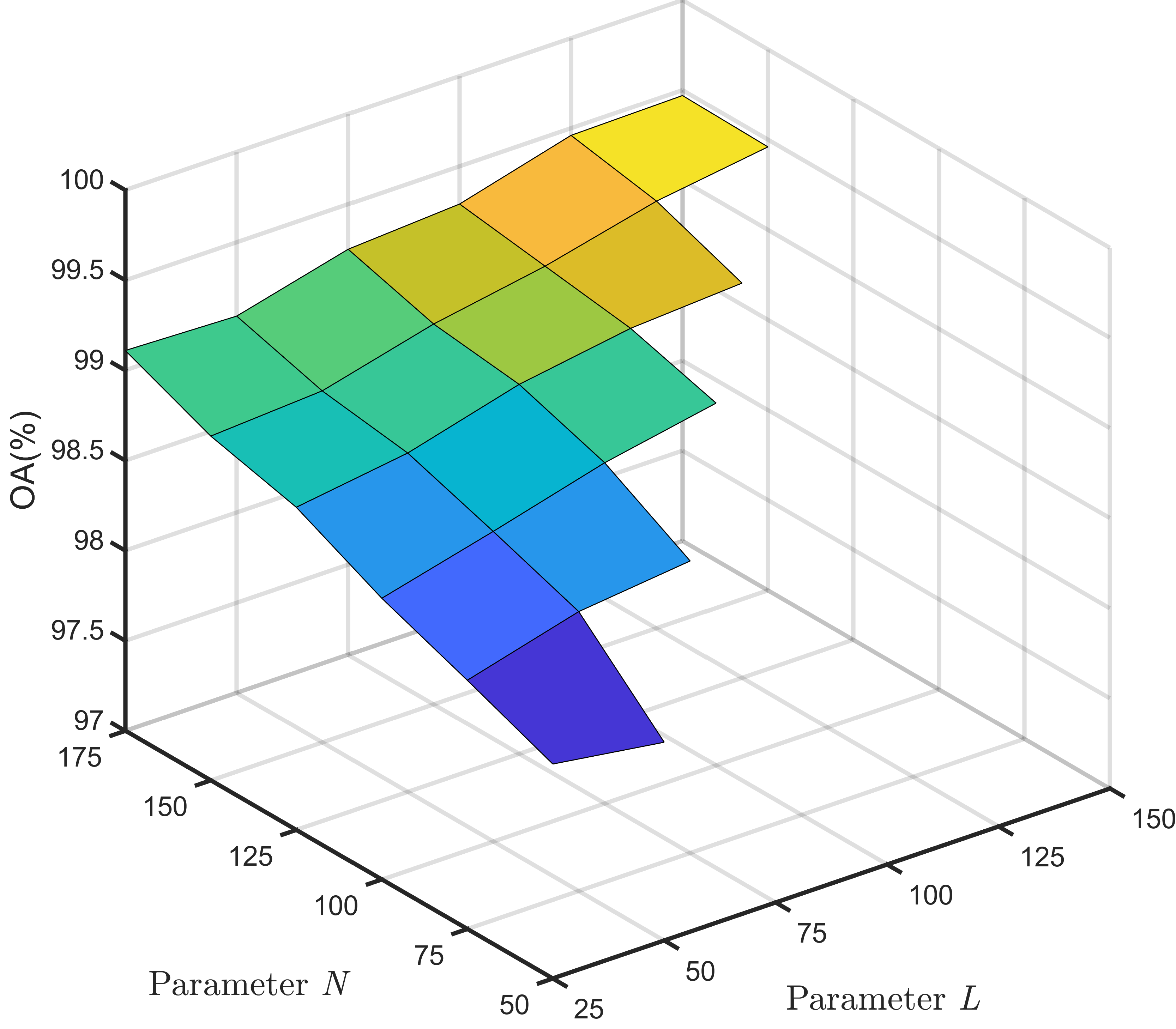}
		 }  \hspace*{0.1em}    
\caption{Parameter analysis of convolutional kernel number in FCN (Parameter $N$) and efficient non-local module (Parameter $L$) over three HSI datasets: (a) IP, (b) PU, and (c) KSC.}
	\label{figNumConv}
\end{figure*}

\subsection{Sensitivity Analysis of Parameters}
In this section, our proposed framework with different parameters are studied over three datasets. Three important parameters, including the convolutional kernel size, the number of convolutional kernels ($N$), and the number of convolutional kernels in the efficient non-local module ($L$), are analyzed to find the optimal values. 

Fig. \ref{figKernelSize} reports the classification accuracy when the convolutional kernel size varies from 3 to 7. As shown in Fig. \ref{figKernelSize}, the OA climes with the increasing of kernels size. The OA improves obviously over IP dataset, from 96.7\% to 98.9\%. However, when the kernel size reaches to 7, the improvement is very small. Therefore, the convolutional kernel size of our network is set to 5. 

We jointly investigate the effect of the number of convolutional kernels in FCN and efficient non-local module (i.e., $N$ and $L$). Specifically, $N$ is set from 50 to 175, while $L$ varies from 25 to 150. The kernel number in efficient non-local module is smaller than that in FCN for dimension reduction. Fig. \ref{figNumConv} shows a climbing trend when the kernel numbers $N$ and $L$ increase. It is notable that the OA improves more clearly with the parameter $N$ increasing. It is reasonable because there are five convolutional layers and more learning parameters are added in the network with a larger $N$.  However, on the other hand, a larger kernel number will also increase the network complexity, leading to more training time. Therefore, to ensure a good classification performance and keep efficiency simultaneously, we set $N$ and $L$ to 150 and 150, respectively.


\section{Conclusion}
\label{SecConclusion}
In this study, we propose a deep neural network (ENL-FCN) to learn non-local information for spectral-spatial HSI classification. By introducing an efficient non-local module into the fully convolutional network, both local and non-local information is extracted. Different from the original non-local module, the long-range contextual feature is aggregated in a criss-cross path to facilitate computation and memory efficiency. By using a recurrent operation, each pixel's response can be captured from all other pixels. The efficient non-local module uses fewer learning parameters and less computational memory. Extensive experiments and ablation study over three HSI datasets demonstrate that the proposed method provides promising classification performance with lower computational cost compared with several state-of-the-art deep learning methods.

\ifCLASSOPTIONcaptionsoff
  \newpage
\fi



\bibliographystyle{IEEEtran}
\bibliography{IEEEabrv,bibtex/DL,bibtex/Oldmethods,bibtex/HSI-CC}
\end{document}